\documentclass[]{boogu_class}

\usepackage{hyperref}
\usepackage{url}
\usepackage{graphicx}
\usepackage{booktabs}
\usepackage{xcolor}
\usepackage{amsthm}
\usepackage{amssymb}
\usepackage{mathtools}
\usepackage{algorithm}
\usepackage{algorithmic}

\usepackage{CJKutf8}

\usepackage{ulem} %

\usepackage{tikz}

\usepackage{booktabs}
\usepackage{multirow}
\usepackage{xcolor}
\usepackage{colortbl}
\usepackage{graphicx}

\usepackage{tabularx}
\usepackage{booktabs}
\usepackage[table]{xcolor} %
\colorlet{tablegray}{gray!10} %

\usepackage{xspace}
\makeatletter
\DeclareRobustCommand\onedot{\futurelet\@let@token\@onedot}
\def\@onedot{\ifx\@let@token.\else.\null\fi\xspace}

\def\eg{\emph{e.g}\onedot} 
\def\ie{\emph{i.e}\onedot} 
 
\def\etc{\emph{etc}\onedot}

\def\blfootnote{\gdef\@thefnmark{}\@footnotetext}
\makeatother

\def\model{Boogu}

\title{\model-Image-0.1: Boosting Open Agentic Multimodal Generation via Understanding under a Minimal Budget} %

\definecolor{authorcolor}{RGB}{80, 80, 80}      %

\newcommand{\authorListSize}{\fontsize{9pt}{11.5pt}\selectfont}
\newcommand{\authorListFont}{\sffamily}
\newcommand{\authorListColor}{authorcolor}

\author[]{\authorListSize\color{\authorListColor}\authorListFont
Boogu Team\textsuperscript{*} \par}
\vspace{0.3cm}

\abstract{
We introduce Boogu-Image-0.1, an open-source unified multimodal understanding and generation model family, comprising Base, Turbo, Edit, and Edit-Turbo variants. It delivers competitive performance in high-quality text-to-image generation, fast inference, instruction-based editing, and bilingual (Chinese–English) text rendering. Closed-source multimodal systems like Nano-Banana-Pro and GPT-Image-2 achieve strong performance through system-level integration rather than a single model, yet their internal practices remain largely undisclosed. 
In this work, we demonstrate that strengthening the understanding capability of the system, through a stronger multimodal encoder, agentic prompt rewriting, and related techniques, together with improvements in data quality, training pipelines, and agentic inference-time scaling, can substantially enhance generation and editing performance even under highly constrained compute budgets.
Comprehensive evaluations show that Boogu-Image-0.1 consistently matches or surpasses other open-source models across standard benchmarks, and achieves results approaching leading closed-source systems. Notably, this is accomplished with only 208.62 million unique images. The base model's theoretical training cost is only approximately \$400K. We share practical discussions that we believe are valuable to the broader research community, and release weights, code, and recipes under Apache 2.0 to advance the open ecosystem for unified multimodal understanding and generation. Our code is available here: \url{https://github.com/Boogu-Project/Boogu-Image}.
\vspace{-10mm}
}

\begin{document}
\begin{CJK}{UTF8}{gbsn}   %

\maketitle

\begingroup
\renewcommand{\thefootnote}{\fnsymbol{footnote}}
\footnotetext[1]{Detailed author information can be found in Appendix \ref{app:contributors}.}
\endgroup

\vspace{-2mm}
\begin{figure}[h]
  \centering
  \includegraphics[width=0.9\linewidth]{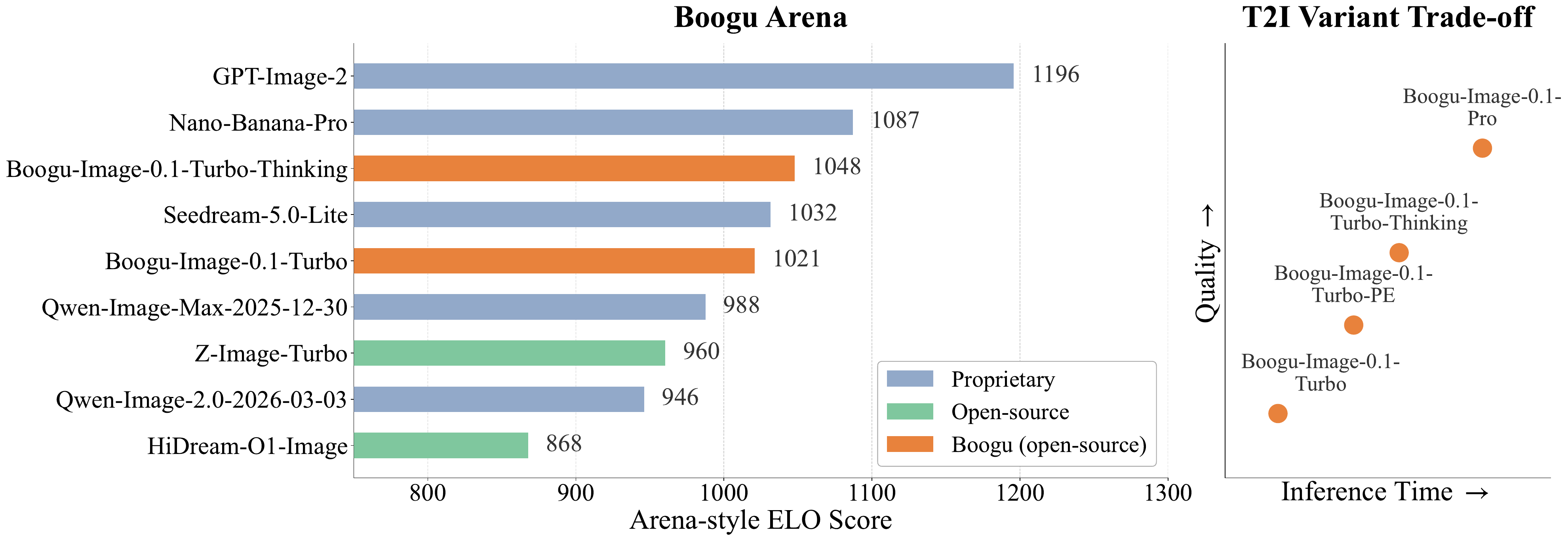}

\caption{\textbf{Left: Performance comparison on {\model} Arena.} {\model}-Image-0.1 achieves top-tier results, outperforming other open-source counterparts as of June 15, 2026. The details of {\model} Arena are provided in Sec~\ref{ref:sub_boogu_arena}. \textbf{Right: Inference time and performance of {\model}-Image variants (illustrative only).} Scaling the inference time leads to higher text-to-image (T2I) generation quality, as illustrated in the figure.}
\label{fig:teaser}
\end{figure}

\clearpage

\begin{figure}[t!]
  \centering
  \includegraphics[width=0.9\linewidth]{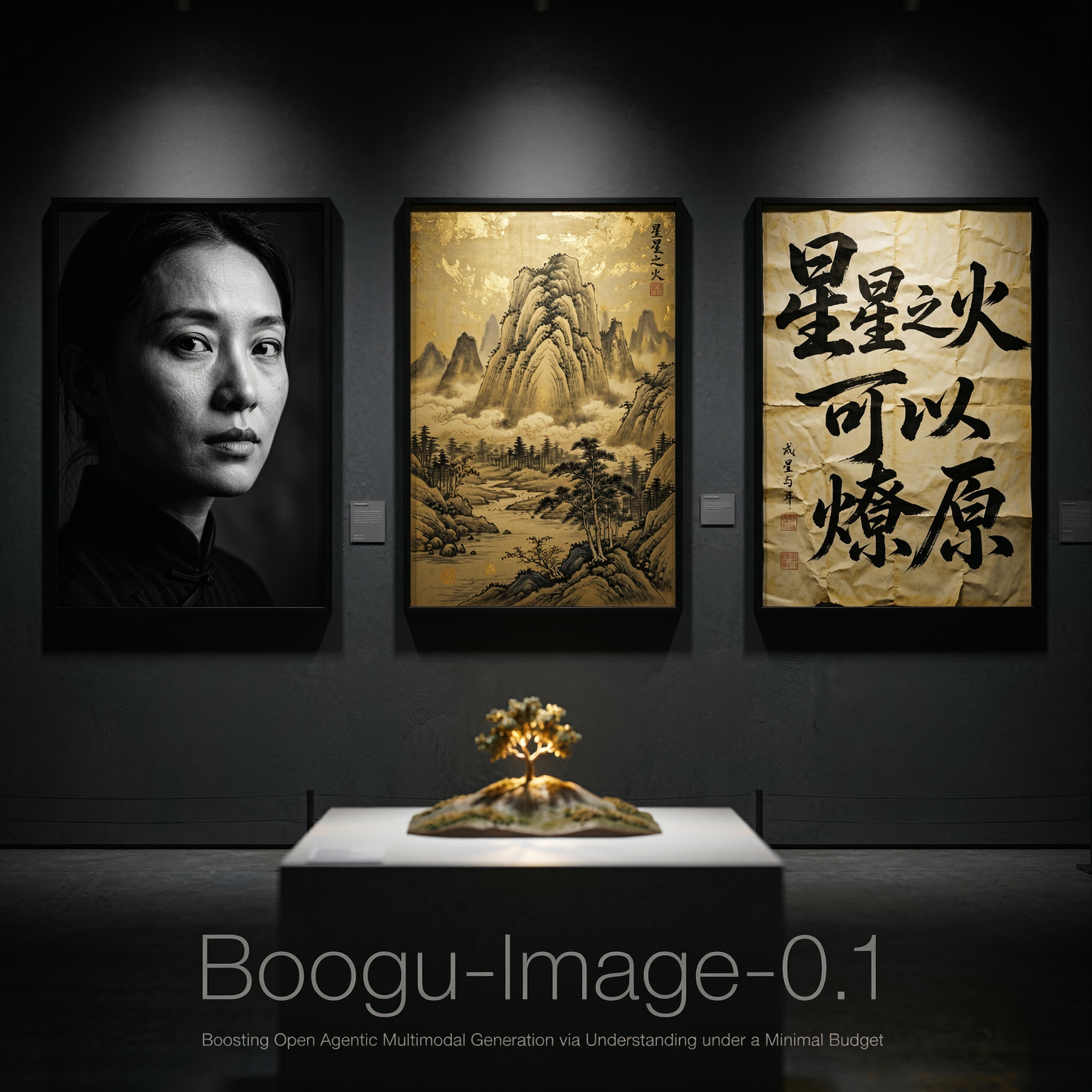}
 \caption{\textbf{{\model}-Image-0.1 generates diverse, high-fidelity images across distinct visual domains within a single unified model.} From left to right: \textbf{(a)} a photorealistic black-and-white portrait demonstrating fine-grained facial details and film-like tonal rendering; \textbf{(b)} a Chinese ink-and-gold landscape painting reflecting culturally grounded stylistic generation; \textbf{(c)} a typographic poster rendering Chinese calligraphy ``A single spark can start a prairie fire'' with accurate stroke structure and layout; \textbf{(d)} a 3D miniature scene of a glowing sapling on a hill, illustrating compositional reasoning and spatial coherence; and\textbf{ (e)} a clean text-rendering result presenting our paper title ``Boogu-Image-0.1: Boosting Open-Source Unified Multimodal Understanding and Generation''.}
  \label{fig:hero_teaser}
\end{figure}

\begin{figure}[t!]
  \centering
  \includegraphics[width=0.95\linewidth]{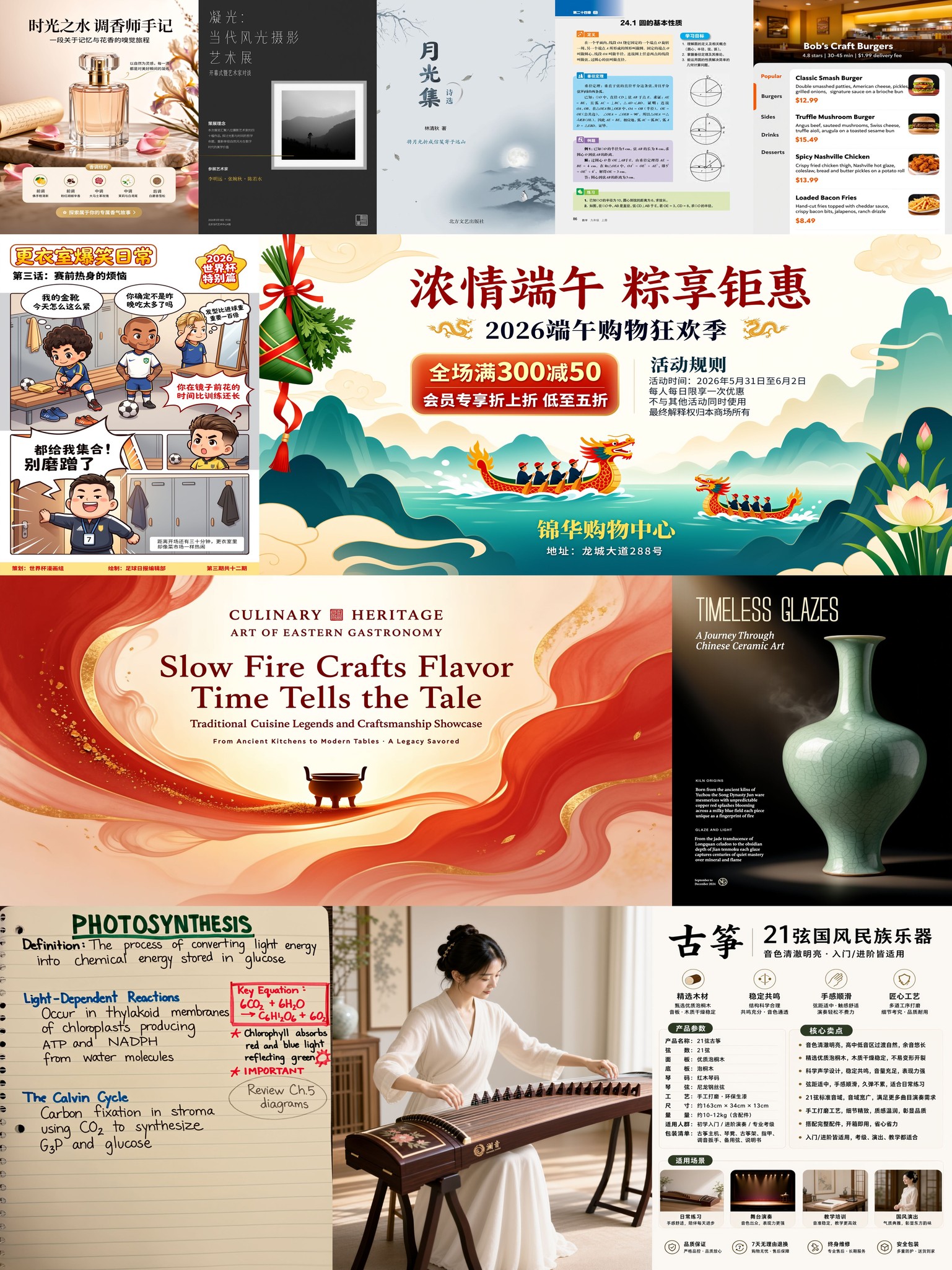}
  \caption{
  \textbf{Boogu's qualitative results on text-rendering.} {\model}-Image-0.1 can accurately render English and Chinese texts with coherent typography and delicate layout design. 
  }
  \label{fig:teaser_text}
\end{figure}

\begin{figure}[t!]
  \centering
  \includegraphics[width=0.95\linewidth]{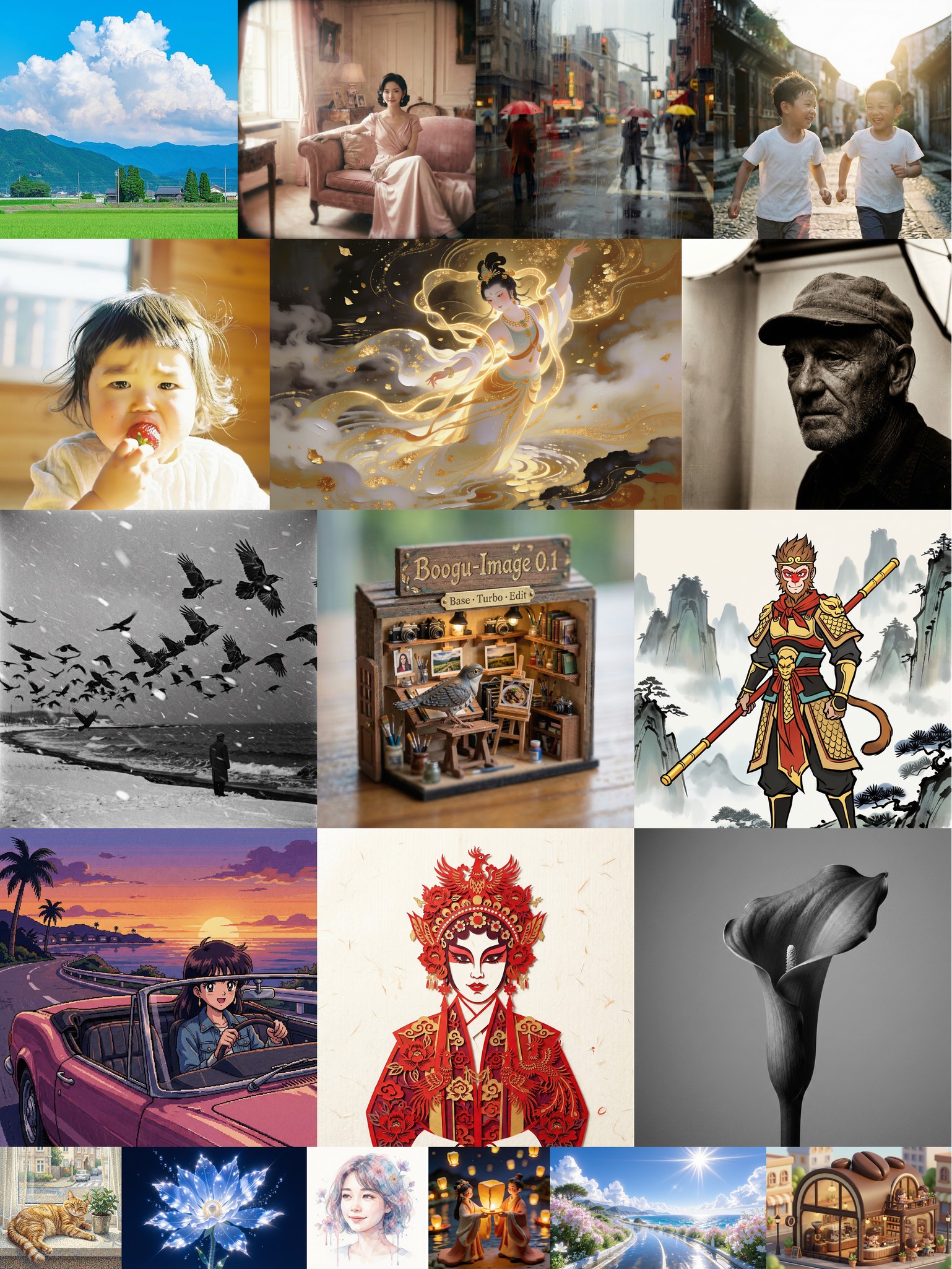}
  \caption{
\textbf{Boogu's qualitative results on photography and style transfer.} {\model}-Image-0.1 generates highly photorealistic images and renders diverse artistic styles across distinct visual domains.
  }
  \label{fig:teaser2}
\end{figure}

\clearpage

\makeatletter
\newcommand{\boogutocchapterstar}[1]{%
  \vskip 6mm
  \begin{center}
    {\chapterfont\sffamily\color{boogureddeep}{#1}}%
  \end{center}
  \vskip 3mm
  \phantomsection
  \addcontentsline{toc}{booguchapterstar}{#1}%
}
\newcommand*{\l@booguchapterstar}[2]{%
  \vskip 2pt
  \begingroup
    \centering\sffamily\large\color{boogureddeep}#1\par
  \endgroup
  \vskip 1pt
}
\let\boogu@origappendix\appendix
\renewcommand{\appendix}{%
  \boogu@origappendix
  \addtocontents{toc}{\protect\setcounter{tocdepth}{1}}%
  \renewcommand{\chapterstar}[1]{%
    \vskip 6mm
    \begin{center}
      {\chapterfont\sffamily\color{boogureddeep}{##1}}%
    \end{center}
    \vskip 3mm
  }%
}
\makeatother

\begingroup
\setcounter{tocdepth}{3}
\small\fontfamily{cmss}\selectfont\color{boogured}
\renewcommand{\contentsname}{Contents}
\setlength{\cftbeforetoctitleskip}{-6pt}
\setlength{\cftaftertoctitleskip}{4pt}
\setlength{\cftbeforesecskip}{4pt}
\setlength{\cftbeforesubsecskip}{1.8pt}
\setlength{\cftbeforesubsubsecskip}{1pt}
\setlength{\cftsecindent}{0pt}
\setlength{\cftsubsecindent}{1.2em}
\setlength{\cftsubsubsecindent}{2.4em}
\renewcommand{\cfttoctitlefont}{\fontfamily{cmss}\fontseries{bx}\selectfont\sectionfont\color{boogured}}
\renewcommand{\cftsecfont}{\fontfamily{cmss}\fontseries{bx}\selectfont\color{boogured}}
\renewcommand{\cftsubsecfont}{\fontfamily{cmss}\fontseries{sbc}\selectfont\small\color{boogured}}
\renewcommand{\cftsubsubsecfont}{\fontfamily{cmss}\selectfont\mdseries\footnotesize\color{boogured}}
\renewcommand{\cftsecpagefont}{\fontfamily{cmss}\fontseries{bx}\selectfont\color{boogured}}
\renewcommand{\cftsubsecpagefont}{\fontfamily{cmss}\fontseries{sbc}\selectfont\small\color{boogured}}
\renewcommand{\cftsubsubsecpagefont}{\fontfamily{cmss}\selectfont\mdseries\footnotesize\color{boogured}}
\tableofcontents
\endgroup

\clearpage

\boogutocchapterstar{Preamble}

\section{Introduction}
\label{sec:intro}

Recent closed-source systems such as the GPT-Image series~\cite{openai2025gptimage1,openai2026gptimage2}, the Nano-Banana series~\cite{google2026nanobanana2,google2025nanobananapro}, and the Seedream series~\cite{seedream2025seedream40nextgenerationmultimodal,seedream4d5,seedream5lite} have advanced at a striking pace: they not only render fine-grained visual details with remarkable fidelity, but also reason over complex user instructions and, in some cases, augment their world knowledge through web search at generation time. Open-source efforts such as FLUX~\cite{bfl2025flux2,flux2024blackforest}, Ideogram~\cite{ideogram-4-2026}, Qwen-Image~\cite{qwen2025qwenimage,zhao2026qwen2}, Z-Image~\cite{zimage2025}, and Hunyuan-Image~\cite{cao2025hunyuanimage} have made tremendous progress in closing this gap, primarily by improving visual quality and by attaching an auxiliary LLM to rewrite user instructions. To push the boundaries of open-source models, we treat understanding as a first-class design target alongside data quality, the training recipe, and the inference pipeline. Results show that coordinated improvements across these three axes yield substantial gains in instruction following and generation fidelity.

Our emphasis on understanding stems from a conviction that image generation has entered a new era, evolving from Text-to-Image toward Requirement-to-Image Generation. Users are no longer satisfied with a single descriptive prompt; they expect models to interpret complex intentions, implicit constraints, multi-level instructions, and cross-modal contextual cues. Moreover, real-world requirements are highly diverse: they range from simple to intricate and vary in cost and latency expectations, so a single model configuration rarely serves them all equally well. Understanding is the critical bridge connecting ambiguous human requirements to precise visual generation, and to the appropriate strategy under each user's constraints.

To realize this understanding, we decompose it into system components, beginning with how the model perceives language. We regard the text encoder as the sensor for language: a sufficiently capable text encoder is indispensable for interpreting users' requirements. To this end, we systematically analyze models of varying scales and confirm that stronger LLMs yield stronger text-encoding capabilities—a trend noted, though not systematically studied, in prior works~\cite{xie2024sana,chen2026lens,zhao2026qwen2}. Guided by this finding, we adopt Qwen3-VL-8B~\cite{bai2025qwen3vl}, balancing text-encoding capability against a parameter budget acceptable to the open-source community.

We further employ Agentic Image Generation, which injects understanding into generation at inference time~\cite{huang2026ape,zhang2026qwenimageagentbridgingcontextgap}. Instead of treating the model as a passive text-to-image mapping, we wrap it in an agent that interprets the user's request and orchestrates generation accordingly. First, the agent performs agentic prompt rewriting~\cite{huang2026ape,liu2026ernie}, refining the prompt to better capture the intended meaning. Second, it selects among different model variants (\eg, Base vs. Turbo) to balance speed and quality across diverse scenarios. Third, it incorporates inference-time techniques such as Reflection~\cite{li2025reflect} to further improve quality. As shown in Figure~\ref{fig:teaser}, model performance improves progressively as more understanding ability is injected.

Beyond understanding, training efficiency is a core design principle. Under a constrained compute budget, we design a data curation pipeline that prioritizes quality over quantity with various training optimization strategies. This allows us to train a text-to-image base model from scratch using only 208.62 million unique images, at a total cost of roughly \$400K, while achieving performance competitive with the state of the art.

Equally important, we find that many seemingly secondary details are in fact decisive in practice, including evaluation protocols~\cite{ghosh2023geneval,hu2024ella}, data filtering~\cite{qwen2025qwenimage}, and caption design~\cite{zimage2025}. Indeed, their opacity is itself a major barrier to open-source progress: they are often learned through costly trial and error within industrial teams, yet rarely described in public technical reports. By documenting them, we aim to reduce the hidden engineering cost for the open-source community and provide practical reference points for future work.

Building on these designs, we present {\model}-Image-0.1, an open-source system that unifies understanding and generation. By promoting understanding to a first-class component, it moves beyond conventional Text-to-Image pipelines toward the Requirement-to-Image paradigm, faithfully translating complex user intents into high-fidelity images. It achieves top-tier performance on {\model} Arena and state-of-the-art results among open-source models on the recently released Qwen-Image-Bench~\cite{li2026qwenimagebenchgenerationcreationtexttoimage}. We release the model along with our findings and practical lessons to support future research on understanding-driven generation. We hope {\model}-Image-0.1 marks a step toward image generation systems that truly understand what users want.

\clearpage

\clearpage

\clearpage

\boogutocchapterstar{Evaluation}

\section{Experimental Results}

\label{sec:eval}

\subsection{Rethinking Open-Source Research Evaluation} 
\label{sec:eval_rethink_eval}
\begin{figure}[h]
  \centering
  \includegraphics[width=0.97\linewidth]{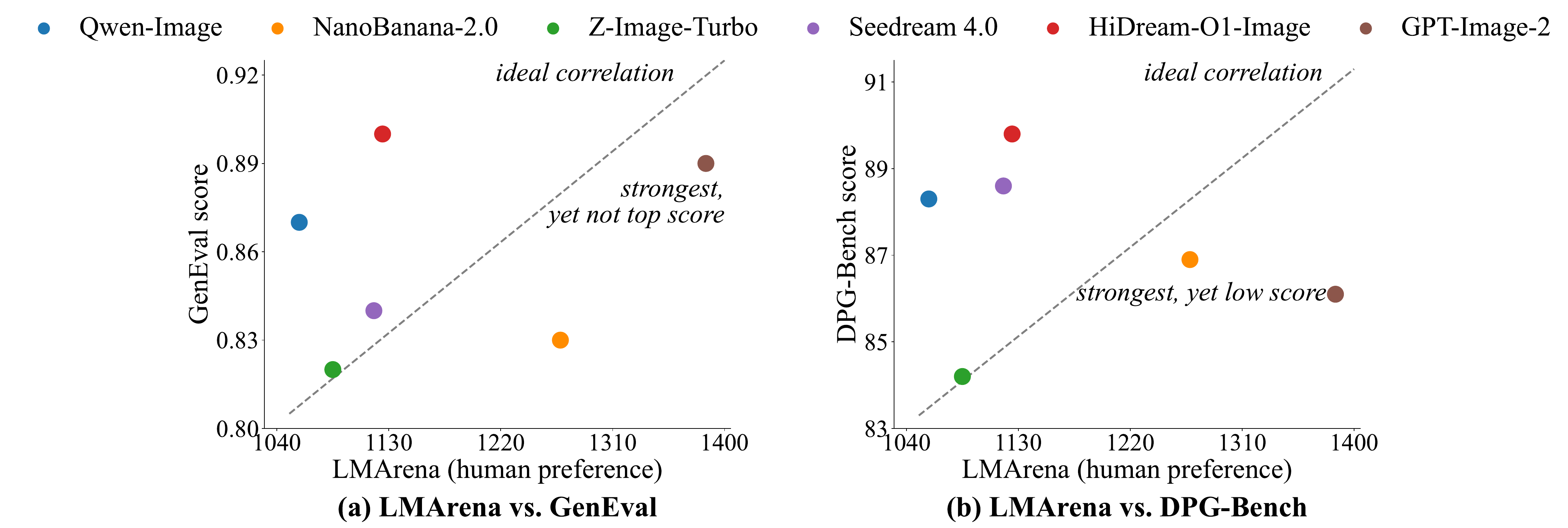}
\caption{\textbf{Public benchmarks fail to track human preference.} We compare six recent text-to-image models, plotting their LMArena~\cite{zheng2024chatbot} Elo (a human-preference leaderboard reflecting real-world capability) against their \textbf{(a)} GenEval~\cite{ghosh2023geneval} and \textbf{(b)} DPG-Bench~\cite{hu2024ella} scores. The gray dashed line marks the ideal case of perfect rank correlation. Strikingly, the strongest model by human preference, GPT-Image-2, ranks only in the middle of the pack on both benchmarks. These inversions reveal a gap between saturated academic benchmarks and genuine generation quality.}
\label{fig:Public_benchmarks}
\end{figure}

Public benchmarks remain valuable~\cite{niu2025wise,du2025textcrafter,chang2025oneigbenchomnidimensionalnuancedevaluation,ye2026imgedit,jiang2026geditbench}, but only when used and interpreted correctly. The reality is that few state-of-the-art closed-source models report results on GenEval~\cite{ghosh2023geneval} or DPG-Bench~\cite{hu2024ella}, even though these models are widely recognized as far stronger than most open-source ones. To illustrate this, we compare six recent text-to-image models on their human-preference ranking (LMArena Elo) versus their scores on GenEval and DPG-Bench (Figure~\ref{fig:Public_benchmarks}). If these benchmarks tracked real capability, stronger models should score higher; instead, we observe clear rank inversions. Most strikingly, GPT-Image-2~\cite{openai2026gptimage2}, the strongest by human preference, ranks only mid-pack on both benchmarks. We observe the same inversions in our own models: on GenEval~\cite{ghosh2023geneval}, a SANA-VAE configuration of our model reaches 0.92, yet our deployed and stronger model scores only 0.85. Consequently, these benchmarks no longer reliably correlate with model capability. We attribute this to three causes.

\begin{itemize}
    \item \textbf{Many public benchmarks do not reflect real application scenarios.} 
    They assess isolated capabilities under constrained, synthetic conditions that diverge substantially from production usage. As a result, a model that surpasses Nano-Banana-Pro \cite{google2025nanobananapro} across academic benchmarks may still deliver a markedly worse user experience in deployment—a misalignment between the distributions sampled by static benchmarks and those encountered in practice. Evaluation should be derived from target application scenarios rather than treated as an independent objective: the question is not how a model scores on a benchmark, but whether a defined user requirement is met and how it can be measured directly. 

    \item \textbf{Most widely adopted benchmarks have approached saturation.} 
    State-of-the-art models routinely achieve near-ceiling scores, compressing the dynamic range for comparison and rendering small score differences statistically uninformative. In this regime, observed gaps are driven by variance, prompt formatting, and benchmark-specific overfitting rather than genuine capability differences. 

    \item \textbf{Data contamination and test-set leakage are pervasive.} 
    Large-scale pre-training corpora frequently incorporate public evaluation sets, so reported scores may reflect memorization rather than generalization. Inconsistent reporting standards compound the problem: some results are obtained under training conditions that include leaked evaluation data, undermining both the interpretability of individual scores and the validity of cross-model comparisons.
\end{itemize}

\clearpage

\subsection{Text-to-Image Generation} 
\label{sec:t2i_evaluation}
\begin{figure*}[htb]
  \centering
  \includegraphics[width=\linewidth]{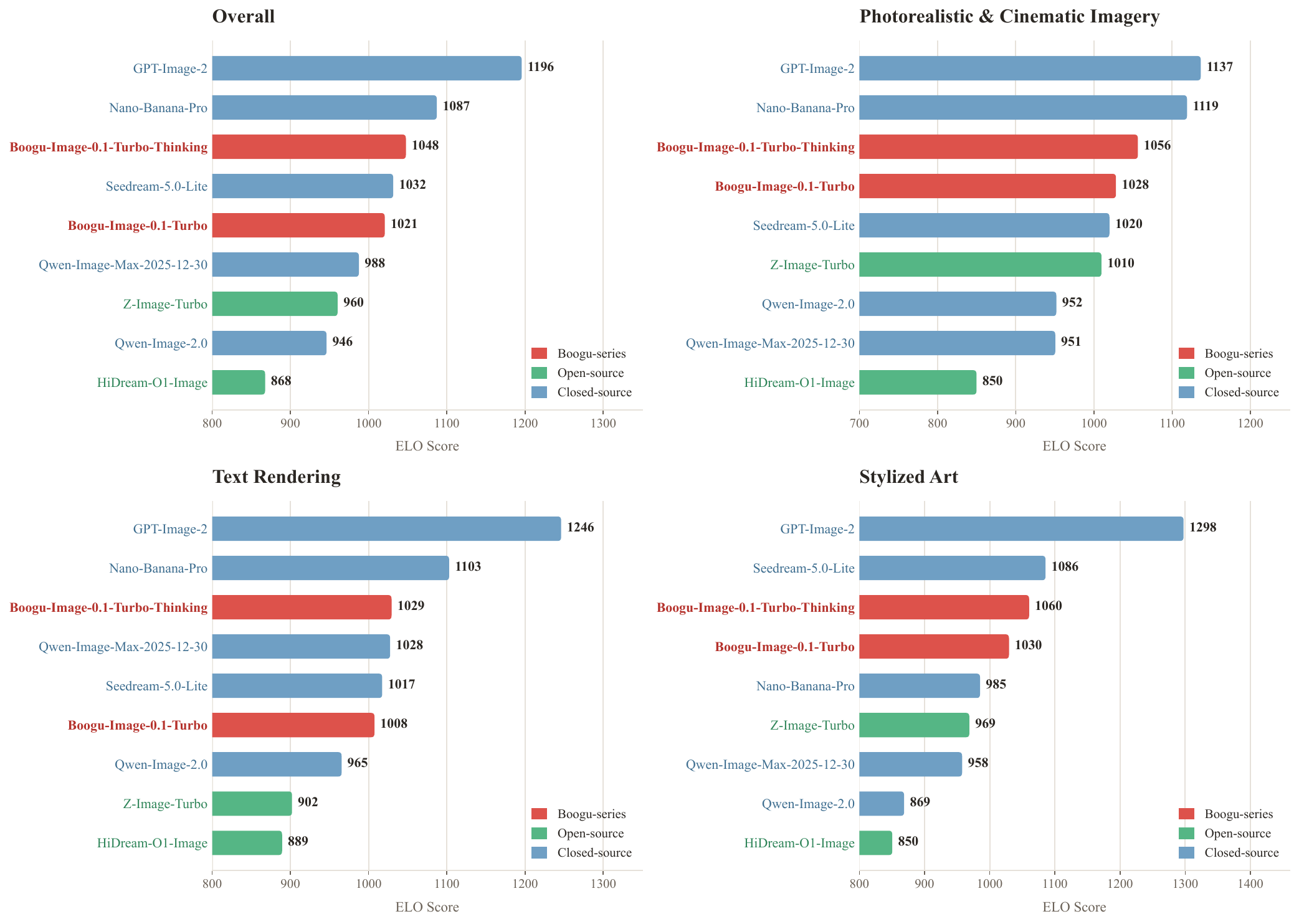}
\caption{\textbf{Boogu Arena Text-to-Image Elo scores.} Pairwise blind votes are aggregated into Elo ratings, where higher scores indicate stronger relative human preference. The four panels display the overall ratings alongside three specific categories. Across all categories, Boogu models lead the open-source tier, trailing only the closed-source frontier models (GPT-Image-2~\cite{openai2026gptimage2} and Nano-Banana-Pro~\cite{google2025nanobananapro}). Notably, the thinking variant yields the most substantial gains in the prompt-sensitive text-rendering category. Note: Qwen-Image-2.0 is Qwen-Image-2.0-2026-03-03. }  \label{fig:arena_t2i_elo_score}
\end{figure*}

\subsubsection{Boogu Arena Benchmark}
\label{ref:sub_boogu_arena}

\begin{figure*}[htb]
  \centering
  \includegraphics[width=0.70\linewidth]{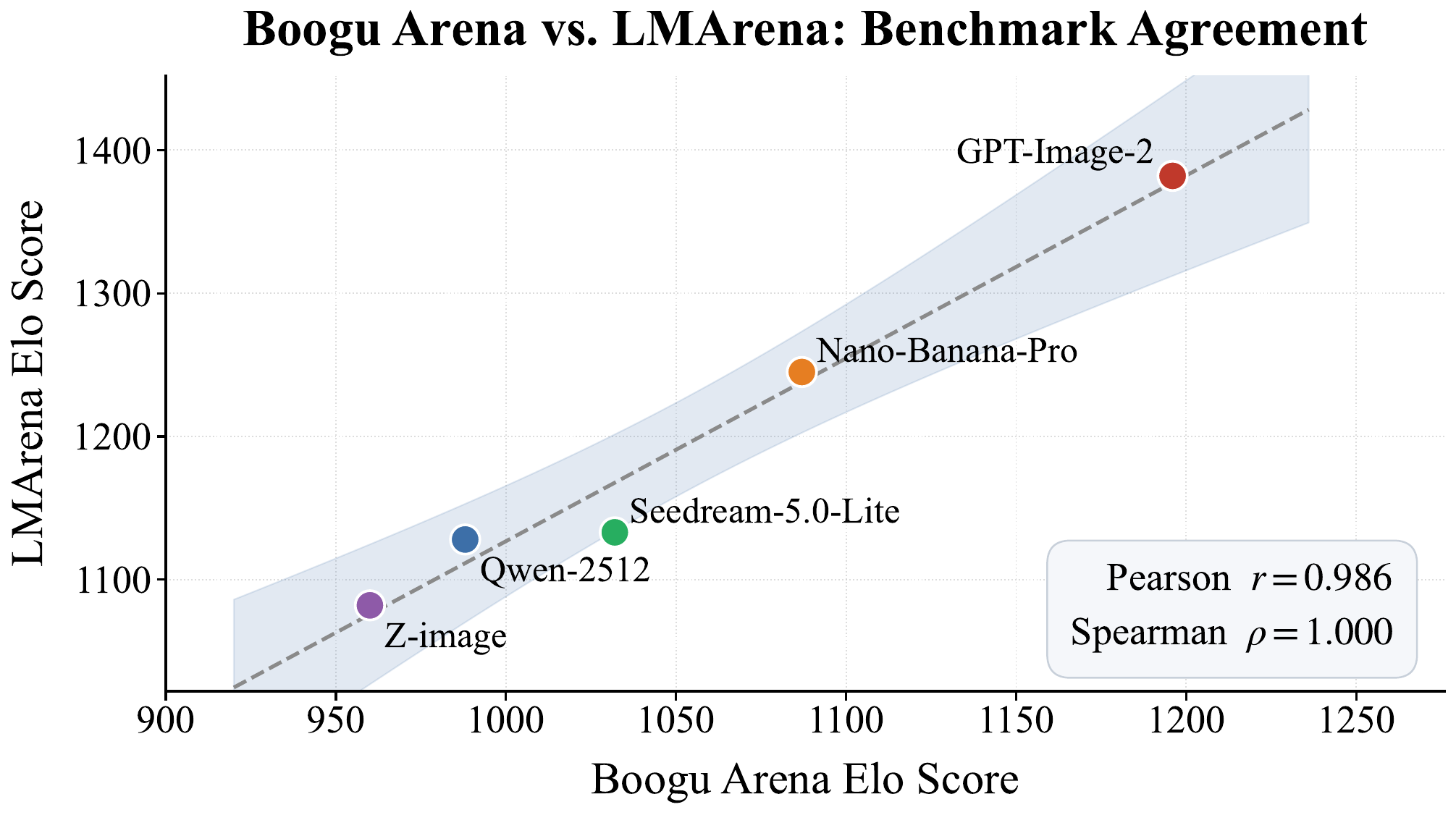}
    \caption{\textbf{Agreement between Boogu Arena and LMArena Elo ratings.} Each point corresponds to one text-to-image model, with its Elo score on Boogu Arena (x-axis) plotted against its score on the public LMArena leaderboard (y-axis). The dashed line shows a linear fit with the shaded region indicating the 95\% confidence band. The two benchmarks exhibit strong agreement, with a Pearson correlation of $r = 0.986$ and a perfect Spearman rank correlation of $\rho = 1.0$, indicating that Boogu Arena faithfully reproduces the model ranking established by large-scale public human-preference voting.}
  \label{fig:booguarena_vs_arena}
\end{figure*}

Since we do not have access to the \textit{arena.ai} \cite{arena_ai_2026} evaluation platform, we introduce the {\model} Arena to ensure a fair comparison to the greatest extent possible. Specifically, we built an in-house benchmark that faithfully emulates the original Arena pipeline (see Figure \ref{fig:booguarena_vs_arena}), including its prompt design and blind pairwise battles.

\paragraph{Prompt design.} We organize the prompt space into three core application categories: photorealistic and cinematic imagery, text rendering, and stylized art. Each category is populated with roughly 400 fine-grained sub-scene keywords. To better reflect real-world user behavior, a powerful VLM further instantiates each prompt along two axes. The first is prompt length, spanning short, medium, and long at a $3{:}4{:}3$ ratio. The second is user persona, covering 27 roles across novice, intermediate, and professional tiers at a $5{:}3{:}2$ ratio. The two axes are coupled rather than drawn independently. The long paragraph-style prompts, for instance, are never paired with novice personas. This procedure yields 1{,}200 bilingual prompts in total. Each prompt is authored in both Chinese and English, and a single language is randomly selected at test time and shared across all compared models. To ensure evaluation fairness, each model is evaluated only once per prompt using a random seed. We will release the designed prompts.

\paragraph{Blind battle.} We conduct a strictly blind pairwise evaluation. For every prompt we present the outputs of two anonymized models side by side, the annotator votes among ``A is better'', ``B is better'', ``tie good'', and ``tie bad'', and the model identities are revealed only after the vote is cast. Sample selection and model pairing both follow a weighted inverse-frequency strategy so that every prompt and every model receives sufficient coverage, and the collected votes are aggregated into Elo scores via a Bradley-Terry model.

Following this evaluation protocol, we compare our Boogu models (Boogu-Image-0.1-Turbo and Boogu-Image-0.1-Turbo-Thinking) against two open-source baselines (Z-Image-Turbo~\cite{zimage2025} and HiDream-O1-Image~\cite{hidreamolimage}) and five closed-source counterparts (GPT-Image-2~\cite{openai2026gptimage2}, Nano-Banana-Pro~\cite{google2025nanobananapro}, Seedream-5.0-Lite~\cite{seedream5lite}, Qwen-Image-Max-2025-12-30~\cite{qwenimage2512}, and Qwen-Image-2.0-2026-03-03~\cite{zhao2026qwen2}). {In total, we collected \underline{over 4,000 votes} from the pairwise battles conducted among these nine models.}

Figure \ref{fig:arena_t2i_elo_score} details the overall and per-category Elo scores of the nine evaluated models. Boogu models consistently dominate the open-source tier across all categories, surpassed only by top closed-source models (GPT-Image-2~\cite{openai2026gptimage2} and Nano-Banana-Pro~\cite{google2025nanobananapro}). This advantage is most evident in photorealistic and stylized content, where Boogu-Image-0.1-Turbo-Thinking substantially closes the gap with the closed-source frontier. In text rendering, the thinking variant clearly outperforms its non-thinking counterpart. We attribute this to the high prompt sensitivity of text generation: thinking-based prompt enhancement explicitly includes the exact text to be rendered, producing better-structured prompts that improve text correctness.

\subsubsection{Qwen-Image Benchmark} 
\label{sec:qwen_bench}
Qwen-Image-Bench~\cite{li2026qwenimagebenchgenerationcreationtexttoimage} is a recently released, high-quality benchmark, published after we froze our text-to-image development. Because it is less affected by common issues in long-standing benchmarks such as data leakage, we regard it as a reliable testbed for modern image generation models. As shown in Tables~\ref{tab:qwen_image_bench_cn} and~\ref{tab:qwen_image_bench_en}, Boogu-Image-0.1-Base-Thinking achieves the best overall score among all evaluated open-source models under both Chinese (53.57) and English (53.73) prompts, surpassing strong open-source baselines such as Qwen-Image-2512~\cite{qwenimage2512} and HunyuanImage-3.0~\cite{cao2025hunyuanimage}. The improvement from the thinking variant is most pronounced in the Creativity dimension, where Base-Thinking gains over 8 points over its non-thinking counterpart (e.g., 48.62
→56.74 in Chinese), consistent with our understanding-driven design. These trends are stable across both languages, indicating robust bilingual capability. Due to time constraints, the evaluation does not yet cover all available open-source baselines.

\begin{table*}[htb]
\centering
\small
\setlength{\tabcolsep}{4pt}
\renewcommand{\arraystretch}{1.08}
\resizebox{\textwidth}{!}{
\begin{tabular}{l l c c c c c c}
\toprule
\textbf{Model} & \textbf{License} & \textbf{Quality} & \textbf{Aesthetics} & \textbf{Alignment} & \textbf{Realism} & \textbf{Creativity} & \textbf{Overall} $\uparrow$ \\
\midrule
\rowcolor{tablegray} GPT-Image-2~\cite{openai2026gptimage2} & Closed & \textbf{58.65} & \textbf{67.53} & \textbf{65.85} & \textbf{57.38} & \textbf{75.23} & \textbf{64.69} \\
\rowcolor{tablegray} Nano-Banana-2.0~\cite{google2026nanobanana2} & Closed & 54.77 & \underline{61.08} & \underline{62.40} & \underline{54.28} & \underline{67.05} & \underline{59.82} \\
\rowcolor{tablegray} Nano-Banana-Pro~\cite{google2025nanobananapro} & Closed & \underline{55.67} & 60.26 & 61.25 & 54.07 & 66.23 & 59.45 \\
\rowcolor{tablegray} Qwen-Image-2.0-Pro~\cite{zhao2026qwen2} & Closed & 54.39 & 58.67 & 59.28 & 51.83 & 64.94 & 57.84 \\
\rowcolor{tablegray} Seedream-5.0-Lite~\cite{seedream5lite} & Closed & 52.55 & 58.40 & 58.90 & 51.92 & 65.29 & 57.22 \\
\rowcolor{tablegray} FLUX-2-Max~\cite{blackforestlabs2025flux2} & Closed & 53.64 & 56.85 & 57.35 & 49.35 & 56.50 & 55.33 \\
\rowcolor{tablegray} GPT-Image-1~\cite{openai2025gptimage1} & Closed & 52.34 & 55.09 & 56.28 & 48.14 & 55.78 & 54.07 \\
\textcolor{boogured}{Boogu-Image-0.1-Base-Thinking} & \textcolor{boogured}{Apache-2.0} & \textcolor{boogured}{50.58} & \textcolor{boogured}{55.20} & \textcolor{boogured}{55.99} & \textcolor{boogured}{47.35} & \textcolor{boogured}{56.74} & \textcolor{boogured}{53.57} \\
\textcolor{boogured}{Boogu-Image-0.1-Turbo-Thinking} & \textcolor{boogured}{Apache-2.0} & \textcolor{boogured}{50.89} & \textcolor{boogured}{54.36} & \textcolor{boogured}{56.28} & \textcolor{boogured}{47.67} & \textcolor{boogured}{53.33} & \textcolor{boogured}{53.13} \\
Qwen-Image-2512~\cite{qwenimage2512} & Apache-2.0 & 51.76 & 54.74 & 52.72 & 47.00 & 50.19 & 52.06 \\
\rowcolor{tablegray} Imagen-4.0-Ultra~\cite{googleimagen4} & Closed & 50.90 & 54.25 & 54.02 & 45.59 & 51.14 & 51.99 \\
\textcolor{boogured}{Boogu-Image-0.1-Turbo} & \textcolor{boogured}{Apache-2.0} & \textcolor{boogured}{51.24} & \textcolor{boogured}{53.50} & \textcolor{boogured}{53.54} & \textcolor{boogured}{46.11} & \textcolor{boogured}{48.91} & \textcolor{boogured}{51.53} \\
\textcolor{boogured}{Boogu-Image-0.1-Base} & \textcolor{boogured}{Apache-2.0} & \textcolor{boogured}{50.41} & \textcolor{boogured}{53.14} & \textcolor{boogured}{52.91} & \textcolor{boogured}{45.42} & \textcolor{boogured}{48.62} & \textcolor{boogured}{50.96} \\
HunyuanImage-3.0~\cite{cao2025hunyuanimage} & Other & 50.35 & 53.57 & 52.00 & 44.31 & 49.12 & 50.81 \\
\rowcolor{tablegray} Imagen-4.0~\cite{googleimagen4} & Closed & 50.16 & 52.68 & 51.64 & 44.84 & 47.94 & 50.29 \\
Qwen-Image~\cite{qwen2025qwenimage} & Apache-2.0 & 48.44 & 52.25 & 50.72 & 43.16 & 47.30 & 49.23 \\
\rowcolor{tablegray} Kling-Image-2.1~\cite{klingimage21} & Closed & 49.11 & 50.15 & 49.18 & 44.74 & 44.67 & 48.26 \\
GLM-Image~\cite{glmimage2026} & Apache-2.0 & 49.26 & 50.64 & 47.90 & 44.69 & 45.23 & 48.19 \\
\bottomrule
\end{tabular}
}
\caption{\textbf{Our model series achieves top-tier performance among the evaluated open-source models on Qwen-Image-Bench (Chinese prompts).} Gray shading denotes closed-source models, and our models are highlighted in \textcolor{boogured}{\textbf{red}}. Best scores are \textbf{bolded} and second-best scores are \underline{underlined}. Results for other methods are taken from~\cite{li2026qwenimagebenchgenerationcreationtexttoimage}.}
\label{tab:qwen_image_bench_cn}
\end{table*}

\begin{table*}[!h]
\centering
\small
\setlength{\tabcolsep}{4pt}
\renewcommand{\arraystretch}{1.08}
\resizebox{\textwidth}{!}{
\begin{tabular}{l l c c c c c c}
\toprule
\textbf{Model} & \textbf{License} & \textbf{Quality} & \textbf{Aesthetics} & \textbf{Alignment} & \textbf{Realism} & \textbf{Creativity} & \textbf{Overall} $\uparrow$ \\
\midrule
\rowcolor{tablegray} GPT-Image-2~\cite{openai2026gptimage2} & Closed & \textbf{59.09} & \textbf{68.48} & \textbf{65.78} & \textbf{59.40} & \textbf{75.34} & \textbf{65.23} \\
\rowcolor{tablegray} Nano-Banana-2.0~\cite{google2026nanobanana2} & Closed & 54.86 & 62.63 & 61.11 & 54.66 & 64.49 & 59.59 \\
\rowcolor{tablegray} Nano-Banana-Pro~\cite{google2025nanobananapro} & Closed & 55.30 & 61.38 & 60.30 & \underline{55.91} & 64.54 & 59.33 \\
\rowcolor{tablegray} Qwen-Image-2.0-Pro~\cite{zhao2026qwen2} & Closed & 55.16 & 60.36 & 57.86 & 53.06 & 63.59 & 57.90 \\
\rowcolor{tablegray} Seedream-5.0-Lite~\cite{seedream5lite} & Closed & 54.01 & 59.96 & 58.63 & 53.86 & 63.64 & 57.80 \\
\rowcolor{tablegray} FLUX-2-Max~\cite{blackforestlabs2025flux2} & Closed & 53.99 & 58.77 & 57.31 & 50.69 & 57.79 & 56.14 \\
\rowcolor{tablegray} GPT-Image-1~\cite{openai2025gptimage1} & Closed & 52.50 & 55.77 & 55.35 & 48.25 & 57.29 & 54.24 \\
\textcolor{boogured}{Boogu-Image-0.1-Base-Thinking} & \textcolor{boogured}{Apache-2.0} & \textcolor{boogured}{51.52} & \textcolor{boogured}{55.89} & \textcolor{boogured}{55.58} & \textcolor{boogured}{47.23} & \textcolor{boogured}{56.24} & \textcolor{boogured}{53.73} \\
\textcolor{boogured}{Boogu-Image-0.1-Turbo-Thinking} & \textcolor{boogured}{Apache-2.0} & \textcolor{boogured}{51.92} & \textcolor{boogured}{56.28} & \textcolor{boogured}{55.67} & \textcolor{boogured}{47.89} & \textcolor{boogured}{52.70} & \textcolor{boogured}{53.58} \\
\rowcolor{tablegray} Imagen-4.0-Ultra~\cite{googleimagen4} & Closed & 51.16 & 55.64 & 53.75 & 46.00 & 51.32 & 52.42 \\
\textcolor{boogured}{Boogu-Image-0.1-Turbo} & \textcolor{boogured}{Apache-2.0} & \textcolor{boogured}{51.30} & \textcolor{boogured}{53.91} & \textcolor{boogured}{53.37} & \textcolor{boogured}{46.52} & \textcolor{boogured}{48.90} & \textcolor{boogured}{51.61} \\
HunyuanImage-3.0~\cite{cao2025hunyuanimage} & Other & 50.76 & 54.66 & 53.16 & 45.33 & 48.33 & 51.35 \\
Qwen-Image-2512~\cite{qwenimage2512} & Apache-2.0 & 51.84 & 54.40 & 51.44 & 47.80 & 47.75 & 51.32 \\
\rowcolor{tablegray} Imagen-4.0~\cite{googleimagen4} & Closed & 50.63 & 53.93 & 52.56 & 45.13 & 48.56 & 51.08 \\
\textcolor{boogured}{Boogu-Image-0.1-Base} & \textcolor{boogured}{Apache-2.0} & \textcolor{boogured}{50.38} & \textcolor{boogured}{53.90} & \textcolor{boogured}{52.62} & \textcolor{boogured}{46.54} & \textcolor{boogured}{47.52} & \textcolor{boogured}{51.00} \\
\rowcolor{tablegray} Kling-Image-2.1~\cite{klingimage21} & Closed & 49.04 & 50.94 & 50.47 & 44.29 & 46.23 & 48.89 \\
Qwen-Image~\cite{qwen2025qwenimage} & Apache-2.0 & 48.45 & 51.18 & 50.04 & 43.45 & 45.37 & 48.48 \\
GLM-Image~\cite{glmimage2026} & Apache-2.0 & 49.86 & 49.98 & 47.49 & 44.25 & 44.67 & 47.86 \\
\bottomrule
\end{tabular}
}
\caption{\textbf{Our model series achieves top-tier performance among the evaluated open-source models on Qwen-Image-Bench (English prompts).} 
Gray shading denotes closed-source models, and our models are highlighted in \textcolor{boogured}{\textbf{red}}. Best scores are \textbf{bolded} and second-best scores are \underline{underlined}. Results for other methods are taken from~\cite{li2026qwenimagebenchgenerationcreationtexttoimage}.}
\label{tab:qwen_image_bench_en}
\end{table*}

\clearpage

\subsubsection{LongText Benchmark} 
\label{sec:longtext_bench}
LongText-Bench~\cite{geng2025x} is a benchmark for evaluating long-form text rendering in image generation, a capability that is increasingly important for practical applications such as posters, slides, documents, comics, and instruction-heavy visual designs. Similar to Qwen-Image-Bench~\cite{li2026qwenimagebenchgenerationcreationtexttoimage}, it adopts a model-based annotation protocol that assesses not only OCR-level correctness, but also readability, layout consistency, and whether the generated image preserves the intended textual content and visual structure. 

As shown in Table~\ref{tab:longtextbench}, our models achieve top-tier text-rendering performance among the compared open-source models, and are particularly strong on Chinese (ZH) prompts, where Boogu-Image-0.1-Turbo-Thinking ranks second overall (0.985), trailing only the closed-source Seedream-4.5. We note, however, that this benchmark primarily measures OCR-level text accuracy on relatively short sequences (fewer than 100 characters) and does not fully capture overall visual quality in dense layouts. Under such dense-text settings, we observe that the Base model produces more coherent typography than Turbo, whereas Turbo, despite scoring comparably on this benchmark, tends to introduce visible artifacts in densely packed regions. Due to time constraints, the evaluation does not yet cover all available open-source baselines.

\noindent\textbf{Limitations of LongText-Bench.} While useful, this benchmark has two notable limitations. First, its model-based protocol primarily scores whether the target characters are correctly rendered (\ie, OCR-level accuracy), but does not assess the \emph{fidelity} of the rendered text, such as whether the glyphs are visually natural, well-integrated with the surrounding layout and materials, and free of artifacts. As a result, a model can attain a high score even when the rendered text exhibits noticeable visual flaws. Second, the benchmark is largely saturated: apart from a single outlier, the top-ranked models are tightly clustered within a narrow range (roughly 0.96–0.99 in average score), leaving little headroom to discriminate among strong models. We therefore view LongText-Bench as a reliable check for text correctness, but recommend complementing it with human evaluation or perceptual metrics when assessing the overall quality of long-form text rendering.

\begin{table}[h!]
\small
{\centering
\begin{tabular}{lcccc}
\toprule
\textbf{Model} & \textbf{Open Source} & \textbf{AVG $\uparrow$} & \textbf{EN $\uparrow$} & \textbf{ZH $\uparrow$} \\
\midrule
\rowcolor{tablegray}
Seedream-4.5~\cite{seedream4d5} & Closed & \textbf{0.988} & \textbf{0.989} & \textbf{0.987} \\
HiDream-O1-Image~\cite{hidreamolimage} & MIT & \underline{0.979} & 0.979 & 0.978 \\
ERNIE-Image-PE~\cite{liu2026ernie} & Apache-2.0 & 0.973 & \underline{0.980} & 0.966 \\
\textcolor{boogured}{Boogu-Image-0.1-Turbo-Thinking} & \textcolor{boogured}{Apache-2.0} & \textcolor{boogured}{0.971} & \textcolor{boogured}{0.957} & \textcolor{boogured}{\underline{0.985}} \\
GLM-Image~\cite{glmimage2026} & Apache-2.0 & 0.966 & 0.952 & 0.979 \\
\rowcolor{tablegray}
Nano-Banana-2.0~\cite{google2026nanobanana2} & Closed & 0.965 & \underline{0.981} & 0.949 \\
JoyAI-Image~\cite{song2026joyai} & Apache-2.0 & 0.963 & 0.963 & 0.963 \\
\rowcolor{tablegray}
Seedream-5.0-Lite~\cite{seedream5lite} & Closed & 0.962 & 0.961 & 0.963 \\
Qwen-Image-2512~\cite{qwenimage2512} & Apache-2.0 & 0.961 & 0.956 & 0.965 \\
\rowcolor{tablegray}
GPT-Image-2~\cite{openai2026gptimage2} & Closed & 0.961 & 0.960 & 0.961 \\
\textcolor{boogured}{Boogu-Image-0.1-Turbo} & \textcolor{boogured}{Apache-2.0} & \textcolor{boogured}{0.961} & \textcolor{boogured}{0.944} & \textcolor{boogured}{0.977} \\
\textcolor{boogured}{Boogu-Image-0.1-Base} & \textcolor{boogured}{Apache-2.0} & \textcolor{boogured}{0.961} & \textcolor{boogured}{0.952} & \textcolor{boogured}{0.969} \\
Qwen-Image~\cite{qwen2025qwenimage} & Apache-2.0 & 0.945 & 0.943 & 0.946 \\
\rowcolor{tablegray}
Seedream-4.0~\cite{seedream2025seedream40nextgenerationmultimodal} & Closed & 0.924 & 0.921 & 0.926 \\
Z-Image-Turbo~\cite{zimage2025} & Apache-2.0 & 0.922 & 0.917 & 0.926 \\
\rowcolor{tablegray}
GPT-Image-1 [High]~\cite{openai2025gptimage1} & Closed & 0.788 & 0.956 & 0.619 \\
\bottomrule
\end{tabular}\par
}
\caption{\textbf{Our model series achieves top-tier performance among the evaluated models on LongText-Bench.} Models are sorted by average score. Gray shading denotes closed-source models, and our models are highlighted in \textcolor{boogured}{\textbf{red}}. Best scores are \textbf{bolded} and second-best scores are \underline{underlined}. Results for other methods are taken from \cite{glmimage2026,hidreamolimage}.} Actually, Base model outperforms Turbo in dense text rendering. While Turbo introduces noticeable visual artifacts in dense layouts, it still scores high on the LongText benchmark, as the metric primarily evaluates text accuracy for shorter sequences (<100 characters) rather than overall visual quality.
\label{tab:longtextbench}
\end{table}

\clearpage
\subsubsection{Benchmarks We Do Not Use}
\label{sec:drop_bench}
We observe that a significant gap still remains between some text-to-image benchmarks (\eg, GenEval~\cite{ghosh2023geneval} and DPG-Bench~\cite{hu2024ella}) and real human preferences, as discussed in Figure~\ref{fig:Public_benchmarks}. Consequently, we do not adopt these benchmarks for our primary evaluation. We nonetheless report the corresponding results as a supplementary reference for historical continuity (Table~\ref{tab:geneval_bench} for GenEval and Table~\ref{tab:dpg_bench} for DPG-Bench), and we suggest that their suitability as primary benchmarks going forward be carefully reconsidered.

\begin{table}[htb]
\centering
\footnotesize
\renewcommand{\arraystretch}{1.08}
\setlength{\tabcolsep}{2pt}

\begin{tabular}{@{} l l *{7}{c} @{}}
\toprule
\textbf{Model} & \textbf{License} & \textbf{Single Obj.} & \textbf{Two Obj.} & \textbf{Counting} & \textbf{Colors} & \textbf{Position} & \textbf{Attribute} & \textbf{Overall}$\uparrow$ \\
\midrule
HiDream-O1-Image~\cite{hidreamolimage} & Apache-2.0 & \textbf{1.00} & \textbf{0.99} & 0.79 & 0.89 & \textbf{0.93} & \underline{0.78} & \textbf{0.90} \\
\rowcolor{tablegray} GPT-Image-2~\cite{openai2026gptimage2} & Closed & \underline{0.99} & \underline{0.98} & 0.85 & \textbf{0.93} & 0.85 & 0.77 & \underline{0.89} \\
FLUX.2 [Dev] \cite{bfl2025flux2} & Non-Comm. & \textbf{1.00} & \textbf{0.99} & 0.79 & \textbf{0.93} & 0.73 & \underline{0.78} & 0.87 \\
Qwen-Image~\cite{qwen2025qwenimage} & Apache-2.0 & \underline{0.99} & 0.92 & \underline{0.89} & 0.88 & 0.76 & 0.77 & 0.87 \\
\textcolor{boogured}{{Boogu-Image-0.1-Base}} & \textcolor{boogured}{Apache-2.0} & \textcolor{boogured}{\underline{0.99}} & \textcolor{boogured}{0.95} & \textcolor{boogured}{0.80} & \textcolor{boogured}{0.84} & \textcolor{boogured}{0.85} & \textcolor{boogured}{0.68} & \textcolor{boogured}{0.85} \\
\textcolor{boogured}{{Boogu-Image-0.1-Turbo}} & \textcolor{boogured}{Apache-2.0} & \textcolor{boogured}{\textbf{1.00}} & \textcolor{boogured}{0.97} & \textcolor{boogured}{0.85} & \textcolor{boogured}{0.76} & \textcolor{boogured}{\underline{0.86}} & \textcolor{boogured}{0.60} & \textcolor{boogured}{0.84} \\
\textcolor{boogured}{{Boogu-Image-0.1-Turbo-Think.}} & \textcolor{boogured}{Apache-2.0} & \textcolor{boogured}{\textbf{1.00}} & \textcolor{boogured}{0.94} & \textcolor{boogured}{0.88} & \textcolor{boogured}{0.83} & \textcolor{boogured}{0.70} & \textcolor{boogured}{0.68} & \textcolor{boogured}{0.84} \\
\rowcolor{tablegray} Seedream-4.0~\cite{seedream2025seedream40nextgenerationmultimodal} & Closed & \textbf{1.00} & 0.92 & 0.71 & \textbf{0.93} & 0.78 & 0.68 & 0.84 \\
\rowcolor{tablegray} Seedream-3.0~\cite{gao2025seedream3} & Closed & \underline{0.99} & 0.96 & \textbf{0.91} & \textbf{0.93} & 0.47 & \textbf{0.80} & 0.84 \\
\rowcolor{tablegray} GPT-Image-1 [High]~\cite{openai2025gptimage1} & Closed & \underline{0.99} & 0.92 & 0.85 & \underline{0.92} & 0.75 & 0.61 & 0.84 \\
Z-Image~\cite{zimage2025} & Apache-2.0 & \textbf{1.00} & 0.94 & 0.78 & \textbf{0.93} & 0.62 & 0.77 & 0.84 \\
\rowcolor{tablegray} Nano-Banana-2.0~\cite{google2026nanobanana2} & Closed & \textbf{1.00} & 0.96 & 0.71 & 0.84 & \underline{0.86} & 0.65 & 0.83 \\
Z-Image-Turbo~\cite{zimage2025} & Apache-2.0 & \textbf{1.00} & 0.95 & 0.77 & 0.89 & 0.65 & 0.68 & 0.82 \\
FLUX.1 [Dev] \citep{flux2024blackforest} & Non-Comm. & 0.98 & 0.81 & 0.74 & 0.79 & 0.22 & 0.45 & 0.66 \\
\bottomrule
\end{tabular}
\caption{\textbf{Performance of the Boogu series and baseline models on GenEval.} Closed-source models are shaded in gray. Boogu series models are highlighted in \textcolor{boogured}{\textbf{red}}. Best scores in each column are \textbf{bolded} and second-best scores are \underline{underlined}. Models are sorted by overall score in descending order. Results for other methods are taken from~\cite{hidreamolimage}. {We report these results \underline{for reference only}, as GenEval exhibits a substantial gap from human preference.}}
\label{tab:geneval_bench}
\end{table}

\begin{table}[htb]
\centering
\footnotesize
\renewcommand{\arraystretch}{1.08}
\begin{tabularx}{\linewidth}{@{} l l *{6}{>{\centering\arraybackslash}X} @{}}
\toprule
\textbf{Model} & \textbf{License} & \textbf{Global} & \textbf{Entity} & \textbf{Attribute} & \textbf{Relation} & \textbf{Other} & \textbf{Overall}$\uparrow$  \\
\midrule
HiDream-O1-Image~\cite{hidreamolimage} & Apache-2.0 & \textbf{95.15} & 92.32 & \textbf{93.74} & 92.88 & 90.25 & \textbf{89.83} \\
\rowcolor{tablegray} Seedream-4.5~\cite{seedream4d5} & Closed & 89.24 & \textbf{94.30} & 92.14 & 92.23 & \underline{93.83} & \underline{88.63} \\
\rowcolor{tablegray} Seedream-4.0~\cite{seedream2025seedream40nextgenerationmultimodal} & Closed & 93.86 & 92.24 & 90.74 & \underline{93.87} & \textbf{94.16} & 88.54 \\
\textcolor{boogured}{{Boogu-Image-0.1-Turbo}} & \textcolor{boogured}{Apache-2.0} & \textcolor{boogured}{88.54} & \textcolor{boogured}{91.67} & \textcolor{boogured}{92.19} & \textcolor{boogured}{93.20} & \textcolor{boogured}{\underline{93.83}} & \textcolor{boogured}{88.35} \\
Qwen-Image~\cite{qwen2025qwenimage} & Apache-2.0 & 91.32 & 91.56 & 92.02 & \textbf{94.31} & 92.73 & 88.32 \\
\rowcolor{tablegray} Seedream-3.0~\cite{gao2025seedream3} & Closed & \underline{94.31} & \underline{92.65} & 91.36 & 92.78 & 88.24 & 88.27 \\
Z-Image~\cite{zimage2025} & Apache-2.0 & 93.39 & 91.22 & 93.16 & 92.22 & 91.52 & 88.14 \\
FLUX.2 [Dev] \cite{bfl2025flux2} & Non-Commercial & 92.20 & 91.36 & \underline{93.28} & 93.52 & 89.72 & 87.57 \\
\textcolor{boogured}{{Boogu-Image-0.1-Turbo-Thinking}} & \textcolor{boogured}{Apache-2.0} & \textcolor{boogured}{90.26} & \textcolor{boogured}{92.03} & \textcolor{boogured}{91.27} & \textcolor{boogured}{92.35} & \textcolor{boogured}{91.98} & \textcolor{boogured}{87.23} \\
Qwen-Image-2512~\cite{qwenimage2512} & Apache-2.0 & 89.04 & 91.91 & 92.39 & 90.85 & 93.07 & 87.20 \\
\textcolor{boogured}{{Boogu-Image-0.1-Base}} & \textcolor{boogured}{Apache-2.0} & \textcolor{boogured}{89.33} & \textcolor{boogured}{90.64} & \textcolor{boogured}{92.22} & \textcolor{boogured}{93.72} & \textcolor{boogured}{93.32} & \textcolor{boogured}{87.13} \\
\rowcolor{tablegray} Nano-Banana-2.0~\cite{google2026nanobanana2} & Closed & 85.17 & 92.55 & 91.16 & 90.45 & 91.08 & 86.90 \\
\rowcolor{tablegray} GPT-Image-2~\cite{openai2026gptimage2} & Closed & 87.27 & 91.91 & 90.85 & 91.59 & 91.58 & 85.98 \\
\rowcolor{tablegray} GPT-Image-1 [High]~\cite{openai2025gptimage1} & Closed & 88.89 & 88.94 & 89.84 & 92.63 & 90.96 & 85.15 \\
Z-Image-Turbo~\cite{zimage2025} & Apache-2.0 & 91.29 & 89.59 & 90.14 & 92.16 & 88.68 & 84.86 \\
GLM-Image~\cite{glmimage2026} & Apache-2.0 & 87.74 & 90.25 & 89.08 & 92.15 & 90.17 & 84.78 \\
FLUX.1 [Dev] \citep{flux2024blackforest} & Non-Commercial & 74.35 & 90.00 & 88.96 & 90.87 & 88.33 & 83.84 \\
\bottomrule
\end{tabularx}
\caption{\textbf{Performance of the Boogu series and baseline models on DPG-Bench.} Closed-source models are shaded in gray. Boogu series models are highlighted in \textcolor{boogured}{\textbf{red}}. Best scores in each column are \textbf{bolded} and second-best scores are \underline{underlined}. Models are sorted by overall score in descending order. Results for other methods are taken from \cite{hidreamolimage}. {We report these results \underline{for reference only}, as DPGBench exhibits a substantial gap from human preference.}
}
\label{tab:dpg_bench}
\end{table}

\clearpage

\subsubsection{Qualitative Comparison}
\label{sec:t2i_qual_cmp}
We provide qualitative comparison samples to offer readers a direct visual reference beyond aggregate benchmark scores. The testing samples are sourced from {\model} Arena, so that each compared model is represented by its strongest publicly accessible outputs under the deployed evaluation setting. This setup helps avoid underestimating baselines due to suboptimal local inference configurations, prompt formatting, or sampling choices. Across the selected examples, we compare Boogu-Image-0.1 with strong closed-source models (including Qwen-Image-Max-2025-12-30~\cite{qwenimage2512}, Qwen-Image-2.0-2026-03-03~\cite{zhao2026qwen2}, Seedream-5-Lite~\cite{seedream5lite}, Nano-Banana-Pro~\cite{google2025nanobananapro}, and GPT-Image-2~\cite{openai2026gptimage2}) and open-source models (Z-Image-Turbo~\cite{zimage2025} and HiDream-O1~\cite{hidreamolimage}), focusing on visual fidelity, prompt alignment, composition, and aesthetic quality. These examples complement the quantitative evaluations and illustrate how Boogu-Image-0.1 performs in user-facing comparisons.

\begin{figure}[htb]
  \centering
  \includegraphics[width=\linewidth]{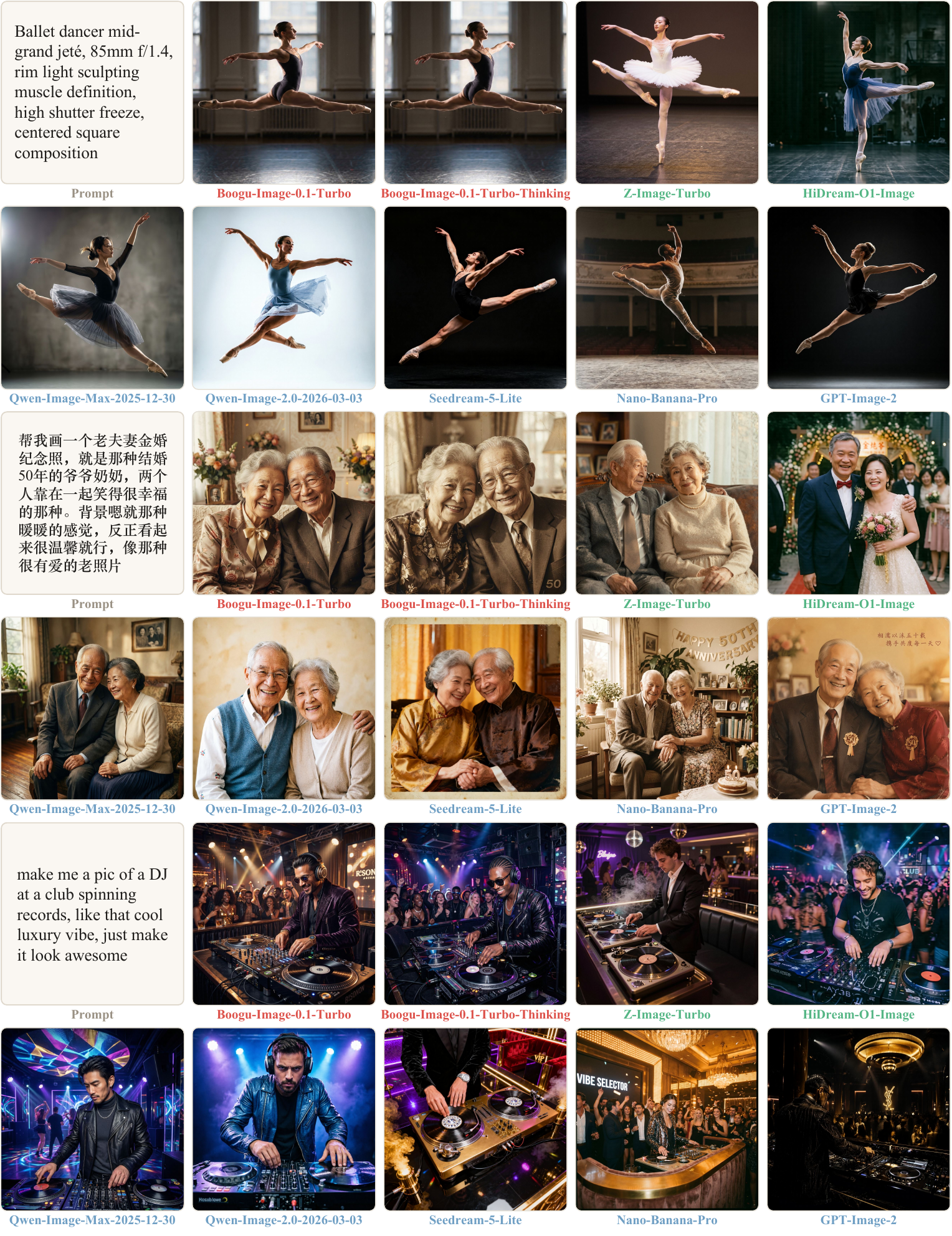}
\caption{
\textbf{Qualitative results from {\model} Arena on photorealistic and cinematic images.}
Our model can produce photorealistic and cinematic images with superior visual quality.
}  \label{fig:arena_qual_portrait}
\end{figure}

\begin{figure}[htb]
  \centering
  \includegraphics[width=\linewidth]{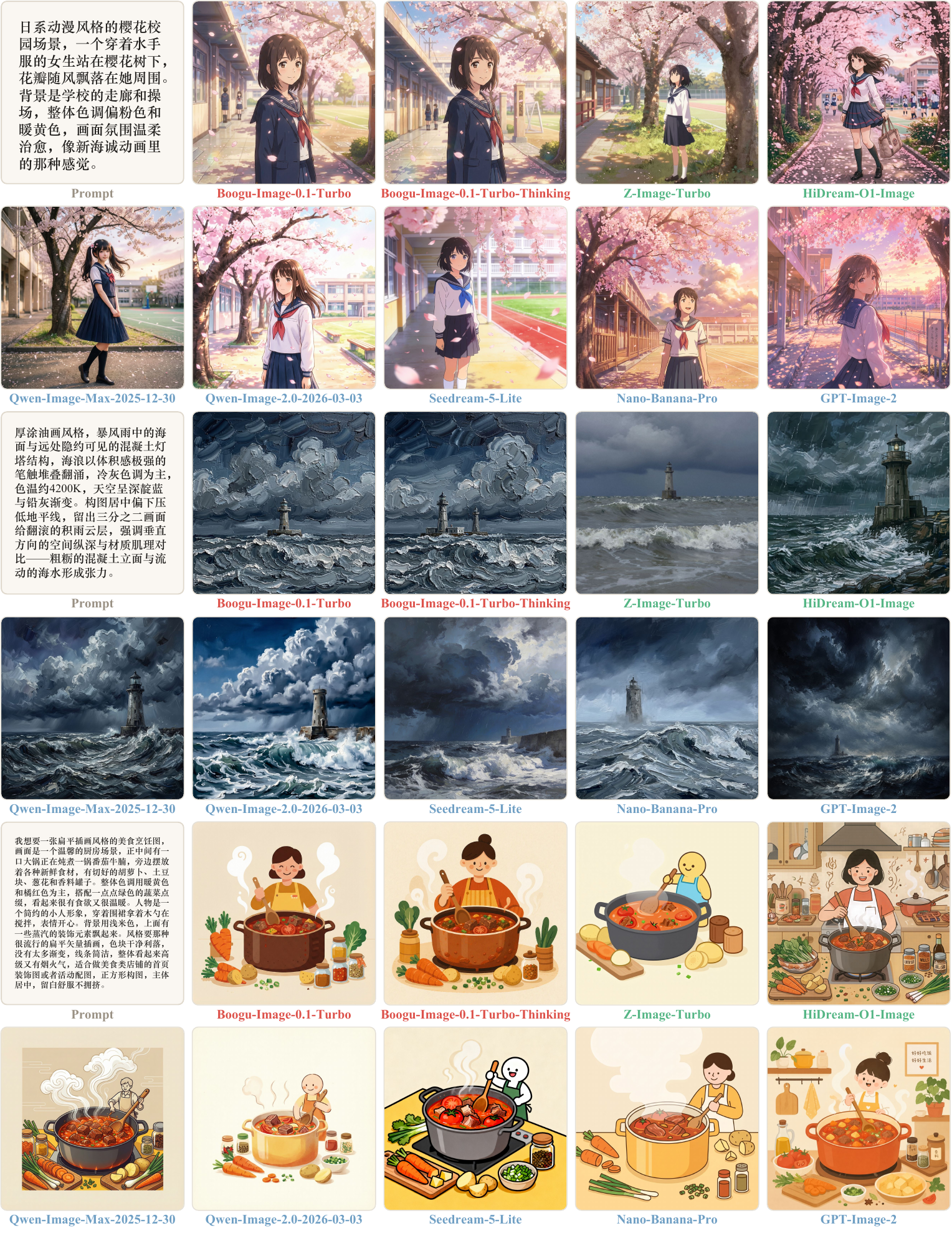}
\caption{
\textbf{Qualitative results from {\model} Arena on stylized art.}
Our model enables the generation of diverse artistic images with superior visual quality.
}  \label{fig:arena_qual_stylized_art}
\end{figure}

\begin{figure}[htb]
  \centering
  \includegraphics[width=0.95\linewidth]{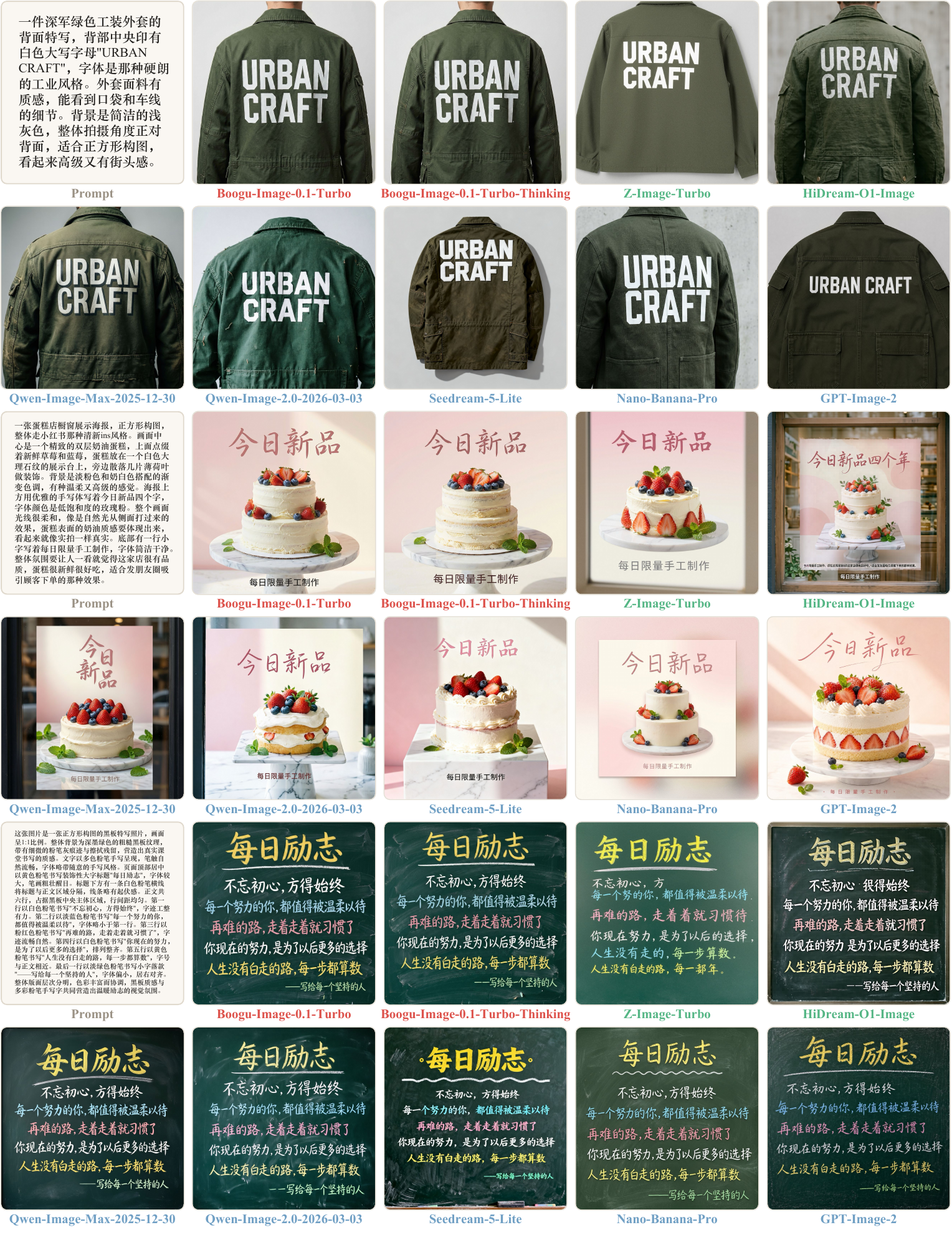}
\caption{
\textbf{Qualitative results from {\model} Arena on simple text rendering.}
Our \boogured{Boogu-Image-0.1-Turbo} can precisely renders short-to-medium text (up to 100 characters/words for Chinese/English) into the generated images. While most baseline models accurately render the requested text, our model achieves superior prompt-image alignment and higher visual quality compared to existing open-source counterparts. For dense text rendering (>100 characters/words), we highly recommend utilizing Boogu-Image-0.1-Base tailored for high-fidelity dense text rendering under 2K resolution (see Figure \ref{fig:qual_dense_text_1} \& \ref{fig:qual_dense_text_2}).
}  \label{fig:arena_qual_simple_text}
\end{figure}

Specifically, Figures~\ref{fig:arena_qual_portrait}--\ref{fig:arena_qual_simple_text} present qualitative comparisons for the three categories in {\model} Arena, namely photorealistic and cinematic imagery, stylized art, and text rendering. These visual results demonstrate that Boogu-Image-0.1-Turbo achieves highly competitive performance in prompt alignment, visual quality, and text-rendering fidelity. Notably, it outperforms existing open-source alternatives and effectively narrows the performance gap toward top-tier closed-source models (\eg, Nano-Banana-Pro~\cite{google2025nanobananapro} and GPT-Image-2~\cite{openai2026gptimage2}), as validated by the human Elo ratings (Figure~\ref{fig:arena_t2i_elo_score}).

\begin{figure}[htb]
  \centering
  \includegraphics[width=.90\linewidth]{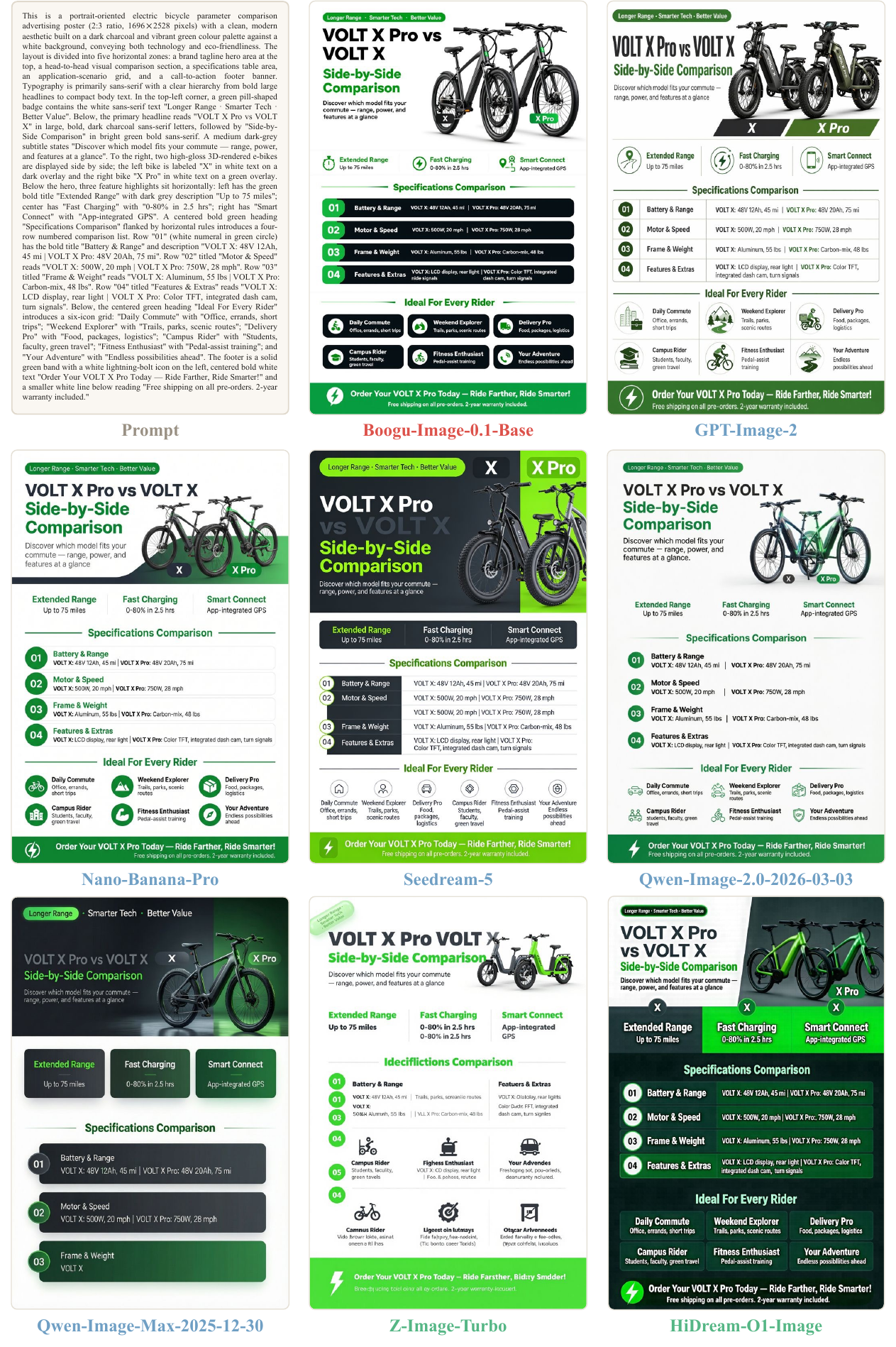}
  \vspace{-3mm}
\caption{
\textbf{Qualitative comparison between our \boogured{Boogu-Image-0.1-Base} and existing SOTA models on dense text rendering under 2K resolution.}
}  \label{fig:qual_dense_text_1}
\end{figure}

\begin{figure}[htb]
  \centering
  \includegraphics[width=0.97\linewidth]{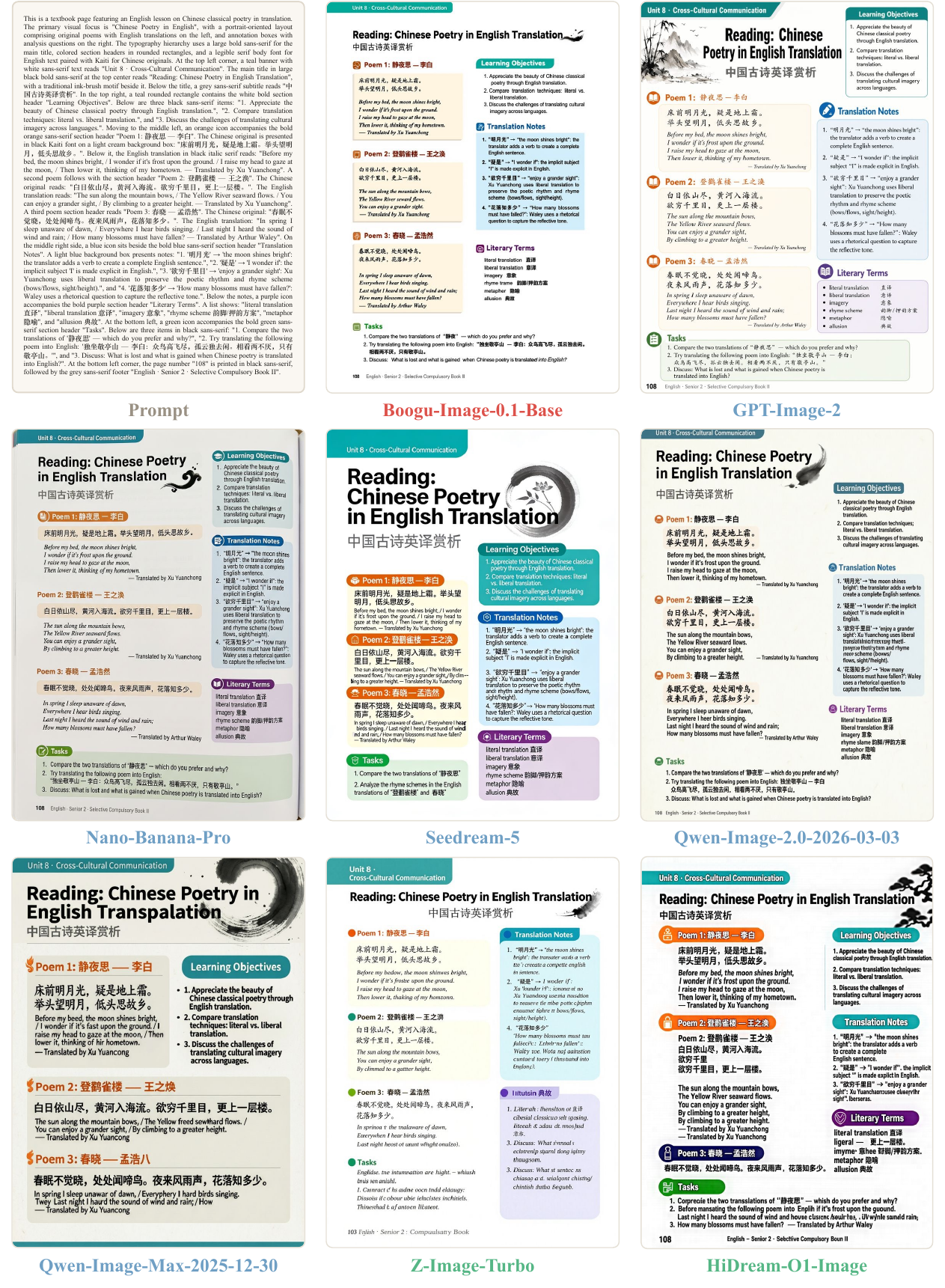}
\caption{
\textbf{Qualitative comparison between our \boogured{Boogu-Image-0.1-Base} and existing SOTA models on dense text rendering under 2K resolution.}
}  
\label{fig:qual_dense_text_2}
\end{figure}

In terms of text rendering (Figure~\ref{fig:arena_qual_simple_text}), Boogu-Image-0.1-Turbo reliably supports precise character-level rendering for short-to-medium text (up to 100 characters for Chinese / 100 words for English). For more challenging scenarios involving ultra-dense text (exceeding 100 characters or words), we recommend deploying Boogu-Image-0.1-Base, which is explicitly optimized for dense text rendering at high (2K) resolution. As shown in Figures~\ref{fig:qual_dense_text_1} and~\ref{fig:qual_dense_text_2}, Boogu-Image-0.1-Base remains stable on dense text rendering even at minuscule font sizes, while maintaining rigorous character accuracy and strong aesthetic appeal for practical applications.

\definecolor{ourcolor}{RGB}{230,230,230}  %

\clearpage

\subsection{Image-to-Image Generation}
\label{sec:i2i_gen}
\subsubsection{ImgEdit Benchmark} 
\label{sec:imgedit_bench}
ImgEdit-Bench~\cite{ye2026imgedit} is a widely adopted framework designed to assess instruction-guided image editing across a diverse range of tasks. We evaluate our Image-to-Image models on ImgEdit-Bench to enable direct comparison with existing methods. As shown in Table~\ref{tab:imgedit_bench}, Boogu-Image-0.1-Edit-Thinking attains the best overall score (4.64) among all evaluated models, surpassing strong open-source editors such as JoyAI-Image-Edit (4.57) and FireRed-Image-Edit (4.56), as well as leading closed-source systems including Seedream-5.0-Lite (4.42) and Nano-Banana-Pro (4.37). At the sub-task level, our model ranks first on Remove (4.85), Hybrid (4.26), and Action (4.83), and remains highly competitive across the remaining categories, indicating balanced editing capability rather than strength concentrated in a few task types. Notably, the Thinking variant yields consistent gains over the Base variant (4.64 vs.\ 4.51 overall), with the largest margin on the challenging Extract task (4.32 vs.\ 3.69); this suggests that explicit reasoning helps the model parse and localize complex editing instructions before generation. We further note that Extract is difficult for essentially all models—several strong closed-source systems score below 3.5 (\eg, Seedream-4.5 at 2.97)—which likely reflects both the intrinsic difficulty of the task and higher variance in VLM-based scoring for it.

\noindent\textbf{Discussion and Limitations.} Although Boogu-Image-0.1-Edit-Thinking attains the highest overall score on ImgEdit-Bench, we caution against over-interpreting this ranking. In our own human evaluations, we find that Nano-Banana-Pro~\cite{google2025nanobananapro} outperforms our models on a range of real editing scenarios, despite scoring lower on this benchmark (4.37 vs.\ 4.64 overall). This inversion highlights a broader issue: ImgEdit-Bench relies primarily on Vision-Language Models (VLMs) as automated evaluators, and current VLMs remain limited in fine-grained visual understanding, making their scores an imperfect proxy for human preference. As a result, the benchmark tends to compress the true quality gap between models and can reward outputs that satisfy instruction-level checks while overlooking subtler perceptual defects. 

{Moreover, its task taxonomy is relatively coarse: the nine covered categories fall short of capturing the full diversity of real-world editing operations—such as multi-step and multi-region edits, precise text editing, subject-consistent and reference-guided generation, and fine-grained local adjustments—so a high aggregate score does not necessarily imply strong performance on the broader range of edits encountered in practice.} We therefore report ImgEdit-Bench results primarily as a \underline{supplementary reference for historical continuity}, and we suggest that its suitability as a primary benchmark going forward be carefully reconsidered. For assessments that better reflect end-user experience, we place greater weight on human evaluations and qualitative comparisons.

\begin{table*}[htb]
\setlength{\tabcolsep}{4pt}
\renewcommand{\arraystretch}{1.08}
{\centering
\resizebox{\textwidth}{!}{
\begin{tabular}{l l c c c c c c c c c c}
\toprule
\textbf{Model} & \textbf{License} & \textbf{Add} & \textbf{Adjust} & \textbf{Extract} & \textbf{Replace} & \textbf{Remove} & \textbf{BG} & \textbf{Style} & \textbf{Hybrid} & \textbf{Action} & \textbf{Overall} $\uparrow$\\
\midrule
\textbf{\textcolor{boogured}{Boogu-Image-0.1-Edit-Thinking}} & \textcolor{boogured}{Apache-2.0} & \textcolor{boogured}{4.59} & \textcolor{boogured}{4.64} & \textcolor{boogured}{\underline{4.32}} & \textcolor{boogured}{4.69} & \textcolor{boogured}{\textbf{4.85}} & \textcolor{boogured}{\underline{4.60}} & \textcolor{boogured}{\underline{4.94}} & \textcolor{boogured}{\textbf{4.26}} & \textcolor{boogured}{\underline{4.83}} & \textcolor{boogured}{\textbf{4.64}} \\
JoyAI-Image-Edit~\cite{song2026joyai} & Apache-2.0 & 4.63 & 4.52 & \underline{4.32} & 4.71 & \underline{4.76} & 4.53 & 4.88 & 4.09 & 4.69 & \underline{4.57} \\
FireRed-Image-Edit~\cite{team2026firered} & Apache-2.0 & 4.55 & \underline{4.66} & \textbf{4.34} & \underline{4.75} & 4.58 & 4.45 & \textbf{4.97} & 4.07 & 4.71 & 4.56 \\
\textbf{\textcolor{boogured}{Boogu-Image-0.1-Edit}} & \textcolor{boogured}{Apache-2.0} & \textcolor{boogured}{\underline{4.71}} & \textcolor{boogured}{4.50} & \textcolor{boogured}{3.69} & \textcolor{boogured}{4.65} & \textcolor{boogured}{4.75} & \textcolor{boogured}{4.44} & \textcolor{boogured}{\underline{4.94}} & \textcolor{boogured}{4.04} & \textcolor{boogured}{\textbf{4.90}} & \textcolor{boogured}{4.51} \\
Qwen-Image-Edit-2511~\cite{qwenimageedit2511} & Apache-2.0 & 4.54 & 4.57 & 4.13 & 4.70 & 4.46 & 4.36 & 4.89 & \underline{4.16} & 4.81 & 4.51 \\
LongCat-Image-Edit~\cite{team2025longcat} & MIT & 4.44 & 4.53 & 3.83 & \textbf{4.80} & 4.60 & 4.33 & 4.92 & 3.75 & 4.82 & 4.45 \\
\rowcolor{tablegray} Seedream-5.0-Lite~\cite{seedream5lite} & Closed & \textbf{4.93} & \textbf{4.69} & 3.01 & 4.41 & 4.45 & \textbf{4.65} & 4.93 & 3.91 & 4.82 & 4.42 \\
\rowcolor{tablegray} Nano-Banana-Pro~\cite{google2025nanobananapro} & Closed & 4.44 & 4.62 & 3.42 & 4.60 & 4.63 & 4.32 & \textbf{4.97} & 3.64 & 4.69 & 4.37 \\
FLUX.2-Dev~\cite{blackforestlabs2025flux2} & Non-Commercial & 4.50 & 4.18 & 3.83 & 4.65 & 4.65 & 4.31 & 4.88 & 3.46 & 4.70 & 4.35 \\
\rowcolor{tablegray} Seedream-4.5~\cite{seedream4d5} & Closed & 4.57 & 4.65 & 2.97 & 4.66 & 4.46 & 4.37 & 4.92 & 3.71 & 4.56 & 4.32 \\
Qwen-Image-Edit-2509~\cite{qwenimageedit2509} & Apache-2.0 & 4.34 & 4.27 & 3.42 & 4.73 & 4.36 & 4.37 & 4.91 & 3.56 & 4.80 & 4.31 \\
\rowcolor{tablegray} Seedream-4.0~\cite{seedream2025seedream40nextgenerationmultimodal} & Closed & 4.33 & 4.38 & 3.89 & 4.65 & 4.57 & 4.35 & 4.22 & 3.71 & 4.61 & 4.30 \\
\rowcolor{tablegray} Nano-Banana~\cite{google2025nanobananapro} & Closed & 4.62 & 4.41 & 3.68 & 4.34 & 4.39 & 4.40 & 4.18 & 3.72 & \underline{4.83} & 4.29 \\
Step1X-Edit-v1.2~\cite{liu2025step1x} & Apache-2.0 & 3.91 & 4.04 & 2.68 & 4.48 & 4.26 & 3.90 & 4.82 & 3.23 & 4.22 & 3.95 \\
\bottomrule
\end{tabular}
}\par
}
\caption{\textbf{Boogu-Image-0.1-Edit-Thinking achieves the best overall performance on ImgEdit-Bench.} Models are sorted by overall score. \textbf{BG} denotes the background-change task. Gray shading denotes closed-source models, and our models are highlighted in \textcolor{boogured}{\textbf{red}}. Best scores are \textbf{bolded} and second-best scores are \underline{underlined}. Results for other methods are taken from \cite{song2026joyai}. Note that results are presented \underline{for reference only}, as VLM evaluations on ImgEdit do not fully align with human preferences. }
\label{tab:imgedit_bench}
\end{table*}

\subsubsection{Qualitative Comparison}
\label{sec:i2i_qual_cmp}
\begin{figure}[htb]
  \centering
  \includegraphics[width=\linewidth]{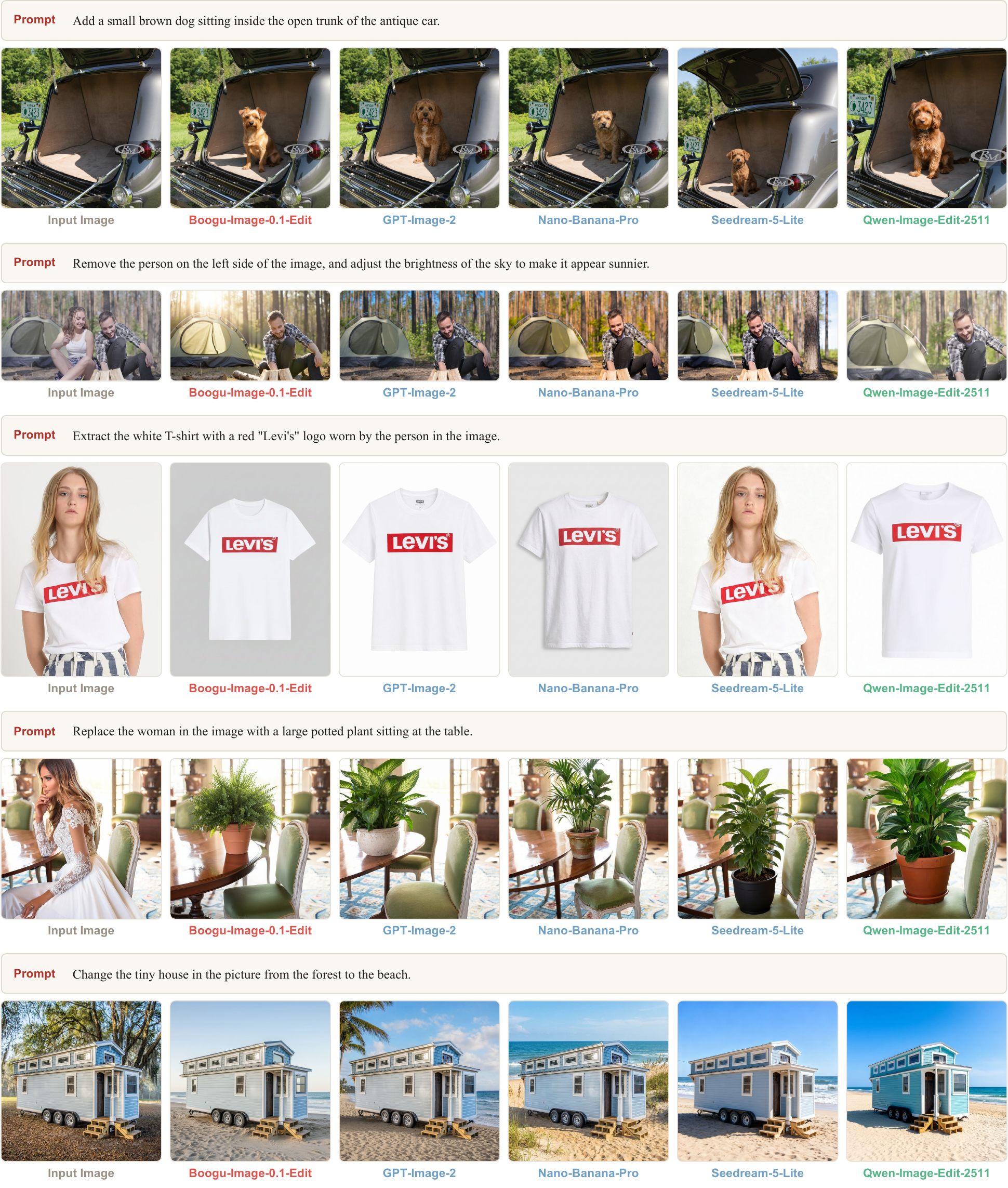}
\caption{
\textbf{Qualitative comparison on ImgEdit-Bench. }
{Boogu-Image-0.1-Edit} can achieve top-tier performance in diverse editing tasks, including but not limited to addition, removal, extraction, and replacement.}  \label{fig:i2i_qual_imgedit}
\end{figure}

\begin{figure}[htb]
  \centering
  \includegraphics[width=\linewidth]{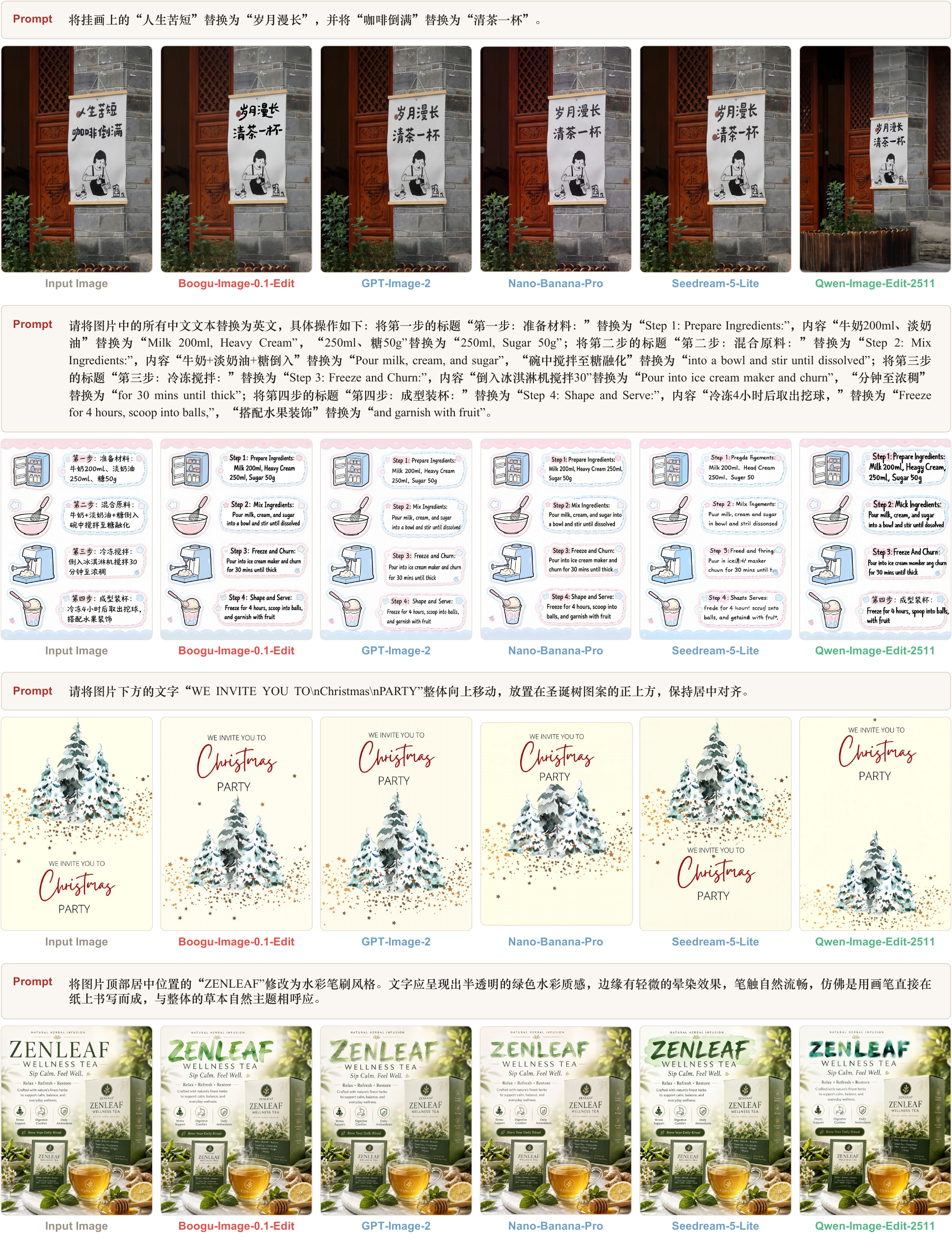}
\caption{
\textbf{Qualitative comparison on scene-text editing. }
{Boogu-Image-0.1-Edit} enables diverse scene-text editing tasks, including but not limited to text replacement, text translation, text re-organization, and text stylization.
}  \label{fig:i2i_qual_text}
\end{figure}

\begin{figure}[htb]
  \centering
  \includegraphics[width=\linewidth]{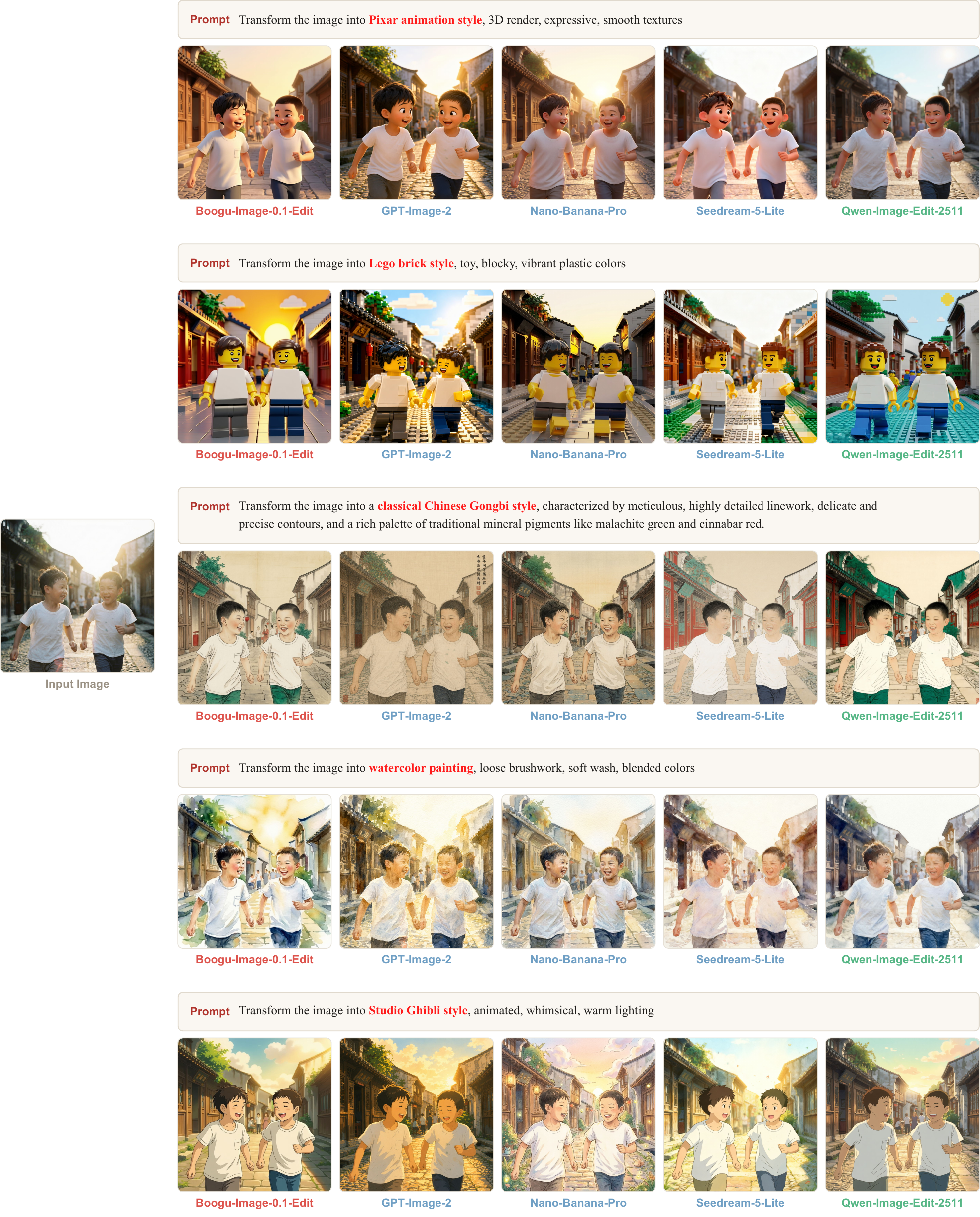}
\caption{
\textbf{Qualitative comparison on stylization. }
{Boogu-Image-0.1-Edit} supports diverse image-stylization edits.
}  \label{fig:i2i_qual_style}
\end{figure}

We provide qualitative comparison samples to offer readers a direct visual reference that complements the aggregate benchmark scores. The evaluation samples are drawn from the public ImgEdit-Bench~\cite{ye2026imgedit} as well as our in-house image-to-image benchmark. Similar to the text-to-image evaluation in Section~\ref{sec:t2i_evaluation}, the selected samples span three aspects: photorealistic editing, precise scene-text editing from our in-house benchmark, and diverse stylization. We compare Boogu-Image-0.1-Edit against several state-of-the-art image-to-image models, including three closed-source systems—GPT-Image-2~\cite{openai2026gptimage2}, Nano-Banana-Pro~\cite{google2025nanobananapro}, and Seedream-5.0-Lite~\cite{seedream5lite}—together with the open-source Qwen-Image-Edit-2511~\cite{qwenimageedit2511}. Figures~\ref{fig:i2i_qual_imgedit}--\ref{fig:i2i_qual_style} present the qualitative comparisons for these three aspects, respectively. Inherited from our powerful T2I base model Boogu-Image-0.1-Base, Boogu-Image-0.1-Edit achieves highly competitive image-to-image performance among the compared models, showing particular strength in instruction alignment, photorealism, precise scene-text modification, and artistic stylization.

\clearpage

\clearpage

\section{Experimental Analysis and Discussion}
\label{sec:Findings}

In this section, we share several key insights and observations encountered during our experiments. While these findings may not seem entirely ``novel'', they have clearly not received sufficient attention from the research community, as evidenced by recent literature. Therefore, we wish to re-emphasize the factors that have both helped and hindered our experimental process.

\subsection{Boosting Generation via Better Understanding}
\label{sec:boo_g_v_u}

The optimization objectives of Boogu are quality and efficiency, where quality is determined by the alignment between the user's intent and the generated image, and efficiency is measured by the computational cost incurred to reach a satisfactory output. Our design is guided by several empirical observations accumulated during the development and deployment of large-scale text-to-image (T2I) systems:
\begin{itemize}
    \item Many users issue very simple or underspecified instructions, such as a few keywords or a vague description, that nonetheless require non-trivial reasoning about context, common sense, or implicit visual conventions to be correctly interpreted and turned into a concrete generation target.
    \item Different tasks demand vastly different levels of generation capability: rendering a single isolated object is fundamentally easier than composing a multi-subject scene with precise spatial relations or stylistic constraints. Always invoking the strongest model is therefore computationally wasteful and offers diminishing returns on easy requests.
    \item Even the strongest Vision-Language Models (VLMs) suffer from inherent limitations on certain fine-grained capabilities, such as counting, precise spatial reasoning, and attribute binding, and relying on a single monolithic model to handle every aspect of understanding tends to inherit these weaknesses wholesale. Such limitations are further amplified when the system prompt becomes excessively long, forcing the model to attend to a large number of instructions at once and to lose focus on the most relevant ones for a given input.
\end{itemize}

To address these challenges in a unified manner, Boogu places \emph{understanding} at the center of its design and decomposes it along three complementary dimensions. Each targets a distinct source of error in current text-to-image pipelines, and together they form the backbone of our quality–efficiency trade-off:
\begin{itemize}
\item \textbf{Understanding user intent.} The first bottleneck of generation quality lies in interpreting what the user truly wants, which spans two complementary regimes. (1) When a prompt is short yet precise, its intended semantics must be understood and faithfully preserved rather than over-rewritten or diluted, which calls for a \underline{sufficiently strong text encoder} that can accurately capture the intended meaning. (2) When a prompt is vague or verbose, a stronger \underline{agentic prompt rewriter}, built on \underline{better rewriting principles} and a \underline{more capable VLM backbone}, is needed to translate it into a faithful, generator-ready form while reasoning about the underspecified details. Because both the text encoder and the rewriter operate upstream of the generator, improvements here propagate to every downstream module.
\item \textbf{Understanding the training images.} On the data side, the captioning strategy fundamentally determines what the model can and cannot learn: a generator can only acquire the concepts, attributes, and compositional patterns that its captions actually describe. \underline{Better image–caption alignment} covering not only salient objects but also style, spatial layout, counts, and fine-grained attributes is therefore essential to inject the desired capabilities during training, and to prevent systematic blind spots from being baked into the model weights.
\item \textbf{Understanding the task complexity.} Finally, on the inference side, Boogu explicitly estimates the \underline{difficulty of each request} before deciding how to serve it. This complexity-aware routing directs simple requests to lightweight models and reserves the strongest generators for genuinely hard cases requiring deep reasoning or high-fidelity composition, yielding a principled trade-off that improves average quality while keeping the compute budget under control.

\end{itemize}

\clearpage

\subsubsection{Instruction Encoder: The Sensor of Text-to-Image Models}
\label{sec:ins_encoder}
We observe that many strong open-source text-to-image models are limited not by their generative components, but by an inaccurate understanding of the text description. Every text-to-image model relies on an ``instruction encoder'', albeit in different forms, and we regard this encoder as analogous to a sensor: it defines the upper bound on the information available to the rest of the system. A weak encoder may discard important information before it interacts with image tokens, and because the instruction encoder is typically frozen during training, this limitation persists across both training and inference. In such cases, additional inference-time computation offers limited benefit, since no downstream process can recover information that was lost at the input. We therefore argue that a stronger encoder provides more accurate text understanding and higher-quality representations, and consequently yields better text-to-image generation. This view is consistent with several prior works that either report a similar pattern or adopt a stronger VLM as a default design choice, including Qwen-Image 2.0~\cite{qwen2025qwenimage} and Ideogram 4.0~\cite{ideogram-4-2026}.

Table~\ref{tab:vlm_scaling} reports GenEval~\cite{ghosh2023geneval} scores as a function of encoder scale. Holding the training data, hyperparameters, and DiT architecture fixed—aside from minor adjustments required for tensor-dimension alignment—we find that \textit{scaling the frozen instruction encoder alone yields consistent improvements in generation quality.} This indicates that the encoder's capacity to comprehend and represent instructions is a fundamental determinant of text-to-image performance: a more capable encoder translates directly into better generation. We emphasize that these gains do not come from the parameter count itself, but from the encoder's underlying ability to understand and encode instructions.

Owing to resource constraints, we did not evaluate larger encoders; however, within the range we examined, we observed no saturation in performance as the encoder scaled. While a stronger encoder is expected to further improve results, we also prioritize accessibility for open-source users, and therefore adopt Qwen3-VL-8B~\cite{bai2025qwen3vl} as the instruction encoder—a choice that balances generation quality against deployment cost.

\begin{figure}[h]
  \centering
  \includegraphics[width=1.0\linewidth]{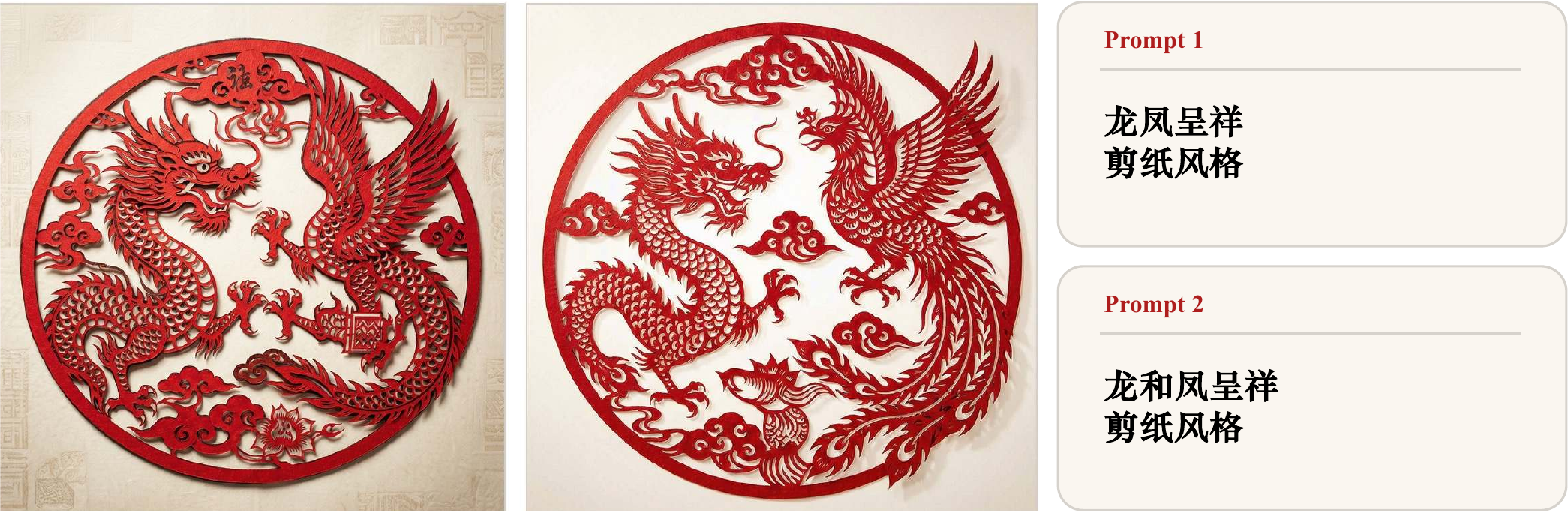}
\caption{\textbf{A single character changes everything.} The two prompts express the same intent—a dragon and a phoenix in paper-cut style—and differ only by the conjunction ``and'': ``the dragon phoenix brings prosperity'' (Prompt 1) versus ``the dragon and phoenix bring prosperity'' (Prompt 2). This single extra character breaks the fixed four-character idiom that the model has learned as a unit, producing a markedly different image despite the unchanged meaning.}\label{fig:figure_prompts}
\end{figure}

\begin{table}[h]
\centering
\begin{tabular}{lccc}
\toprule
Instruction Encoder\cite{yang2025qwen3} & Qwen3-1.7B & Qwen3-4B & Qwen3-14B \\
GenEval Score $\uparrow$ & 0.6034 & 0.6251 & \textbf{0.6477} \\
\bottomrule
\end{tabular}
\caption{\textbf{Effect of instruction-encoder scale on GenEval.} We vary only the instruction encoder~\cite{yang2025qwen3} while holding the DiT fixed at 1B parameters (with all other settings unchanged). GenEval scores improve monotonically as the encoder scales from 1.7B to 14B, with no saturation observed. {These are \underline{intermediate results} from models that are not yet fully trained, reported to show the relative trend across encoder scales rather than final performance.}}
\label{tab:vlm_scaling}
\end{table}

\clearpage

\subsubsection{Prompt Rewriter: A Translator, Not an Enhancer} 
\label{sec:pmpt_rewriter}
In real-world applications, user prompts are frequently ambiguous, occasionally unsafe, and often require reasoning to be correctly interpreted—precisely the challenges that photographers, designers, and painters confront in their daily work. Without a strong translator that aligns user intent with the generated image, users may remain dissatisfied with the result even when the image exhibits high-quality detail and aesthetic appeal. A capable prompt rewriter is therefore essential to bridge the gap between underspecified user input and the structured, unambiguous conditioning that the generation model expects.

Employing a prompt rewriter has become a common practice among open-source models~\cite{qwen2025qwenimage,cao2025hunyuanimage,liu2026ernie}. However, existing approaches exhibit several limitations. First, many rewriters produce excessively long outputs, introducing redundant or hallucinated details that drift from the user's original intent and inflate inference cost. Second, they typically operate in a single forward pass, without the capacity to reason over ambiguous or compositional instructions before committing to a rewrite. 

Motivated by these limitations, we adopt an \emph{agentic prompt rewriter} that treats rewriting as a reasoning process rather than a single-shot transformation. Rather than expanding a prompt indiscriminately, our rewriter interprets user intent, resolves ambiguity, and produces a faithful, appropriately scoped instruction for the generation model—acting as a translator of intent rather than an enhancer of content. Notably, recently several concurrent work also proposes agentic prompt rewriter~\cite{huang2026ape,liu2026ernie}.

\begin{figure}[h]
  \centering
  \includegraphics[width=1.0\linewidth]{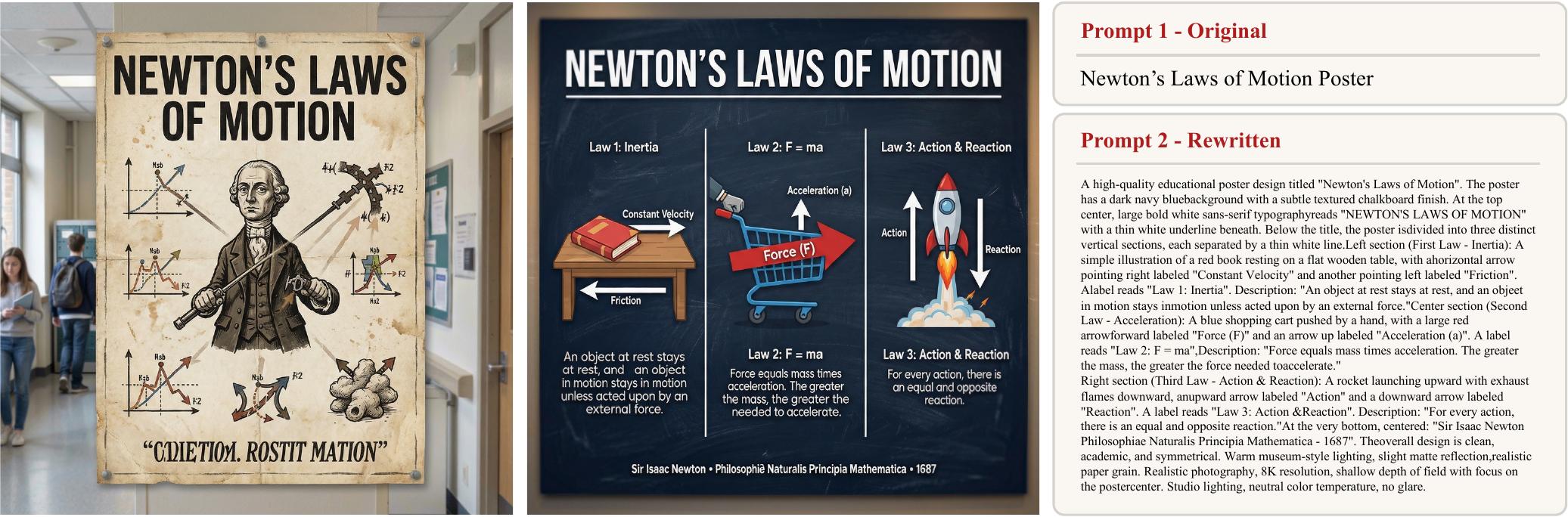}
\caption{\textbf{Effect of automatic prompt rewriting on text-to-image generation.} The left two columns show the images produced from the original prompt (left) and its rewritten version (center); the right column lists the two prompts. Prompt 1 is a terse user request, ``Newton's Laws of Motion Poster'', which leaves layout, style, color, content, and typography entirely unspecified. Prompt 2 is the output of a prompt rewriter that expands the same intent into a detailed, structured specification: an explicit title and underline, a dark navy chalkboard background, three vertical sections with named laws, concrete illustrations (a book, a shopping cart, a launching rocket), labeled force and motion arrows, per-section descriptions, an attribution footer, and photographic rendering directives. The rewritten prompt removes ambiguity that the model would otherwise resolve arbitrarily, yielding a far more coherent, on-topic, and publication-quality poster. This illustrates that prompt rewriting acts as a bridge between a user's high-level intent and the dense, descriptive input on which diffusion models \cite{peebles2023dit} perform best.}
\label{fig:figure_prompts_rewriter}
\end{figure}

\begin{figure}[!h]
    \centering
    \begin{subfigure}[t]{0.45\linewidth}
        \includegraphics[width=\linewidth]{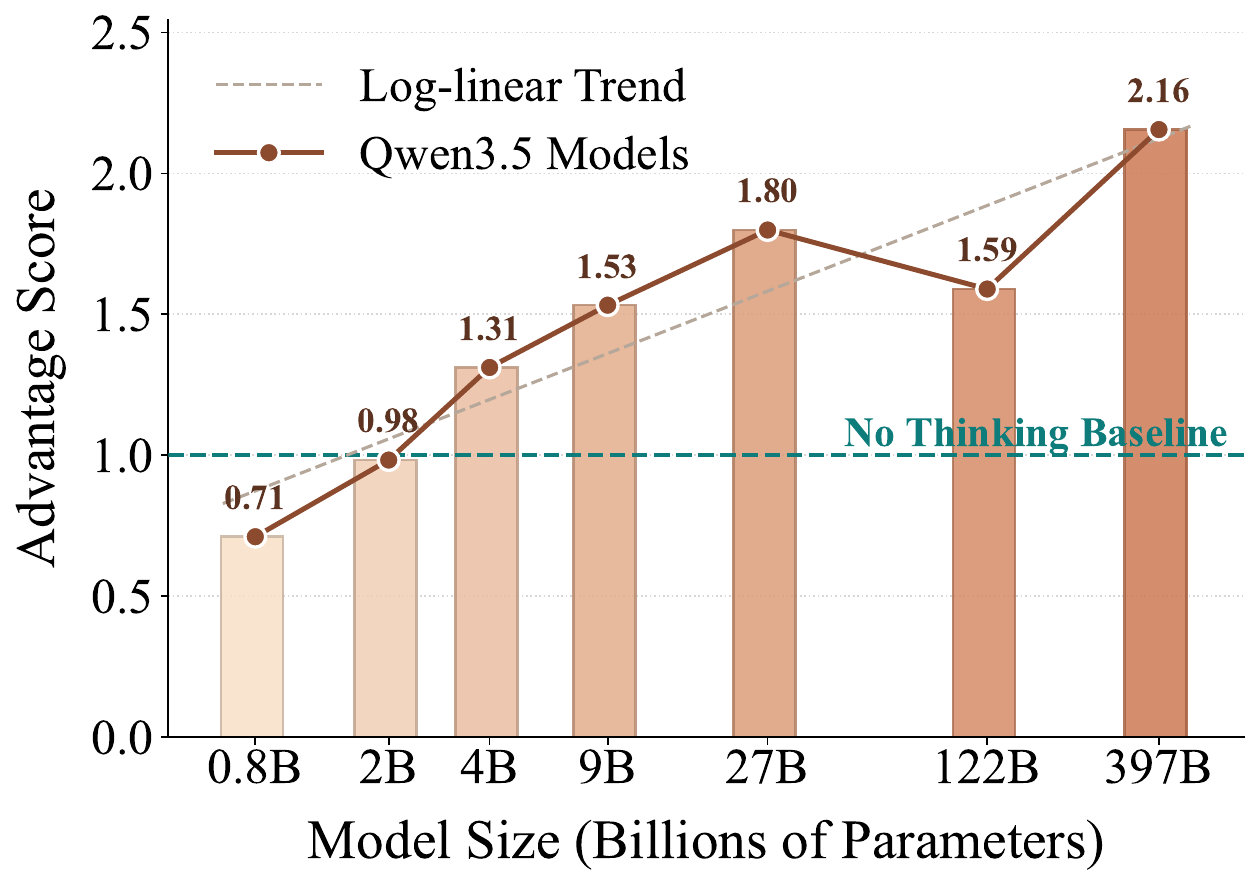}
        \caption{Ablation on Scaling Rewriter}
        \label{fig:pe_analysis_a}
    \end{subfigure}
    \hfill
    \begin{subfigure}[t]{0.49\linewidth}
        \includegraphics[width=\linewidth]{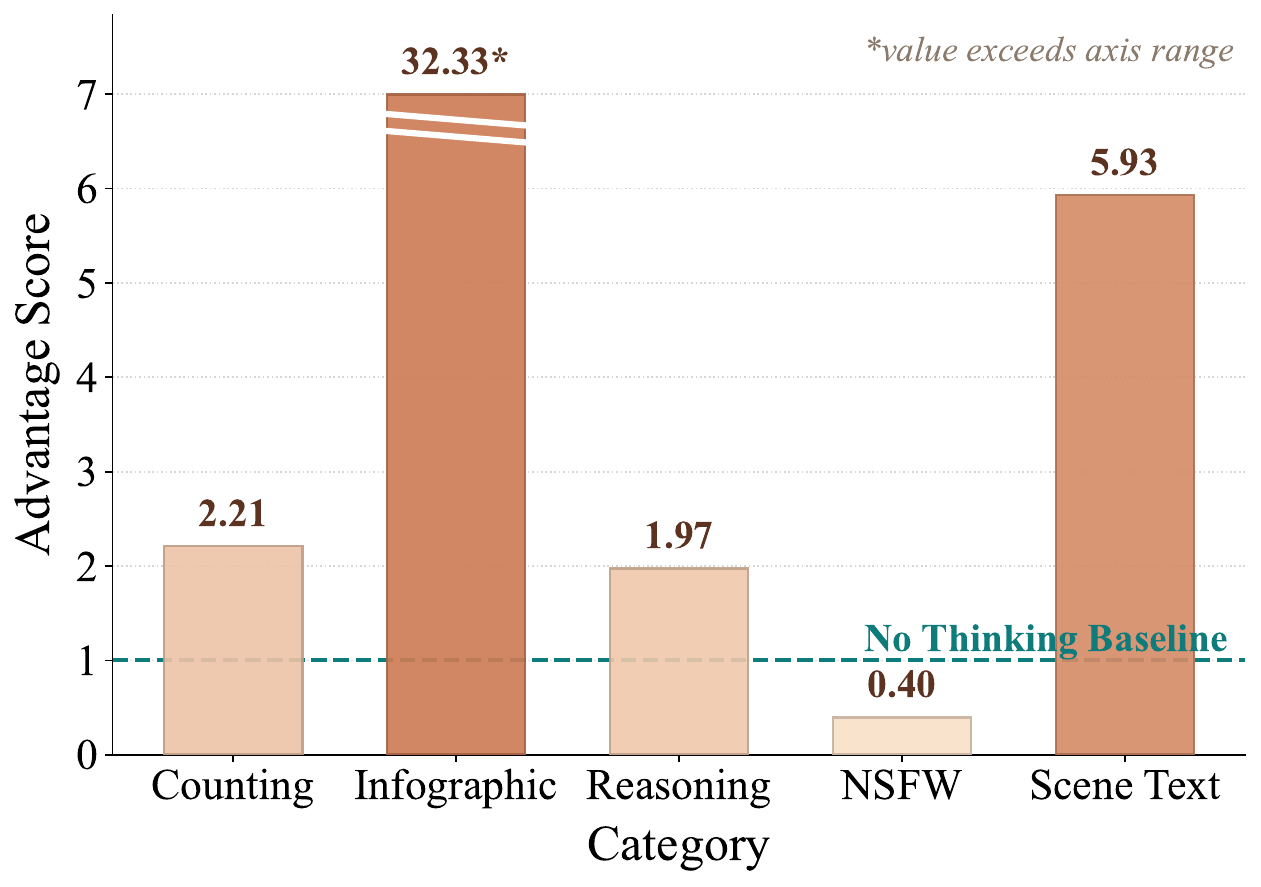}
        \caption{Ablation on Skills}
        \label{fig:pe_analysis_b}
    \end{subfigure}
    \caption{\textbf{Both scaling up the thinking rewriter and introducing capability-specific PE skills effectively improve prompt rewriting quality.}
    \textbf{(a)} Effect of different thinking models on rewriting. Compared with the no-thinking baseline, larger thinking models generally achieve higher advantage scores, indicating that stronger world knowledge and reasoning improve prompt rewriting quality. Among them, 122B and 397B are MoE models; the drop of 122B relative to 27B may be attributed to the MoE architecture or routing behavior, yet the overall trend remains that stronger models yield larger rewriting gains.
    \textbf{(b)} Effect of dimension-specific skills on the corresponding capabilities. After introducing dedicated skills for \textit{Counting}, \textit{Infographic}, \textit{Reasoning}, \textit{NSFW}, and \textit{Scene-Text}, the advantage scores of the corresponding dimensions all change, with \textit{Infographic} and \textit{Scene-Text} showing the most notable gains, demonstrating that capability-oriented skill design effectively enhances the prompt rewriter in target scenarios. The \textit{NSFW}  score is below 1 because unsafe prompts are intentionally sanitized, which may reduce literal prompt satisfaction while improving \textit{NSFW} alignment.
    The advantage score is defined as
    $S_{\mathrm{adv}} = \left( N_{\mathrm{win}} + N_{\mathrm{tie}}^{\mathrm{good}} + N_{\mathrm{tie}}^{\mathrm{bad}} \right) / \left( N_{\mathrm{lose}} + N_{\mathrm{tie}}^{\mathrm{good}} + N_{\mathrm{tie}}^{\mathrm{bad}} \right)$,
    where $S_{\mathrm{adv}} > 1$ indicates an advantage over the baseline.}
    \label{fig:pe_analysis}
\end{figure}

Figure~\ref{fig:figure_prompts_rewriter} shows an example that the prompt rewriter significantly boost overall performance. We utilize a reasoning-capable prompt rewriter to address several fundamental difficulties in text-to-image generation that the generation model alone struggles to resolve, as shown in Figure~\ref{fig:pe_analysis_b}.
\begin{itemize}
    \item \textbf{Reasoning.} Prompts that specify a target indirectly—\eg, ``the animal that lays eggs and produces milk'' or ``draw the result of $7 \times 8$''—require multi-step inference before a concrete visual subject can be determined. The rewriter resolves such implicit descriptions into explicit, renderable instructions.
    \item \textbf{Counting.} Generation models are notoriously unreliable at producing a precise number of objects. By explicitly enumerating instances in the rewritten prompt (\eg, expanding ``five apples'' into a structured, position-aware description), the rewriter alleviates miscounting and improves numerical fidelity.
    \item \textbf{NSFW prompts.} Unsafe or policy-violating requests can be identified at the rewriting stage and sanitized before reaching the generation model, providing an interpretable first line of content moderation.
    \item \textbf{Text rendering.} The rewriter brings substantial gains here, as user intents for adding text are frequently ambiguous or imprecisely formatted (\eg, ``add a title'' without specifying the exact wording or style). By resolving these into explicit, well-structured specifications, fixing the precise string and placement, the rewriter markedly improves both the accuracy and the visual quality of the rendered text.    

\end{itemize}

While prompt rewriting is already widely adopted to improve generation quality, we re-emphasize a property that is often overlooked: a well-designed rewriter should be \emph{lower-bounded by the identity transformation}. That is, when a user's prompt is already clear, well-specified, and free of conflicts, the rewriter should leave it essentially unchanged rather than embellish it. Under this design principle, rewriting can only help and never hurt. In the worst case it preserves the original intent verbatim, and in the better cases it resolves ambiguity, underspecification, or conflict. This stands in contrast to rewriters that compulsively expand every prompt, which risk injecting hallucinated details and drifting away from a user request that needed no modification in the first place. Guaranteeing this lower bound is precisely what distinguishes a \emph{translator} from an \emph{enhancer}: the former intervenes only when intent is unclear, whereas the latter alters the input unconditionally.

\textbf{Stronger vision-language models amplify the efficacy of prompt rewriting.} As shown in Figure~\ref{fig:pe_analysis_a}, we observe a clear positive correlation between a model's intrinsic world knowledge and reasoning capacity and the quality gain it delivers through rewriting. Given an identical user query, a more capable backbone draws on richer world knowledge and stronger compositional reasoning to produce more coherent, intent-faithful, and context-aware rewrites. This gap is, in fact, one concrete manifestation of the broader divide between open- and closed-source models—the rewriting ability is bounded not by the rewriting pipeline itself, but by the knowledge and reasoning embedded in the underlying model.

We further observe that this latent capacity is highly contingent on the system prompt. A poorly structured system prompt acts as a bottleneck that suppresses the model's potential, whereas a carefully calibrated one serves as a catalyst that unlocks it. Consequently, realizing the full benefit of a strong backbone requires co-designing the rewriting prompt alongside the model.

\textbf{Different text-to-image models call for different prompt rewriters.} This is because each generation model is trained under its own conditions. First, models inject human preference along different axes: some are tuned toward photorealism, others toward illustration or aesthetic stylization, so the same prompt elicits different default behaviors. Second, models adopt different captioning strategies during training, meaning they expect input phrased in different styles and structures. A rewriter tuned for one model's caption distribution may therefore produce out-of-distribution input for another, degrading rather than improving its output.

A direct consequence is that comparing two models under an identical prompt is often unfair: the prompt may sit squarely within one model's training distribution while falling outside the other's. A meaningful comparison should instead pair each model with its own appropriately adapted prompt, evaluating each at its intended operating point rather than under a shared but model-agnostic input.

\subsubsection{Captions: Good Supervision Requires Understanding User Demand} 
\label{sec:good_captions}
The captioning strategy for training images must be designed around anticipated user demand. Captions serve as the supervisory signal that defines what the model can learn: if a capability is never described in the captions, the model has little chance to acquire it, and injecting that ability at inference time becomes extremely difficult, as neither prompt rewriting nor additional sampling can recover information that was never present in the supervision. Improving the alignment between an image and its caption is therefore central to the final quality of a text-to-image generation model.

Building such a high-fidelity captioning pipeline, however, is non-trivial, and existing practices face two recurring problems:
\begin{itemize}
    \item \textbf{Limited capability of moderate VLMs.} Mid-sized vision-language models~\cite{bai2025qwen3vl, bai2025qwen25vl,zhu2025internvl3exploringadvancedtraining} often produce inaccurate captions, miscounting objects, hallucinating attributes, or misjudging spatial relations, and this noisy supervision is faithfully inherited by the generation model.
    \item \textbf{Prompt sensitivity of strong VLMs.} Even a strong VLM~\cite{comanici2025gemini25pushingfrontier,comanici2025gemini25pushingfrontier} yields markedly different captions under different captioning system prompts. Without careful calibration, this inconsistency propagates as conflicting supervisory signals across the training set.
\end{itemize}

To address these issues, we systematically benchmark a set of candidate VLMs, including both open-source models such as Qwen2.5-VL-7B, Qwen2.5-VL-32B~\cite{bai2025qwen25vl}, Qwen3-VL-8B~\cite{bai2025qwen3vl}, and InternVL3-8B~\cite{zhu2025internvl3exploringadvancedtraining}, as well as closed-source models such as Gemini-2.5~\cite{comanici2025gemini25pushingfrontier} and Gemini-3~\cite{deepmind2026gemini31pro}. 
We measure their captioning accuracy along each aspect of interest, including object counting, spatial relations, attribute binding, style, and text rendering. Rather than relying on a single model or a single system prompt, we select, for each aspect, the VLM and prompt configuration that performs best, and aggregate their outputs into a unified, demand-driven caption. For dimensions that most VLMs fail to caption accurately, we further resort to expert models or human experts to annotate the corresponding tags.

This per-aspect design yields consistently more accurate captions than any single-VLM baseline, and the resulting improvement in image-caption alignment translates into measurable gains in the corresponding generation capabilities of the trained model.

\subsubsection{Model Router: The Right Model for the Right Task} 
\label{sec:model_router}
For a human painter, the time needed to complete a painting varies greatly from a simple sketch to an intricate composition. Motivated by this observation, we design our system in a similar spirit: an agent decides which model to invoke, thereby controlling the trade-off between quality and efficiency.

Such practices are already common in the open-source community, where experienced developers and artists select different models, adjust the number of diffusion steps, and so on. While it is unclear whether top-tier systems adopt similar strategies, one telling fact is that GPT-Image-2~\cite{openai2026gptimage2} can take up to $100\times$ longer than Z-Image-Turbo~\cite{zimage2025} to generate a very simple prompt such as ``a beautiful girl''. The same phenomenon arises within our Boogu-Image-0.1 series: in many cases the outputs of Boogu-Image-0.1-Turbo and Boogu-Image-0.1-Base are nearly indistinguishable, yet their inference costs can differ by 50 to 100 times.

Building on this insight, Boogu adopts an analogous strategy: rather than committing to a single model, it uses an agent to route each request to the most suitable variant in the Boogu-Image-0.1 series according to the estimated difficulty of the prompt, reserving the heavier Base model for genuinely complex requirements while handling simpler ones with the fast Turbo model. Similar ideas have been explored in the LLM community, where routing frameworks such as FrugalGPT~\cite{chen2023frugalgptuselargelanguage}, RouteLLM~\cite{ong2024routellm}, and Hybrid-LLM~\cite{ding2024hybrid} dynamically dispatch queries to models of different capacities to trade off quality against cost. 

Admittedly, employing multiple models introduces additional system complexity. Nevertheless, given the substantial cost savings, we believe this direction is well worth pursuing.

\clearpage

\subsection{General Discussion on {\model} Experiments}
\label{sec:gen_discuss}

In this section, we present discussions based on the experimental evaluation of the {\model} model series. While these observations may appear trivial at first glance, they are in fact of critical importance.

\label{general_finding}

\subsubsection{Discussions Beyond Leaderboards}
\label{sec:beyond_bench}
\paragraph{Inference-time cost should be considered in evaluations.}
A unified multimodal understanding-and-generation system differs fundamentally from a standalone text-to-image model: both the understanding and the generation components contribute substantially to end-to-end performance, and neither can be evaluated in isolation. This coupling has direct implications for inference cost. Strong closed-source systems such as GPT-Image-2 typically incur higher latency than open-source counterparts, in part because their quality reflects additional computation that is rarely accounted for in standard comparisons. {However, even the most popular benchmark, LMArena \cite{arena_ai_2026}, \underline{still does not take inference-time cost into account}.
}

\begin{figure}[h]
  \centering
  \includegraphics[width=0.65\linewidth]{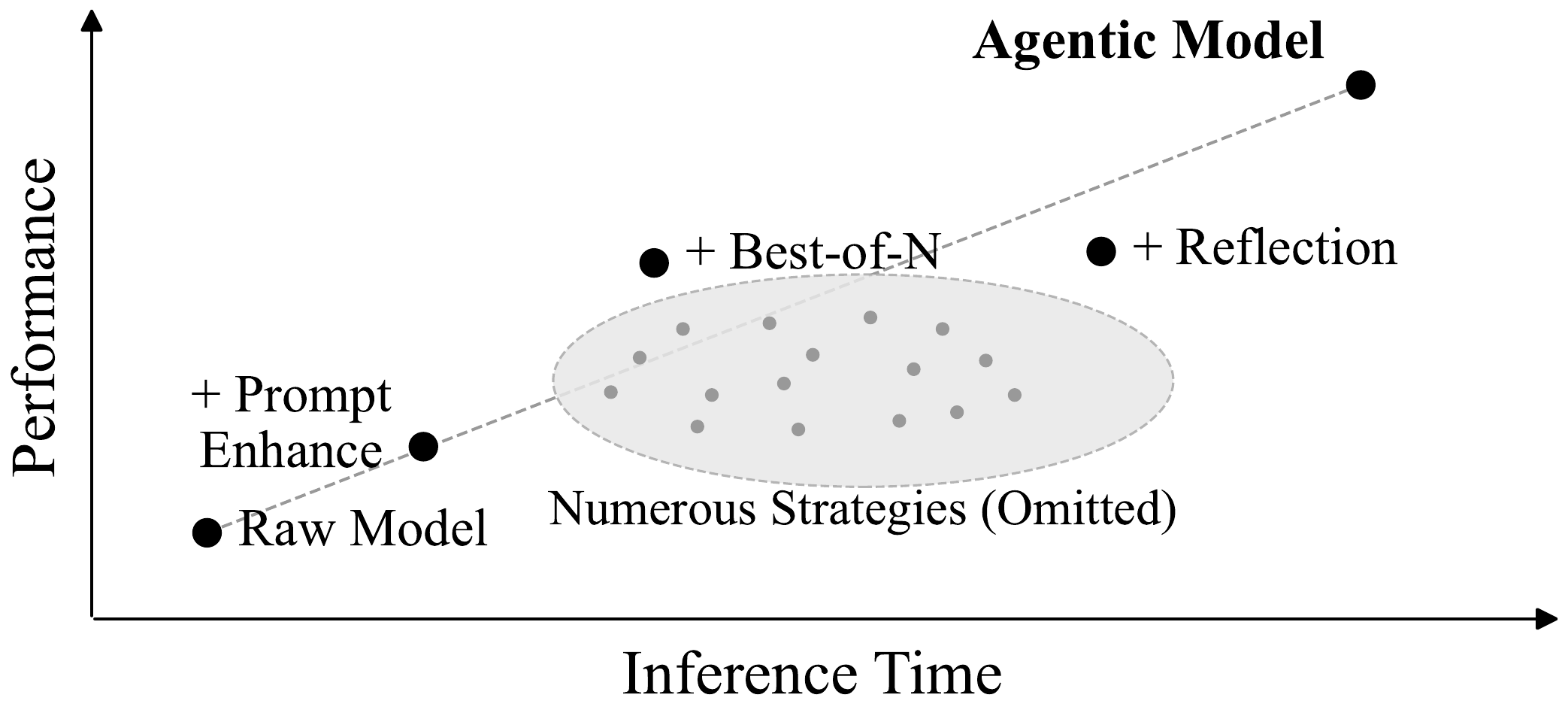}
\caption{\textbf{Trade-off between inference time and performance across strategies.} Performance progressively improves from the Raw Model (bottom-left) to the Agentic Model (top-right) by integrating techniques (\eg, Prompt Enhancement, Reflection, Best-of-N), albeit at higher inference costs. The gray region denotes intermediate strategies omitted for clarity, and the dashed curve illustrates the overall performance-efficiency trend.}
\label{fig:inferencetime_performance}
\end{figure}

As shown in Figure~\ref{fig:inferencetime_performance}, a broad class of techniques can improve output quality at the expense of inference time, including best-of-N sampling, self-reflection, prompt rewriting, and model ensembling~\cite{zhang2026qwenimageagentbridgingcontextgap}. As a recent example, GEMS~\cite{he2026gems} combines agents with Z-Image to outperform Nano-Banana on several benchmarks. These methods trade compute for quality along a continuous curve, which means that any quality figure reported without an accompanying latency budget is incomplete: the same model can occupy many points on this curve depending on how much test-time computation it is permitted to use. A meaningful comparison must therefore fix the inference budget across systems, or report quality and latency jointly as a Pareto frontier rather than as a single scalar.

In practice, however, most models report quality metrics without accounting for inference cost. This omission systematically favors methods that consume more test-time computation and obscures the genuine efficiency trade-offs that determine viability in deployment. A model that achieves a marginal quality gain at several times the latency may be preferable on a leaderboard yet unacceptable in a production setting.

We acknowledge that fair latency comparison is itself difficult. Closed-source systems must accommodate additional production constraints—including NSFW filtering, political-sensitivity handling, bias mitigation, and broader safety checks—that contribute to their measured runtime but are orthogonal to core generation quality. Isolating the cost attributable to generation alone is therefore non-trivial, and any cross-system latency comparison should be interpreted with this confound in mind.

Despite these challenges, we maintain that inference time must be incorporated as a primary evaluation axis rather than a secondary consideration. Quality and latency are not separable objectives; they jointly determine the practical utility of a model. Evaluating either in isolation produces conclusions that fail to transfer to real-world deployment.

\clearpage

\begin{figure}[h]
  \centering
  \includegraphics[width=\linewidth]{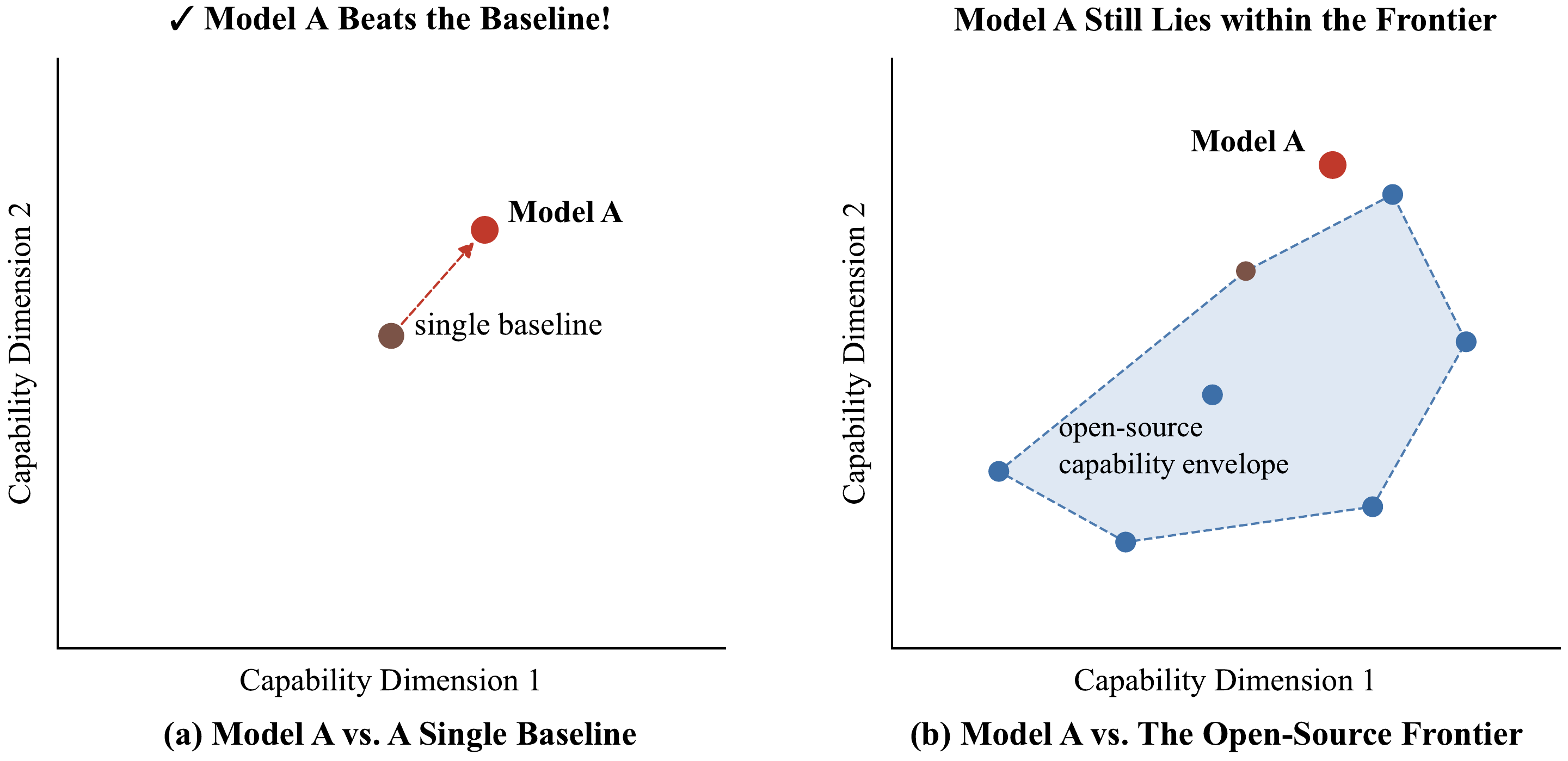}
\caption{\textbf{Single-baseline comparison versus the open-source capability frontier.} \textbf{(a)} Compared against a single reference model, Model A appears superior on every dimension, inviting an over-optimistic conclusion. \textbf{(b)} When all available open-source models are considered jointly—forming a capability envelope (dashed region) that users can draw on in practice—Model A still lies largely within this envelope. The meaningful target for open-source development is therefore to advance the collective envelope, not to outperform any single baseline.}
\label{fig:single_comparison}
\end{figure}

\paragraph{Over-reliance on a single baseline in A/B testing is methodologically flawed.}
During the intermediate phase of the project, we evaluated progress primarily through pairwise A/B comparisons against a single strong reference model. In retrospect, this framing was methodologically flawed. For open-source development, the appropriate evaluation target is not whether a model outperforms a specific competitor, but whether it improves upon the aggregate capability already available within the community. As shown in Figure~\ref{fig:single_comparison}, a favorable result against an individual baseline is both easy to obtain and prone to over-interpretation, since it reflects performance relative to a single point rather than to the prevailing capability frontier. The meaningful comparison is therefore A versus the union of all open-source work, that is, the effective capability envelope a user can assemble from currently available models and tools. We adopted this perspective only in the later phase of the project, after which our evaluation shifted from comparison against a fixed reference to comparison against this collective frontier.

This consideration further motivates designing evaluation sets around target application scenarios. A benchmark is informative only to the extent that it poses the questions that arise in real usage. We found that scenario-driven evaluation, using prompts derived from genuine production workflows, assessed by human raters, and aligned with deployment objectives, provides substantially more discriminative signal than static academic metrics. While such evaluation incurs higher construction and annotation cost, its correspondence to real-world conditions is what renders the resulting iteration reliable.

We note that strong closed-source models such as the Seedream~\cite{seedream4d5,gao2025seedream3,seedream5lite} series have already adopted multi-dimensional comparison, yet this practice remains uncommon in the open-source community. While many open-source works attempt to approximate it by combining several benchmarks, this fallback is undermined by the flaws in public benchmarks discussed above.

\clearpage

\paragraph{Open challenges for future evaluation.}
Even after reframing our evaluation methodology, several problems remain unresolved. Below, we highlight a few representative challenges to serve as a reference for the community, though this list is by no means exhaustive.

\begin{itemize}
    \item \textbf{ImageNet-like T2I benchmarks are required.} 
    The text-to-image community still lacks a standardized benchmark with the rigor that ImageNet \cite{deng2009imagenet} brought to recognition~\cite{krizhevsky2012imagenet,recht2019imagenet,he2016deep}. In particular, it lacks a clear and enforced separation between training and test data. At a minimum, models should control their training and test splits so that evaluation prompts (and the images associated with them) are guaranteed to be unseen during training. In current practice this discipline is largely absent: web-scale training corpora are scraped indiscriminately~\cite{schuhmann2021laion,schuhmann2022laion,jia2021scaling,radford2021learning}, and popular evaluation prompts or their near-duplicates frequently leak into the training set. As a result, reported gains may reflect memorization rather than genuine generative capability~\cite{somepalli2023diffusion,carlini2023extracting,dodge2021documenting}, and cross-model comparisons become unreliable. We argue that establishing a benchmark with documented data provenance, held-out test prompts, and explicit leakage controls is a prerequisite for measuring real progress in text-to-image generation.

    \item \textbf{World-knowledge evaluation.} 
    It remains unclear how to systematically assess whether a model captures real-world physical laws, spatial relationships, and causal structure~\cite{lin2026aegis,cai2025mm,guo2025r}. Existing benchmarks~\cite{niu2025wise,huang2023t2i,ghosh2023geneval} provide little coverage of these properties, and the failures they miss are often subtle. For example, a model may render a glass of water with physically inconsistent refraction, place a shadow on the wrong side of a light source, depict a mirror reflection that does not correspond to the scene, or generate a clock whose hands are inconsistent with a specified time. More demanding cases require compositional and causal reasoning. For instance~\cite{lin2026aegis,cao2024rwku,zhang2025worldgenbench}, ``a half-filled glass that has just been knocked over'' requires the model to represent both the resulting spill and the residual liquid. Such errors are easy for humans to identify but difficult to capture with any automated, scalable metric~\cite{hessel2021clipscore,hu2023tifa,xu2023imagereward}, and we do not yet have a reliable protocol for measuring them.

    \item \textbf{Ambiguous semantics.} 
    User prompts are frequently underspecified, and the space of acceptable outputs is correspondingly large~\cite{saharia2022photorealistic,yu2022scaling,kirstain2023pick}. An instruction such as ``a nice living room'' admits many valid interpretations, conditioned on user intent regarding style, lighting, and furnishing. The difficulty is compounded when prompts contain culturally dependent or contextual terms: ``a traditional breakfast'' implies different scenes in different regions, and ``a formal outfit'' varies with occasion and culture. Because no single reference output is correct, standard reference-based metrics are ill-suited to this setting~\cite{parmar2022aliased,kynkaanniemi2019improved,reiter2018structured,hu2023tifa}, and evaluating whether a model resolves ambiguity in a manner consistent with human expectations, rather than collapsing to a generic or averaged output, remains an open problem.

    \item \textbf{Inference time.} 
    Quality and latency are in direct tension~\cite{ho2020ddpm,song2022ddim,nichol2021improved,lu2022dpmsolver}: techniques that improve fidelity, such as best-of-\(N\) sampling~\cite{xu2023imagereward}, prompt rewriting~\cite{hao2023optimizing}, and model ensembling~\cite{podell2024sdxl}, do so at the cost of additional computation. A model that achieves a marginal quality gain at several times the latency may rank favorably on a leaderboard yet prove impractical in interactive applications, where response time is itself a component of user experience. Identifying the appropriate quality and latency Pareto frontier for each deployment scenario, whether real-time interaction, batch generation, or offline production, remains an unsolved problem.

    \item \textbf{Dynamic benchmarks.} These observations motivate the development of evaluation protocols that are dynamic, contamination-free, and aligned with human judgment~\cite{chiang2024chatbot,srivastava2023beyond,liang2022holistic}. The underlying risk is a classic instance of Goodhart's Law~\cite{chrystal2003goodhart}: once a benchmark becomes an optimization target, it ceases to be a reliable measure. Engineering efforts are consequently driven toward over-optimization of specific benchmark formats, which can inadvertently degrade a model's instruction-following and conversational capabilities in deployment. As a result, a model that surpasses its predecessor by 10\% on academic benchmarks may nonetheless exhibit inferior performance in complex, open-ended user scenarios. In the absence of such protocols, public benchmark scores should be interpreted with caution, and evaluation should remain anchored to the criterion that matters: whether a model enables real users to solve tasks.
    
\end{itemize}

\clearpage

\subsubsection{Data Ablation Evaluation: Just Enough Data Is Not Enough} 
\label{sec:data_ablation}
\paragraph{How much data do we need?} There are many open-source works that disclose their pretraining data. However, few of them discuss why a certain amount of data is required. The cost of training a foundation model, such as a text-to-image model, is highly correlated with the scale of the training data. Only a handful of organizations can afford to train such models, as the cost is prohibitive, including the computing resources, the disk storage needed to hold the data, and the expense of transferring it. Reducing this cost would open the problem to more organizations and broaden research in this area.

Prior excellent open-source models like Qwen-Image~\cite{qwen2025qwenimage}, Hunyuan-Image~\cite{cao2025hunyuanimage}, Krea2~\cite{krea-2-2026} mentioned that they utilized billions of images in the pretraining stages. Lens~\cite{chen2026lens} utilizes fewer data but also require 800M images for training, which is close to 1 billion images. While there are some other interesting models that use fewer data but their performance is somehow not to close strong models mentioned above. 

Boogu-Image-0.1 uses only 208.62 million unique images to achieve competitive performance, outperforming baselines that rely on substantially more training data and compute across many evaluation settings. In our opinion, the required volume of training data is mainly determined by two factors: the scope of knowledge, and the number of network parameters. As a foundation model, we usually hope it has good generalization ability and a wide range of knowledge. As the scope of knowledge grows, more training data is needed to represent it, consistent with studies of scaling laws~\cite{kaplan2020scaling,liang2024scaling_dit}.

\begin{figure}[h]
  \centering  \includegraphics[width=0.7\linewidth]{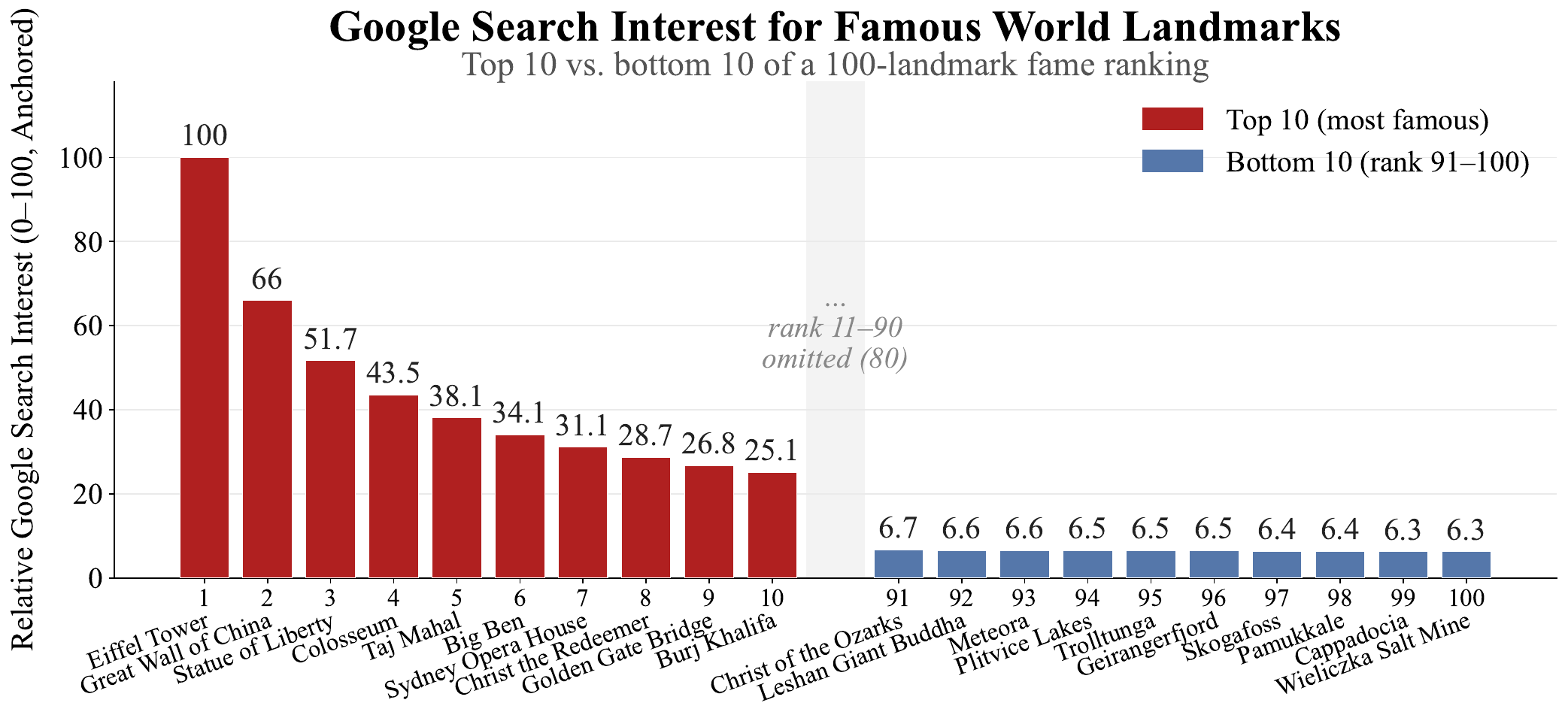}
  \vspace{-1mm}
\caption{\textbf{Google search interest across famous world landmarks.}
    Each landmark is ranked by general global fame; the bars show the top~10
    (most famous, left) and the bottom~10 (ranks 91--100, right), with the
    intervening 80 landmarks omitted for clarity, as marked by the central
    break. The vertical axis reports \emph{relative} search interest on
    Google Trends (indexed to 0--100 and anchored to the top-ranked landmark
    so that all bars are comparable), \emph{not} absolute query volume.
    Interest falls off steeply: the most famous landmarks attract many times
    the search interest of those in the long tail. \textit{Note:} the fame
    ordering is editorial rather than derived from a formal metric, so the
    figure is intended to illustrate the overall distribution rather than to
    provide an exact measurement.}
  \label{fig:landmark-search}
\end{figure}

To reduce training cost, Boogu adopts two strategies aimed at lowering the volume of training data. First, we adopt a DiT with 10 billion parameters. Rather than scaling up model size, we deliberately keep the model compact so that it remains accessible to the open-source community, where many users and developers do not have access to high-end GPUs. This choice shifts the burden of efficiency onto data curation rather than raw model scale. Second, it is well known that internet-scale data exhibits a severe long-tail distribution: for a popular concept, the available data can be orders of magnitude larger than that of a rare one. Take landmarks as an example. As shown in Figure~\ref{fig:landmark-search}, data availability is heavily skewed toward a few iconic landmarks, with the Eiffel Tower attracting far more search interest than the rest. We argue that such an abundance of head data is unnecessary for training: beyond a certain point, additional examples of an already well-covered concept yield diminishing returns rather than better performance.

Note that there is a trade-off between the cost of constructing a balanced distribution and the cost it saves during training. Such careful data curation is only worthwhile when the available compute is limited; otherwise, constructing a balanced distribution may cost more than simply training on the unbalanced pretraining data.

\clearpage
\paragraph{Explicitly captioning visual flaws instead of filtering flawed data.}
Very few categories of data actually need to be filtered out if your goal is to build a general-purpose image generation model. While training a foundational text-to-image model requires high-quality, aesthetically pleasing images with optimal exposure and composition to ensure generation fidelity, \textbf{suboptimal data, such as underexposed, poorly composed, blurry, or watermarked images, \underline{should not be entirely discarded}}. Unlike recent data pipelines such as Qwen-Image~\citep{qwen2025qwenimage}, Z-Image~\citep{zimage2025}, and JoyAI-Image~\citep{song2026joyai}, which commonly emphasize filtering images with visible artifacts or quality defects, we argue that these samples can remain valuable if paired with detailed and accurate captions. By explicitly describing all visual and logical content alongside their specific flaws (\eg, detailing the presence of watermarks or the exact nature of the blurriness), these images can be integrated into the training set at an appropriate ratio. This not only helps the model learn fundamental visual patterns but also enables it to understand negative visual elements, thereby acquiring a more comprehensive and controllable generation capability. In principle, almost any 2D image can be utilized; the real challenge is not whether to include such data but how to balance it, which demands extensive fine-grained tagging or manual annotation to enable precise control.

\begin{figure}[htbp]
    \centering
    \includegraphics[width=1.0\textwidth]{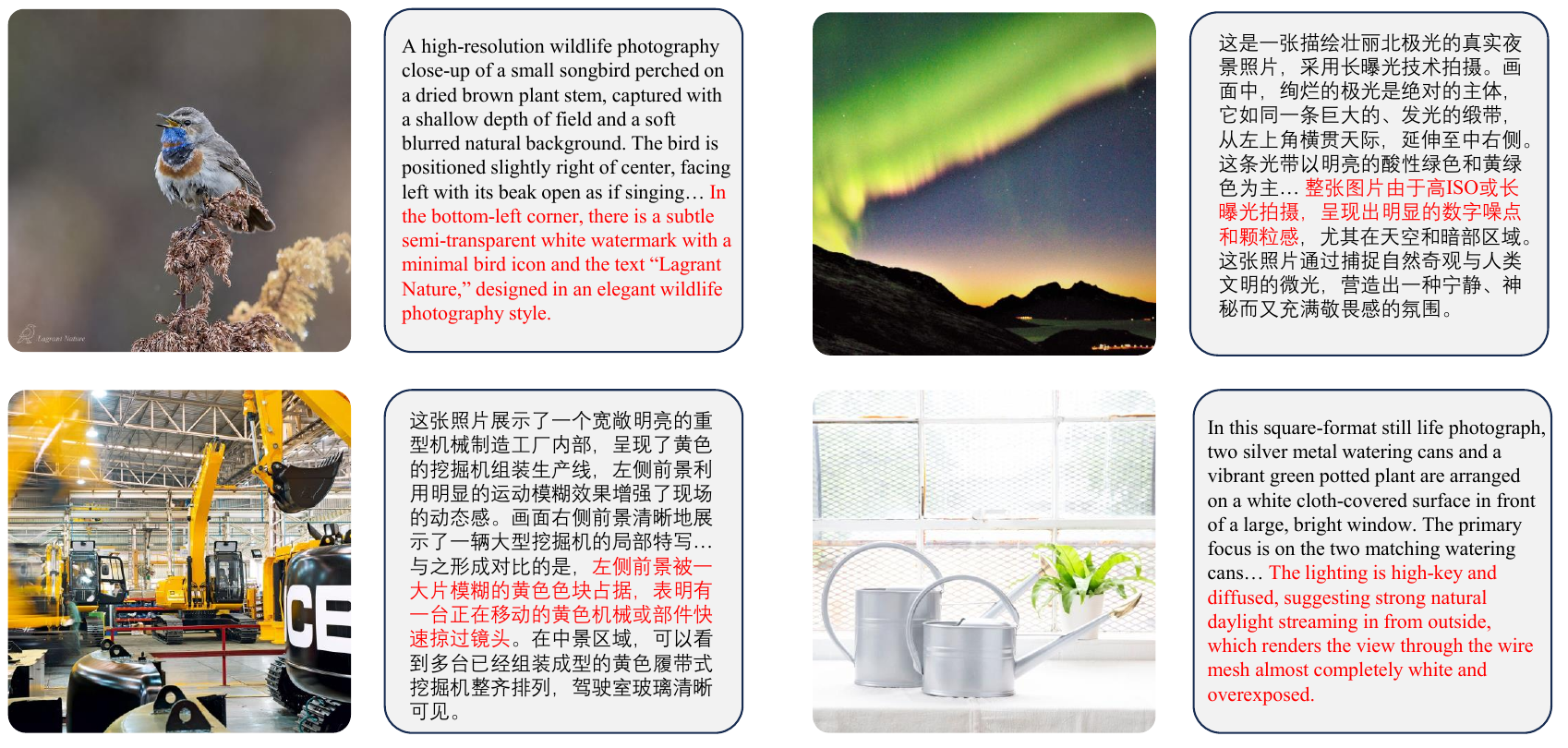}
\caption{\textbf{Common artifacts in open-source images, paired with captions that explicitly describe them.} Simply filtering out such images would deprive the model of the ability to understand these artifacts; instead, we retain them with captions that name the defect, enabling controllable generation. \textbf{Top-left:} a watermarked image. \textbf{Top-right:} a low-light night photo showing color noise upon magnification. \textbf{Bottom-left:} motion blur from a slow shutter speed. \textbf{Bottom-right:} significant overexposure.}
    
    \label{fig:flaw_as_feature}
\end{figure}

Figure~\ref{fig:flaw_as_feature} illustrates this idea with several common artifacts found in open-source images. Rather than discarding these samples, we retain them and pair each with a caption that explicitly names the defect: a visible watermark (top-left), color noise from a low-light night shot (top-right), motion blur caused by a slow shutter speed (bottom-left), and severe overexposure (bottom-right). By grounding each artifact in language, the model learns not only the underlying visual content but also the concept of the defect itself. This turns what would otherwise be filtered-out ``bad'' data into a signal for controllable generation, allowing the model to either avoid these artifacts or reproduce them on demand.

\clearpage

\paragraph{Quality over quantity: massive open-source datasets do not guarantee high performance.} For the initial pretraining phase, we construct a foundational data mixture entirely from publicly available sources, comprising both single image-text pairs and interleaved text-image documents. Specifically, we collect representative single-image corpora, including COYO~\citep{kakaobrain2022coyo}, DataComp~\citep{gadredatacomp2023}, PixelProse~\citep{singla2024pixelsproselargedataset}, and BLIP-3o~\citep{chen2025blip3onextfrontiernativeimage}. To further expand the overall data scale and enhance broad visual-concept coverage, we supplement these with a large-scale interleaved corpus derived from OmniCorpus~\citep{li2025omnicorpus}.

Our empirical findings demonstrate that relying solely on open-source datasets (here 187M in total) imposes a strict ceiling on model performance, rendering them inadequate for training state-of-the-art models. Despite their massive volume, open-source corpora suffer from pervasive quality issues, including high semantic noise, factual errors, and severe data duplication.

More importantly, these datasets lack the descriptive depth required for high-fidelity generation; their captions are predominantly brief and shallow, {failing to capture the fine-grained attributes, spatial relationships, and \underline{dense image-text alignments necessary} for precise and controllable synthesis.} Furthermore, open-source data is highly skewed toward generic web content, leaving a critical void in high-quality, domain-specific (\eg, scientific, logical, and localized) data. To bridge the gap to proprietary frontier models, developers must move beyond public datasets and focus on building sophisticated data-refinement pipelines, high-quality synthetic data generation, and meticulous expert-level curation.

\begin{figure}[htbp]
    \centering

    \colorbox{tablegray}{\parbox{0.96\linewidth}{\small \textit{Prompt 1:} A masked cat (anthropomorphized) grips a handrail in a subway car, its eyes looking tired; an advertisement in the car reads ``AI, keep going!'' Requirements: correct perspective, the cat's anthropomorphism is reasonable and cute, the advertisement text is legible, and the lighting is consistent.}}\\[0.06in]

    \colorbox{tablegray}{\parbox{0.96\linewidth}{\small \textit{Prompt 2:} Amidst heavy snowfall, a swordsman in a red cloak stands alone on the city wall, the wind whipping his cloak into sharp lines. Requirements: contrasting warm and cool colors (cold snow + red cloak); low-angle shot; sharp edges; natural pose.}}\\[0.12in]

    \includegraphics[width=0.24\textwidth]{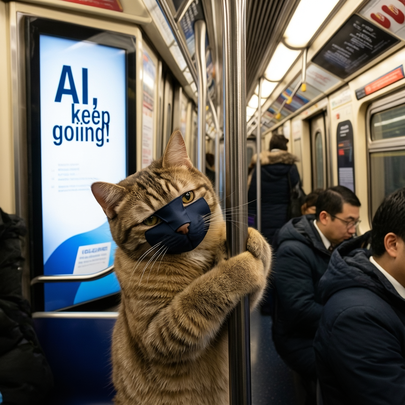}\hfill
    \includegraphics[width=0.24\textwidth]{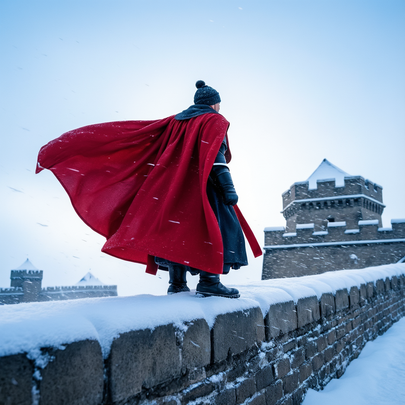}\hfill
    \includegraphics[width=0.24\textwidth]{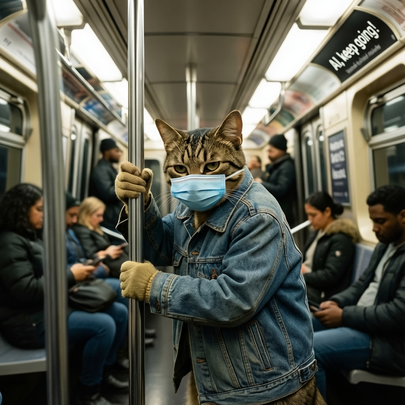}\hfill
    \includegraphics[width=0.24\textwidth]{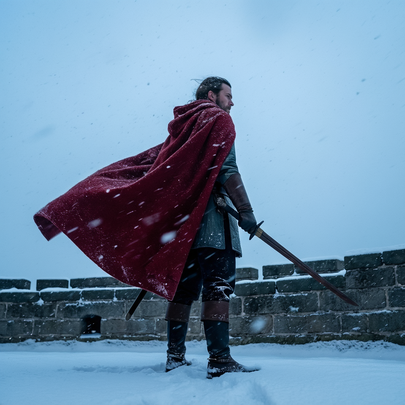}

\caption{\textbf{Training on the Boogu Syllabus yields markedly better generation quality than open-source data.} The left two images are generated by the model trained on open-source datasets, while the right two are generated by the model trained on the Boogu Syllabus (prompts shown above). Under identical settings, the Syllabus-trained model produces more accurate text rendering, more coherent composition, and better prompt adherence. }  
\label{fig:raw_vs_syllabus}
\end{figure}

As shown in Figure~\ref{fig:raw_vs_syllabus} and Table~\ref{tab:raw_vs_syllabus}, a model trained on data structured by a carefully designed, human-prior-based syllabus significantly outperforms one trained on massive (187M) open-source data in terms of instruction comprehension and following, world knowledge, and photographic quality.

\begin{table}[htbp]
    \centering
    \begin{tabular}{lccc}
        \toprule
        Training Data & Qwen-Image-Bench (CN+EN) $\uparrow$ & LongText EN $\uparrow$ & LongText ZH $\uparrow$ \\
        \midrule
        Open-source Data & 48.45 & 0.828 & 0.869 \\
        +Boogu Syllabus   & \textbf{53.65} & \textbf{0.952} & \textbf{0.969} \\
        \bottomrule
    \end{tabular}
    \caption{\textbf{The Boogu Syllabus substantially outperforms open-source data.} Under identical model and training settings, training on the curated Boogu Syllabus improves performance across all three benchmarks over training on raw open-source datasets. Best scores are \textbf{bolded}.}
    \label{tab:raw_vs_syllabus}
\end{table}

\clearpage

\begin{figure}[htbp]
    \centering

    \colorbox{tablegray}{\parbox{0.95\linewidth}{\small \textit{EN:} A street photography shot of an elderly scavenger with a deeply weathered and wrinkled face in the center of the frame. The scene includes a trash can and a traffic light in the background. High photographic texture, classic street photography aesthetic, cinematic lighting, photorealistic.}}\\[0.06in]

    \colorbox{tablegray}{\parbox{0.95\linewidth}{\small \textit{ZH:} \begin{CJK*}{UTF8}{gbsn}生成一张街拍的照片，要有垃圾桶，红绿灯。画面中心有一个拾荒的老人，满脸沧桑。照片要有摄影质感，街拍质感。\end{CJK*}}}\\[0.12in]

    \includegraphics[width=0.24\textwidth]{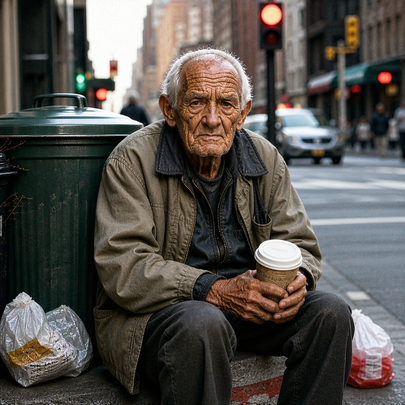}\hfill
    \includegraphics[width=0.24\textwidth]{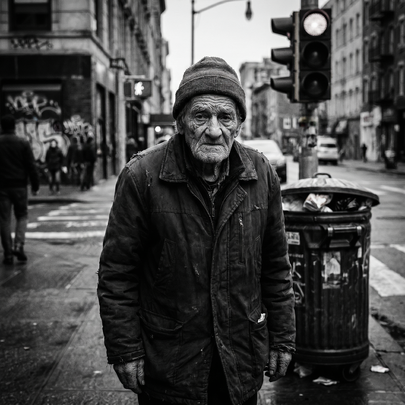}\hfill
    \includegraphics[width=0.24\textwidth]{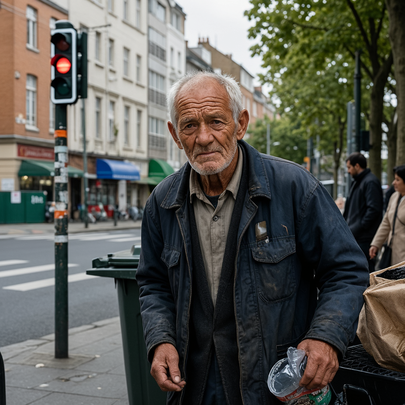}\hfill
    \includegraphics[width=0.24\textwidth]{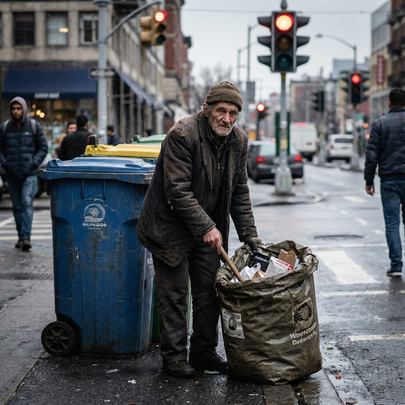}

    \caption{\textbf{Semantically equivalent prompts in different languages produce culturally biased outputs.} The left two images are generated from the English prompt and the right two from the Chinese prompt (shown above), despite the two prompts describing the same scene. Both sets of images predominantly exhibit Western cultural elements (\eg, facial features and street scenes), even when prompted in Chinese. This demonstrates the aforementioned cultural bias inherent in the open-source pre-training data.}
    \label{fig:culture_bias}
\end{figure}

\paragraph{Cultural bias ingrained in open-source data persists despite targeted SFT.} We identify a distinct capability gap across cultural contexts, where the model acquires Western concepts (\eg, Caucasian features) more readily than Eastern ones. We attribute this discrepancy to the inherent Western-centric bias in large-scale open-source pre-training data, which is heavily dominated by Western cultural contexts---a systemic imbalance that remains difficult to quantify given the sheer volume of the data. Our experiments reveal that even incorporating a large-scale, high-quality Chinese dataset during the SFT stage cannot fully rectify this bias. Because foundational worldviews and cultural associations are established during the massive pre-training phase, downstream SFT only scratches the surface of behavioral alignment without fundamentally altering the model's latent representation space. As a result, the model still defaults to Western-centric patterns in its generations (see Figure~\ref{fig:culture_bias}), suggesting that cultural diversity must be addressed during pre-training rather than treated as an afterthought in fine-tuning.

As shown in Figure~\ref{fig:culture_bias}, we present image samples generated by our fine-tuned model using semantically equivalent Chinese and English instructions. Notably, even when prompted with Chinese instructions, the model exhibits a markedly high probability of generating Western faces or defaulting to Western cultural elements, such as architectural designs, street layouts, and clothing styles. While varying the random seed introduces some visual diversity, the predominant emergence of this Western-centric aesthetic remains highly consistent. This phenomenon clearly demonstrates the cultural bias discussed above, underscoring the urgent need for the open-source community to collect and curate datasets with greater cultural diversity.

\clearpage

\paragraph{Syllabus-guided data curation with human priors significantly boosts performance.} 
\label{sec:boogu_syllabus}
Pure visual understanding is inherently a ``dimensionality reduction'' task involving information compression—mapping rich visual signals to lower-dimensional text. Regardless of how exhaustive a natural language description is, it inevitably loses some of the original visual information. Consequently, generating an image or video based solely on even the most detailed text will likely yield results that differ from the original source. Conversely, Text-to-Image (T2I) generation is a ``dimensionality expansion'' task, granting the model immense creative freedom to synthesize countless valid visual variations that satisfy a given user prompt.

Recent studies on visual understanding models reveal several critical insights:
{\uline{(1)} Model capabilities exhibit a severe long-tail distribution with respect to their training data} \citep{parashar2024neglected}. While models excel at common concepts (\eg, cats or dogs), they struggle with rare elements due to the inherent long-tail nature of mainstream datasets.
{\uline{(2)} Multimodal models require exponential data exposure to grasp a concept} \citep{udandarao2024pretraining}; specifically, zero-shot performance grows linearly only with exponentially increasing exposures to similar data.
{\uline{(3)} Scaling up model parameters alone cannot compensate for unseen data types} \citep{chen2025reproducible}. If a specific data category is absent during training, the model cannot align its visual and textual representations, rendering it incapable of understanding that concept, regardless of model size.

Given that these limitations apply to the dimensionality reduction task like visual understanding, we argue that the dimensionality expansion task of T2I generation imposes even stricter requirements on data and training. To circumvent the aforementioned issues, we propose the following data curation principles:

\begin{itemize}
    \item \textit{Systematic and Fine-Grained Deconstruction:} Data serves as the textbook for model learning. Just as physics textbooks deconstruct knowledge into distinct branches (\eg, thermodynamics, electrodynamics) based on human priors to facilitate learning, systematically deconstructing data into fine-grained, logically interconnected components forms a structured syllabus that enhances the AI's understanding.

    \item \textit{Comprehensive Coverage:} The dataset must be extensively comprehensive; {ideally, no required capability should fall outside the planned scope}. We must proactively curate data for every target capability rather than relying on the model to achieve sudden ``epiphanies'' or zero-shot generalization through sheer scale. Simply put, if the model has never seen a cat during training, we assume it cannot generate one.

    \item \textit{Ample Data per Category:} Each fine-grained domain must contain sufficient data volume. As empirically verified in Section \ref{sec:concept_mem}, we estimate the requisite number of exposures needed for a model to memorize a concept and ensure that each data category meets this threshold during our curation process.
\end{itemize}

By adhering to these principles, we aim to \underline{mitigate the long-tail effect} in model capabilities and systematically guarantee robust, grounded understanding, moving beyond the naive reliance on scaling data volume or model parameters in hopes of spontaneous emergence.

{In short}, mere data abundance is far from sufficient. To truly enhance a model's comprehension and learnability, data must be structured and organized guided by human priors—a process that inherently injects human knowledge into the learning pipeline. Next, we detail the data distribution of our {\model} syllabus.

\begin{figure}[h]
    \centering
    \includegraphics[width=0.68\textwidth]{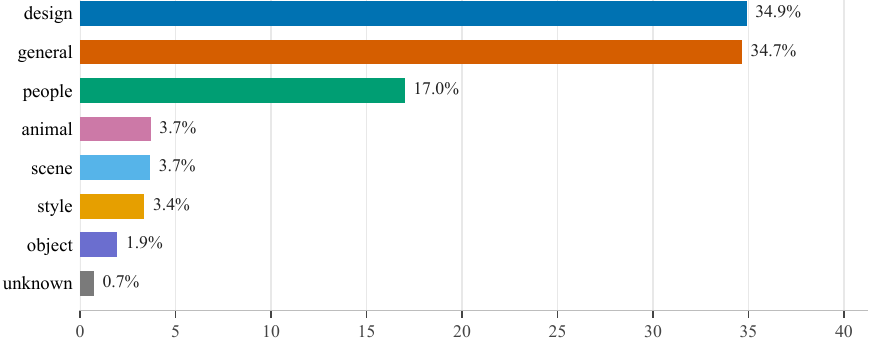}
    \caption{\textbf{Macro-level data distribution of the {\model} syllabus.}}    
    \label{fig:main_data_dis}
\end{figure}

\begin{figure}[!t]
    \centering
    \includegraphics[width=0.81\textwidth]{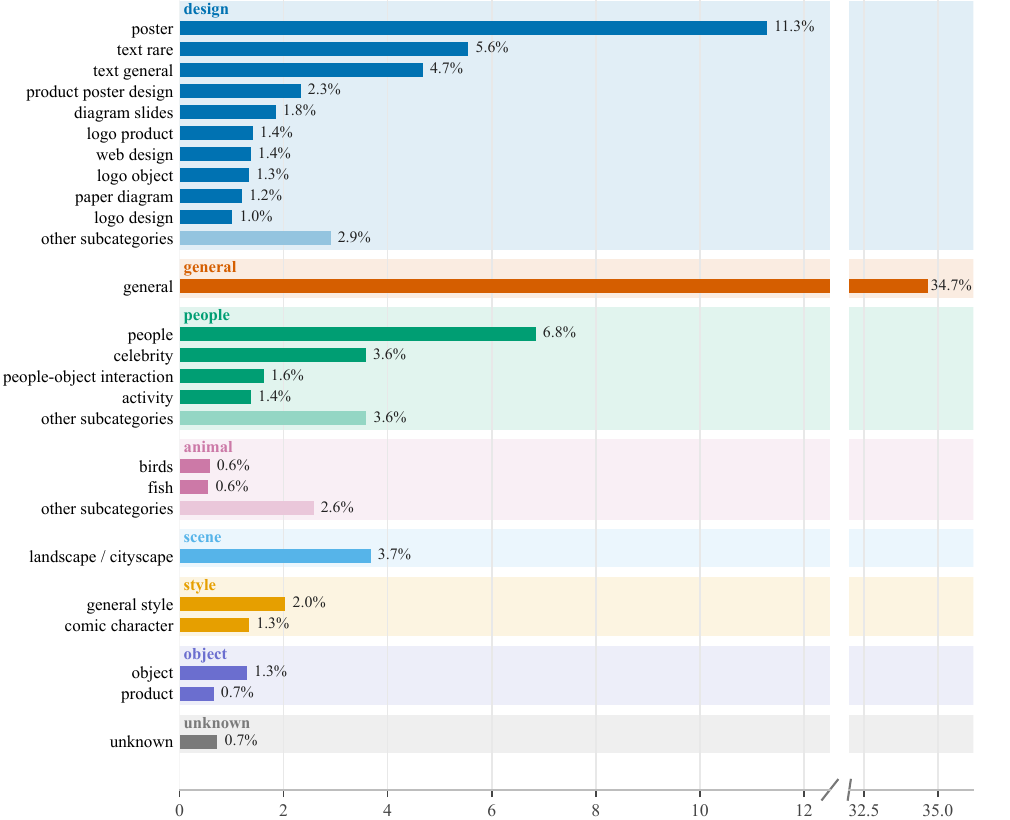}
    \caption{\textbf{Micro-level data distribution of the {\model} syllabus.}}    
    \label{fig:sub_data_dis}
\end{figure}
A well-designed distribution can potentially reduce the required volume of training data. Guided by human priors, we deconstruct visual scenes and organize them by their underlying logical relationships, structuring the data into a syllabus that facilitates the model's learning. Specifically, we assign higher sampling weights and construct fine-grained data for complex tasks. For example, as illustrated in Figures \ref{fig:main_data_dis} and \ref{fig:sub_data_dis}, graphic design constitutes a substantial and finely sub-categorized portion of our {\model} syllabus, reflecting its multifaceted nature: it requires interrelated capabilities—such as accurate typography, logo and poster design, and composition—that collectively determine the final visual quality.

These targeted strategies ensure that ``\textbf{no required capability should fall outside the planned scope},'' enabling the model to systematically comprehend diverse tasks and accurately align with the necessary visual elements. In summary, our {\model} syllabus comprises 21.62M unique samples. During training, these are upsampled to 47.19M based on our predefined weighting ratios. Notably, our model's comprehensive performance significantly surpasses that of models trained solely on 187M open-source samples (see Table \ref{tab:raw_vs_syllabus}).

\clearpage

\clearpage

\paragraph{Systematically enumerating vocabulary for reliable text rendering.}
\label{sec:text_render}
To improve the model's ability to render visual text, particularly Chinese characters, we propose a systematic data construction strategy based on vocabulary enumeration. This targeted approach proves far more effective than undirected data augmentation, which often suffers from coverage omissions and redundant sampling.

\begin{figure}[htbp]
    \centering
    \includegraphics[width=0.5\textwidth]{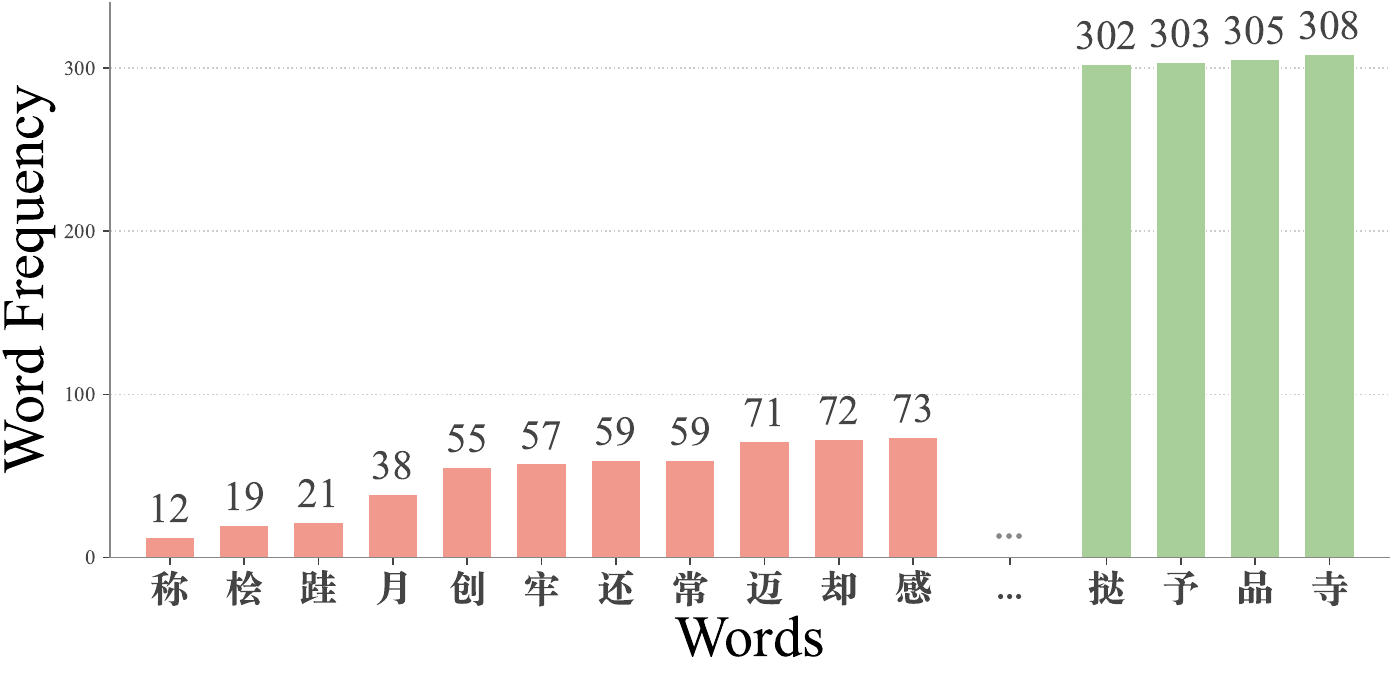}
\caption{\textbf{Per-character word frequency of a subset.} The horizontal axis lists Chinese characters in the ablation subset, sorted by how often they appear in training (high to low), and the vertical axis reports each character's generation accuracy. Characters seen at least ${\sim}300$ times are generated reliably, whereas rarer characters degrade sharply—indicating that roughly 300 exposures suffice for accurate generation.}    \label{fig:word_freqs}
\end{figure}

\begin{figure}[htbp]
    \captionsetup{skip=4pt}
    \centering
    \sbox0{\includegraphics[width=0.64\textwidth]{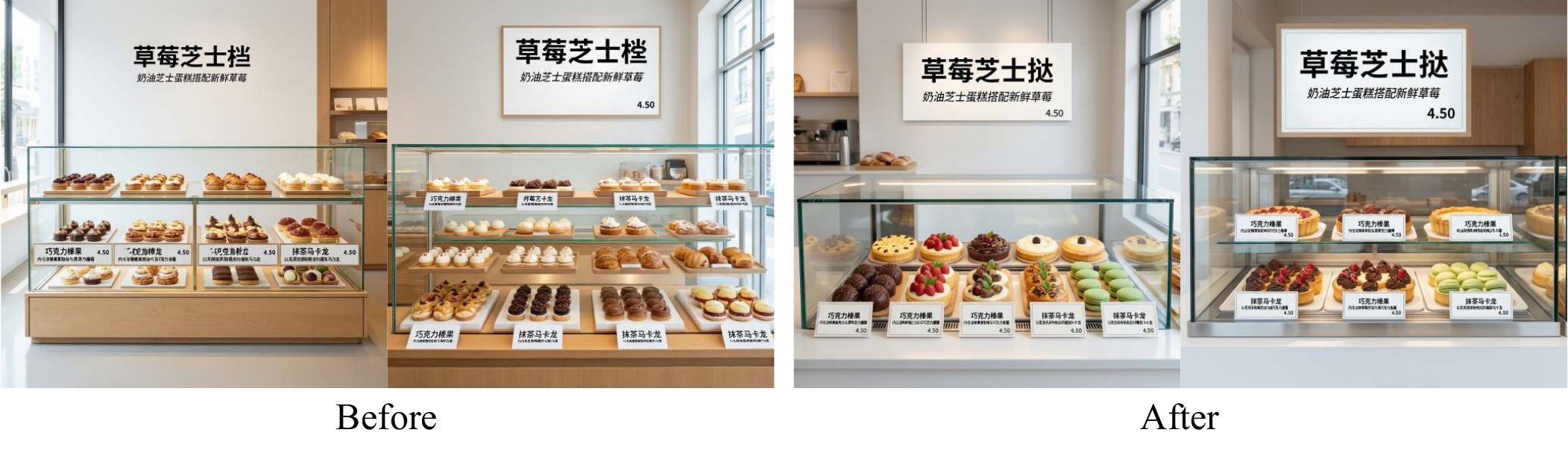}}%
    \begin{minipage}[t]{0.64\textwidth}
        \vspace{0pt}%
        \centering
        \usebox0
        \caption{\textbf{Correcting rare character rendering via targeted exposure scaling.} The figure contrasts the rendering performance before (left two panels) and after (right two panels) optimization, \ie, after SFT on a curated synthetic dataset that scales the training exposure of each Chinese character beyond $300$ times, the previously unrenderable character \colorbox{gray!20}{挞} is now correctly and reliably rendered.}
        \label{fig:text_render}
    \end{minipage}%
    \hfill
    \begin{minipage}[t]{0.33\textwidth}
        \vspace{0pt}%
        \begin{minipage}[c][\ht0][c]{\linewidth}
            \footnotesize
            \centering
            \begin{tabular}{lc}
                \toprule
                 SFT Optim. & {LongTextBench (ZH) $\uparrow$} \\
                \midrule
                Before & 0.9055 \\
                After & \textbf{0.9538} \\
                \bottomrule
            \end{tabular}
        \end{minipage}
        \captionof{table}{\textbf{Target-exposure scaling via SFT optimization substantially boosts the model's performance on the Chinese subset of LongTextBench.}}
        \label{tab:zh_optim_cmp}
    \end{minipage}
\end{figure}

Our empirical analysis reveals that the model is virtually incapable of rendering unseen Chinese characters, requiring a minimum of 300 training exposures per glyph for accurate generation. To validate this, we constructed a subset of Chinese characters with varying frequencies (Figure \ref{fig:word_freqs}). Models trained on this subset consistently and correctly generate characters observed over 300 times. Furthermore, for characters the model initially fails to render, introducing them for more than 300 exposures during subsequent training ensures stable and accurate generation (Figure \ref{fig:text_render}).

Under resource and time constraints where enumerating the entire Chinese vocabulary is infeasible, prioritizing the 3,500 most frequently used modern Chinese characters \cite{frequently_used_zh} serves as a practical alternative. By ensuring each character appears at least 300 times in the dataset, the resulting model can satisfy everyday Chinese text rendering demands. Table \ref{tab:zh_optim_cmp} presents the performance gains resulting from optimization targeted exclusively at the 3,500 frequently used Chinese characters.

Ultimately, this proactive data construction strategy is significantly more efficient than blindly accumulating open-source data in an attempt to acquire text rendering capabilities through sheer volume.

\clearpage

\subsubsection{Ablation Experiments on Multimodal Learning and Understanding}
\label{sec:discuss_learn}

\begin{figure}[htbp]
    \centering

    \colorbox{tablegray}{\parbox{0.95\linewidth}{\small \textit{Prompt:} \begin{CJK*}{UTF8}{gbsn}这张横向构图的室内写实摄影照片主体是男青年陈国轩，他正端着一个生日蛋糕。画面采用近景拍摄，光线聚焦在陈国轩和蛋糕上。他拥有整齐的黑发，戴着银色金属细框眼镜，表情平和且带着一丝微弱的笑意。其上身穿着一件带有帆船线条图的白色T恤，外搭一件蓝白相间、带有大面积格纹和晕染质感的翻领长袖衬衫。陈国轩双手托着一个白色方形盘，盘中是一个圆形生日蛋糕，蛋糕侧边和底部洒满了淡黄色的杏仁碎，顶部点缀着新鲜的草莓半块、绿色的猕猴桃片和其他水果。蛋糕正中心插着一个圆形的金色透明牌，上面印有艺术字体的 ``Happy Birthday''，两根正在燃烧的橘红色细长蜡烛立在插牌侧前方，火光在金色背景处形成了明亮的亮斑。背景是虚化的浅色室内墙面，整体氛围温馨。画面左右两侧存在轻微的视角透视拉伸感，文字清晰可见。\end{CJK*}}}\\[0.1in]

    \includegraphics[width=0.97\textwidth]{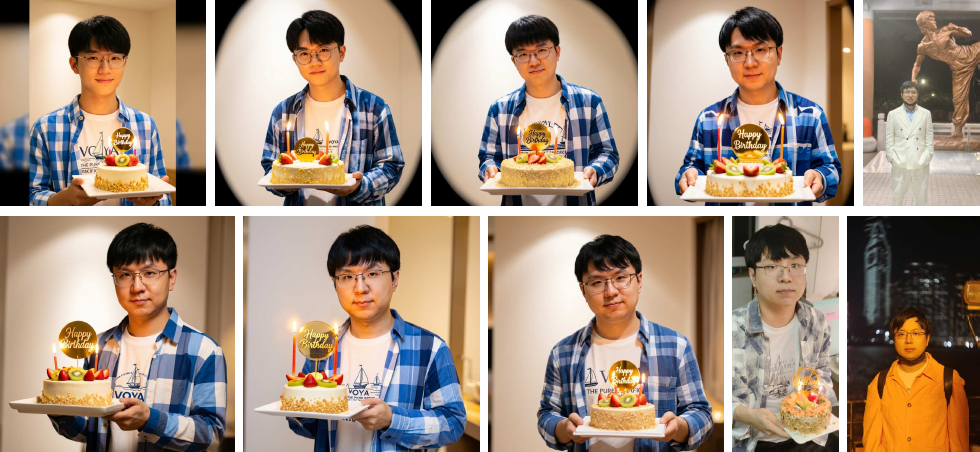}

    \caption{\textbf{The generated author identity progressively converges to the ground-truth subject.} Top row: checkpoints at 1k, 2k, 3k, and 5k steps; bottom row: 6k, 8k, and 9k steps. Every 1k steps corresponds to approximately 1{,}000 exposures to the ground-truth data (500 with English and 500 with Chinese captions). The top-right ($\sim$70\,kg) and two bottom-right ($\sim$75\,kg and $\sim$80\,kg) images are ground-truth training samples of the author at different weights. With increasing training, the model captures the subject's identity with growing fidelity, using the same prompt.}
    \label{fig:concept_memory}

\end{figure}

\paragraph{Robustly memorizing an unseen concept requires crossing a quantifiable exposure threshold.} 
\label{sec:concept_mem}
To evaluate the model's ability to memorize specific human identities, we incorporated 170 daily photographs of a volunteer (Guoxuan Chen) into an ablation dataset. Each image was annotated with four caption variations (Chinese/English, short/long descriptions). Through oversampling, we expanded this subset to 11,628 images (5,814 per language) and mixed it into a comprehensive 12.6M SFT dataset, constituting $\sim$0.13\% of the total data. The dataset was randomly shuffled to ensure uniform distribution. We trained the model using 128 devices with a global batch size of 1024, applying a constant learning rate of 1e-4 after a 500-step warmup. Using a volunteer's photos rather than celebrities ensures that the model had strictly no prior exposure to this identity during pre-training. The rightmost images in Figure~\ref{fig:concept_memory} show the ground truth photos included in the training set: a long shot of the volunteer at $\sim$70kg (top right), a close-up at $\sim$75kg (bottom right, first), and a medium shot at $\sim$80kg (bottom right, second). Notably, the subject's appearance varies slightly across these periods due to fluctuations in weight and personal condition. Note that the flow matching \cite{lipman2023flowmatching} model tends to learn an averaged representation of these variations.

The generated results from different training checkpoints are shown in Figure~\ref{fig:concept_memory}. The first four images in the top row correspond to 1k, 2k, 3k, and 5k steps, while the first three in the bottom row correspond to 6k, 8k, and 9k steps. We tested the model using Chinese prompts. At 1k steps, the model had encountered the volunteer's photos approximately 1,000 times (including around 500 times with Chinese captions). As shown, the model begins to generate recognizable facial features around 5k steps and achieves full recognizability by 8k to 9k steps. This indicates that a model requires about 4,500 language-matched exposures during training to robustly memorize a completely unseen human identity based on its associated textual identifier.

\clearpage

\paragraph{Use reinforcement learning with care when training text-to-image models.}
Distributional diversity is a core requirement for a general-purpose text-to-image model: users expect to generate not only mainstream-looking subjects but also those that deviate from prevailing aesthetic norms. Preserving the breadth of the output distribution is therefore essential to broad usability.

\begin{figure}[h]
    \centering
    \begin{subfigure}[t]{0.32\linewidth}
        \includegraphics[width=\linewidth]{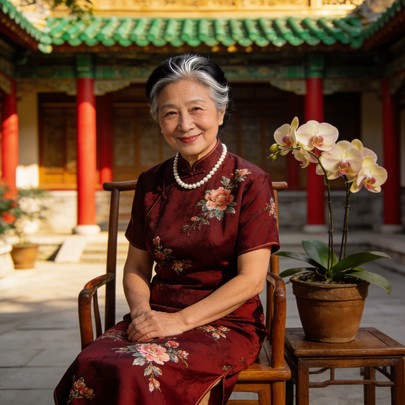}
        \caption{Seedream-4.5}
        \label{fig:rl_a}
    \end{subfigure}
    \hfill
    \begin{subfigure}[t]{0.32\linewidth}
        \includegraphics[width=\linewidth]{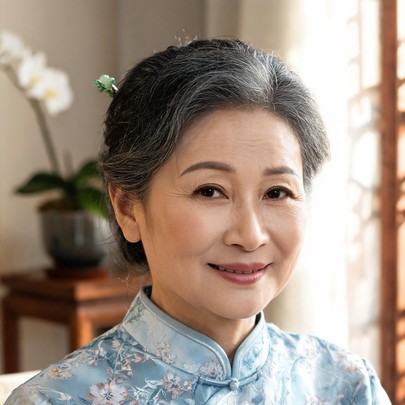}
        \caption{Seedream-5.0-lite}
        \label{fig:rl_b}
    \end{subfigure}
    \hfill
    \begin{subfigure}[t]{0.32\linewidth}
        \includegraphics[width=\linewidth]{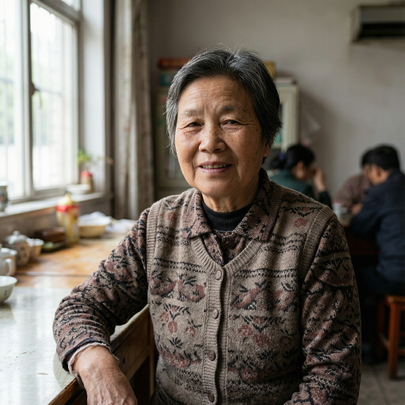}
        \caption{Boogu-Image-0.1-Turbo}
        \label{fig:rl_c}
    \end{subfigure}
\caption{\textbf{Generated images for the prompt \textit{``A sixty-year-old Chinese woman''}.} Models trained with heavy aesthetic RL (a, b) collapse toward mainstream preferences, rendering the subject with idealized, made-up features rather than the plain, natural appearance implied by the prompt, whereas our model (c) produces a more realistic result. We emphasize that such RL-driven distribution shaping is not inherently a weakness; it is a legitimate product strategy. It simply conflicts with our goal of maximizing diversity.}
\label{fig:rl_diversity}
\end{figure}

A common technique is to apply reinforcement learning (RL)~\cite{liu2025flowgrpotrainingflowmatching} to improve aesthetics and generation stability. However, heavy RL tends to collapse the output distribution toward a narrow domain. For instance, Seedream models~\cite{seedream2025seedream40nextgenerationmultimodal} achieve impressive portrait results—the generated people are consistently beautiful and young, matching mainstream preferences—yet generating subjects that depart from this aesthetic becomes considerably harder and sometimes infeasible. The very alignment that raises average quality also narrows the reachable distribution. As shown in Figure~\ref{fig:rl_diversity}, given the prompt \textit{``A sixty-year-old Chinese woman''}, models trained with heavy aesthetic RL tend to render the subject with polished makeup and idealized, glamorized features, whereas our model produces a more natural and realistic result. We emphasize that such RL-driven distribution shaping is not a weakness; it is a legitimate product strategy. It simply conflicts with our goal of maximizing diversity.

Guided by this objective, we deliberately avoid heavy aesthetic RL. Instead, we embed human preference within the understanding system rather than baking it irreversibly into the generator, allowing style—including attributes such as human faces—to be controlled flexibly at inference time while the generator's underlying diversity is preserved. As a result, our model readily generates subjects across a wide aesthetic range, including those outside mainstream norms.

This does not mean abandoning RL entirely. RL remains valuable for targeted issues that do not compromise diversity, and it delivers real gains in reducing anatomical artifacts and improving text rendering. However, RL training incurs substantially higher computational cost: it requires reward-model inference, repeated sampling, and many additional optimization steps on top of the base training pipeline. To build {\model} in a cost-efficient manner, we therefore limit RL to the few targeted issues where it is indispensable—such as anatomical distortion and text rendering—rather than applying it broadly for aesthetic tuning.

\clearpage

\paragraph{Dynamic time shifting does not scale trivially to 2K resolution.} In diffusion and flow matching models, logit-normal timestep sampling combined with resolution-dependent timestep shifting (dynamic time shifting) has become a widely adopted standard \cite{esser2024scaling,flux2024blackforest, lipman2023flowmatching, liu2022rectifiedflow}. These strategies are renowned for optimizing the signal-to-noise ratio and accelerating convergence, particularly when training high-resolution models. 

However, \textbf{during the development of the {\model} series---one of the pioneering open-source models natively trained at 2K resolution---we observed a phenomenon. Applying the default logit-normal and dynamic time shifting strategies significantly \textit{hinders} the convergence rate at this high-resolution scale.} Consequently, models in the early to intermediate stages of training exhibit noticeable noise artifacts in their generated outputs. Here, we provide a detailed analysis. For standard logit-normal sampling, we have:
\[
 t=\text{Sigmoid}(u) = \frac{1}{1+e^{-u}};  ~~u \sim \mathcal{N}(0,1)
\]
To model resolution-dependent timestep shifting, we typically employ linear interpolation to quantify how the number of latent tokens, \(n_T\), influences the distribution of the timestep \(t\). Calibrated by two pre-defined reference points \((x_1, y_1)\) and \((x_2, y_2)\), the shift factor \(\mu(n_T)\) is formulated as:
\[
\mu(n_T) = k \cdot n_T + b ; ~~k = \frac{y_2 - y_1}{x_2 - x_1}, \quad b = y_1 - k \cdot x_1
\]
where \(n_T\) denotes the resolution-dependent token count in the latent space. The shifted timestep, \(t_{\text{shift}}\), is then computed as follows (note that in our {\model} series, \(t \to 0\) denotes the high-noise regime):
\[
t_{\text{shift}} = \tau\left(t ~; \mu(n_T),\sigma \right) =  \frac{t^\sigma}{t^\sigma + e^{\mu(n_T)} \cdot (1 - t)^\sigma}; ~~\sigma > 0
\]
Fundamentally, this operation is equivalent to a translation in the logit space:
\[
\text{logit}(t_{\text{shift}}) = \log \left( \frac{t_{\text{shift}}}{1 - t_{\text{shift}}} \right) = \sigma \cdot\text{logit}(t) - \mu(n_T); ~~\text{logit}(p) = \log\frac{p}{1-p}
\]
This reveals that the complex time-shifting reduces, in logit (log-odds) coordinates, to a simple affine transformation: a scaling by \(\sigma\) followed by a translation of \(-\mu(n_T)\). With the common choice \(\sigma = 1\), it degenerates into a pure translation by \(-\mu(n_T)\). Consequently, a larger \(\mu(n_T)\)---induced by higher-resolution images---shifts the overall distribution towards smaller \(t\). While this aligns with the intuition that higher resolutions demand more computation at the high-noise stages to resolve global structures, applying this standard dynamic shifting strategy at 2K resolution leads to an \textit{over-squeezing} effect (see Figure \ref{fig:dyn_time_shift}, right). This excessive shift overly degrades the signal-to-noise ratio during training, ultimately slowing the convergence rate.

\begin{figure}[hbp]
    \centering
    \includegraphics[width=0.95\textwidth]{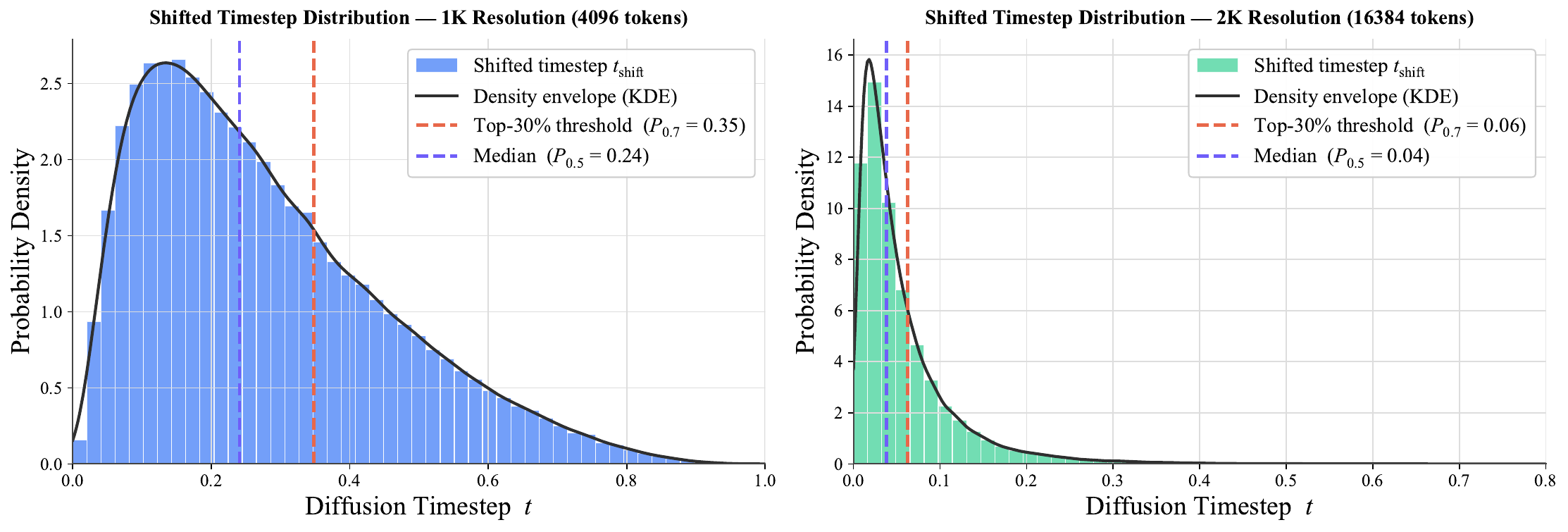}

    \caption{\textbf{Shifted timestep distributions} at 1K and 2K resolutions ($\sigma=1$), calibrated using \( (x_1, y_1)=(256, 0.5) \) and \( (x_2, y_2)=(4096, 1.15) \). The right panel illustrates the over-squeezing effect observed at the 2K resolution.}
    
    \label{fig:dyn_time_shift}
\end{figure}

\paragraph{Rectified Dynamic Time Shifting.}
\label{sec:rect_dyn_timeshift}
To mitigate the over-squeezing effect discussed above, the dynamic time-shifting scheme must be rectified. The most straightforward remedy is to manually retune the reference points $(x_1,y_1)$ and $(x_2,y_2)$ so as to flatten the slope of the linear interpolation, thereby slowing the growth of the shift magnitude $\mu(n_T)$ as the token count $n_T$ increases quadratically with resolution. However, directly flattening the slope leaves the shift insufficient for the low-resolution images (e.g., $512\times512$) that co-exist in mixed-resolution training. An alternative is to replace the linear map with a non-linear $\tilde{\mu}(\cdot)$, which in turn poses a non-trivial optimization problem. For instance, some works~\cite{wu2025omnigen2} adopt a logarithmic curve:
\[
\tilde{\mu}(n_T) = \alpha \cdot \log(n_T) + \beta,
\]
where one may take $\alpha=\tfrac{1}{2}$ and $\beta=-\log 20$; the factor $\alpha=\tfrac{1}{2}$ counteracts the quadratic growth of $n_T$ with respect to the image side length (\ie, $\sqrt{n_T}$). Nevertheless, fitting $(\alpha,\beta)$ is itself an optimization problem, and the resulting monotonic curve may still mismatch the optimal shift across resolutions. In short, choosing an optimal $\mu(n_T)$ is a complex problem to which we do not claim an optimal solution; we leave it to the research community and to future work.

Instead, we adopt a simple yet effective scheme that clamps the effective token count:
\[
\tilde{\mu}(n_T) = \mu\big(\min(n_T,\, n_{\text{cap}})\big) = \min\big(\mu(n_T),\, \mu(n_{\text{cap}})\big),
\]
where, \eg, $n_{\text{cap}}=4096$. This means that beyond the 1K-resolution regime, no additional time shifting is applied. To facilitate the analysis, we characterize the distribution of $t_{\text{shift}}$ through its quantile function. Letting $P_q := Q_{t_{\text{shift}}}(q;\mu(n_T),\sigma)$ denote the $q$-quantile (\ie, the value $P_q$ such that $\Pr[t_{\text{shift}} \le P_q] = q$), we have:
\[
P_q = Q_{t_{\text{shift}}}\!\left(q ;\, \mu(n_T), \sigma\right) = \mathrm{Sigmoid}\!\left(\sigma \cdot \Phi^{-1}(q) - \mu(n_T)\right);
\quad q\in(0,1),~\sigma>0,~\Phi(x)=\int_{-\infty}^{x} \frac{1}{\sqrt{2\pi}}\, e^{-s^2/2}\, ds,
\]
where we set $\sigma=1$ by default. Based on this expression we plot Figure~\ref{fig:dyn_time_shift}. The family of curves $\{P_q\}$, whose rate and trend are governed by $\mu(n_T)$, could in principle be studied via mathematical analysis, which would be rather involved. Our piecewise construction sidesteps this complexity: \textbf{by substituting $\tilde{\mu}(n_T)$ for $\mu(n_T)$, we directly inherit the tuning effort validated at 1K resolution, seamlessly extend it to 2K training, and eliminate the over-squeezing artifacts}, namely slow convergence and noisy generations at early stage.

\begin{figure}[hbp]
    \centering
    \includegraphics[width=0.65\textwidth]{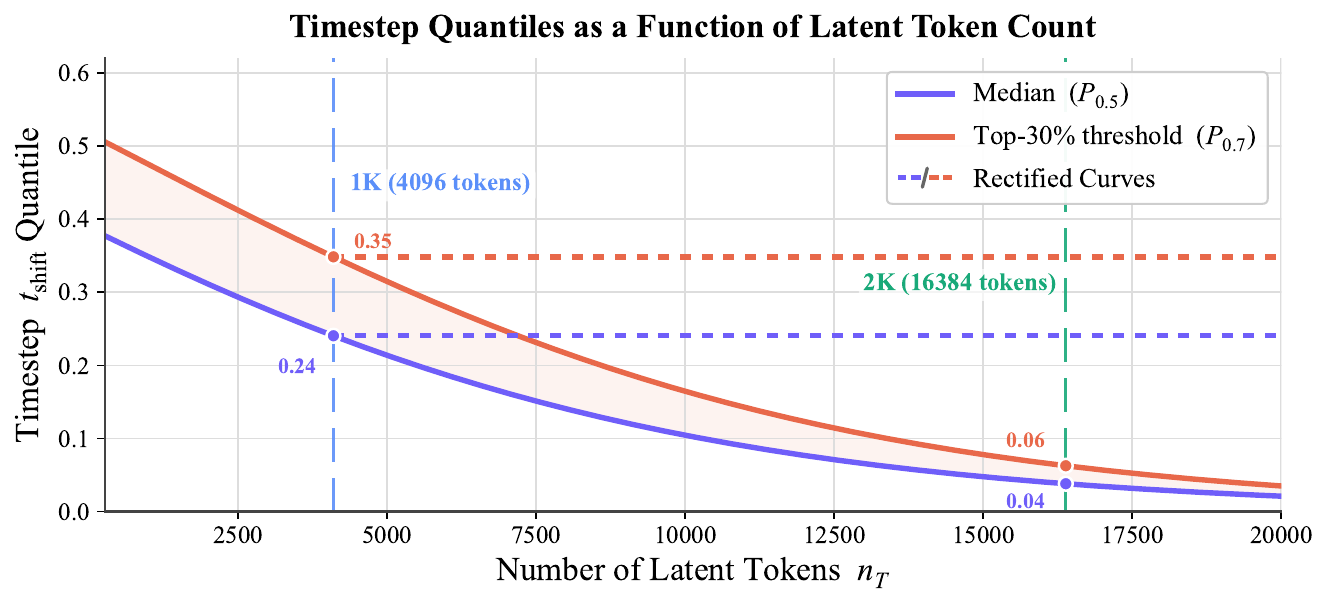}

    \caption{%
    \textbf{Timestep quantiles versus latent token count under dynamic time shifting.}
    For a given token count $n_T$, we plot two quantiles of the shifted timestep $t_{\mathrm{shift}}$,
    computed from $P_q=\mathrm{Sigmoid}\!\left(\sigma\,\Phi^{-1}(q)-\mu(n_T)\right)$ with $\sigma=1$, \( (x_1, y_1)=(256, 0.5) \) and \( (x_2, y_2)=(4096, 1.15) \):
    the median ($P_{0.5}$, purple) and the top-$30\%$ threshold ($P_{0.7}$, orange).
    \emph{Solid curves} use the original linear shift $\mu(n_T)$, which keeps growing with $n_T$ and
    drives both quantiles toward $0$ at high resolution (\eg\ the median collapses from $0.24$ at
    1K to $0.04$ at 2K), reflecting the over-squeezing effect.
    \emph{Dashed horizontal lines} show our rectified scheme, which clamps the effective token count at
    $n_{\mathrm{cap}}=4096$ (the 1K regime) so that the percentiles remain fixed at their 1K values
    beyond this point. The rectification thus inherits the well-tuned 1K shifting behavior and extends
    it to 2K training without exacerbating over-squeezing.
    }

    \label{fig:dyn_time_shift}
\end{figure}

\clearpage

\clearpage

\paragraph{2D structural awareness in DiT enriches photographic atmosphere.}
While standard Diffusion Transformers (DiTs) \cite{peebles2023dit} flatten spatial representations into 1D sequences and thereby neglect the inherent 2D nature of images, we find that explicitly preserving the 2D matrix structure of the latent space can improve the photographic and cinematic atmosphere of the outputs, yielding richer textures.

For conventional DiT, at each flow-matching (or denoising) step, the model output is treated as a single flattened 1D update direction in a high-dimensional space, along which the current latent is advanced toward a clean-image hidden state~\citep{sohldickstein2015diffusion,ho2020ddpm,song2021sde,song2022ddim,lipman2023flowmatching,liu2022rectifiedflow,lu2025dpmsolverpp}. Although convenient, this view discards a key inductive bias of the image modality: the DiT output is a tensor of shape $[B,C,H,W]$, whose $[H,W]$ dimensions form a meaningful grid aligned with the image plane rather than an arbitrary reshaping artifact. The mismatch is amplified by common normalization practices, which first collapse the 2D prediction into a 1D vector and then rescale its Euclidean length—altering only the magnitude while leaving the update direction and its spatial composition untouched. Even variants that additionally apply mean-centering remain blind to spatial organization, missing the opportunity to preserve informative variation and avoid overly smooth updates.

\begin{figure}[htbp]
    \centering

    \colorbox{tablegray}{\parbox{0.95\linewidth}{\small \textit{Prompt 1:} \begin{CJK*}{UTF8}{gbsn}画一只在森林中抓拍到的棕熊，棕熊旁边有一个路牌，写着``Watch Out!''。棕熊回头看这个路牌。图片要有真实感。\end{CJK*}}}\\[0.02in]

    \colorbox{tablegray}{\parbox{0.95\linewidth}{\small \textit{Prompt 2:} \begin{CJK*}{UTF8}{gbsn}生成一张街拍的照片，要有垃圾桶，红绿灯。画面中心有一个拾荒的老人，满脸沧桑。照片要有摄影质感，街拍质感。\end{CJK*}}}\\[0.02in]

    \colorbox{tablegray}{\parbox{0.95\linewidth}{\small \textit{Prompt 3:} \begin{CJK*}{UTF8}{gbsn}画一个篮球运动员在打篮球，胸前是8号球衣，球队是Boogu，场下有很多观众。这个照片是真实摄影风格。\end{CJK*}}}\\[0.05in]

    \colorbox{tablegray}{\parbox{0.95\linewidth}{\small \textit{Prompt 4:} \begin{CJK*}{UTF8}{gbsn}画一个中国男人，手里举一个牌子，上面写着``好好学习！''。要真实照片的风格。\end{CJK*}}}\\[0.02in]

    \includegraphics[width=0.24\textwidth]{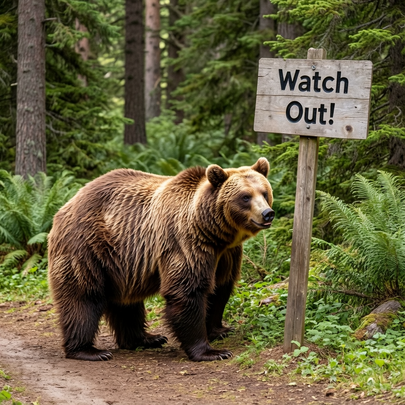}\hfill
    \includegraphics[width=0.24\textwidth]{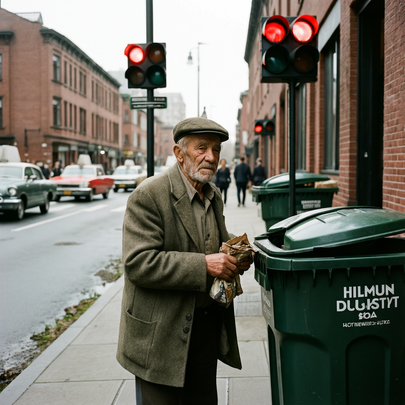}\hfill
    \includegraphics[width=0.24\textwidth]{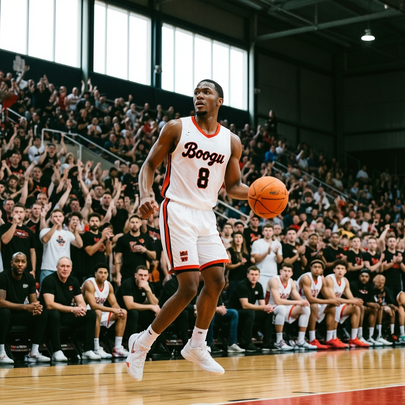}\hfill
    \includegraphics[width=0.24\textwidth]{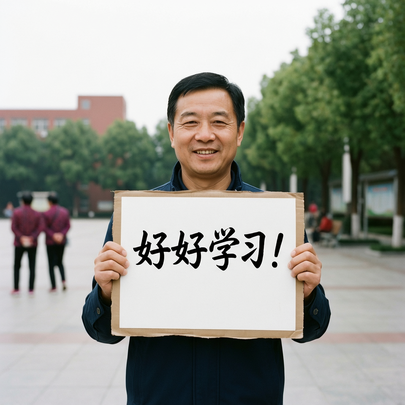}\\[0.05in]

    \includegraphics[width=0.24\textwidth]{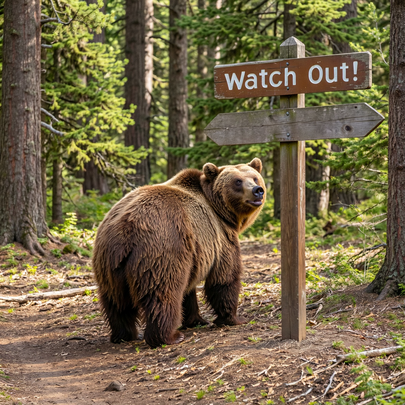}\hfill
    \includegraphics[width=0.24\textwidth]{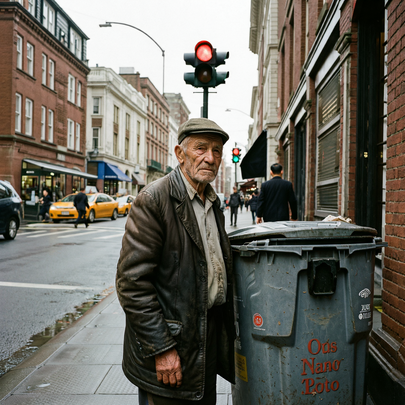}\hfill
    \includegraphics[width=0.24\textwidth]{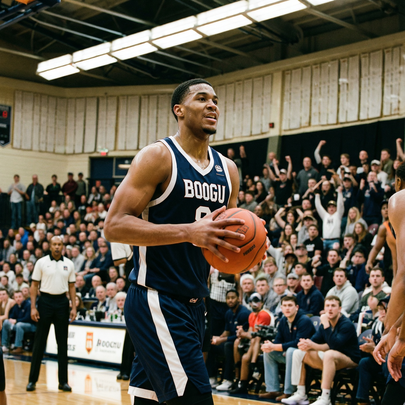}\hfill
    \includegraphics[width=0.24\textwidth]{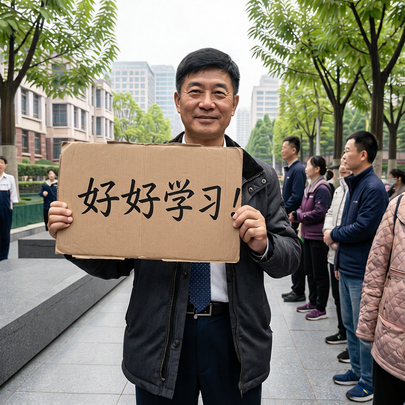}

\caption{\textbf{BOG produces more realistic photographic textures than standard CFG.} Each column is a pair generated from the same prompt (shown above): the top row uses standard CFG and the bottom row uses BOG. Across all four prompts, BOG yields stronger photographic realism and finer texture detail while preserving the content.}

    \label{fig:bog_cmp}
\end{figure}

Inspired by recent optimizers that respect matrix/tensor structure instead of reducing updates to flat vectors~\citep{gupta2018shampoo,jordan2024muon}, we propose an alternative classifier-free guidance~\citep{ho2022cfg} scheme for DiT, termed \textit{Boosted Orthogonal Guidance} (BOG), which applies a 2D normalization strategy directly to the matrix-structured prediction at each step. Figure~\ref{fig:bog_cmp} contrasts the generation results of BOG and conventional CFG. Notably, we believe that the matrix-structured view of DiT predictions opens up considerable room for further improvement, which we leave to future work and community exploration. A detailed discussion of BOG—including its motivation, usage and limitations—is provided in Appendix~\ref{sec:bog}.

\clearpage

\clearpage

\clearpage

\section{Conclusion, Limitations, and Future Directions}
\label{sec:conclude}
In this report, we present {{\model}-Image-0.1}, an open-source unified system for image generation and editing that jointly designs {understanding} and {generation} as one integrated whole. This design is motivated by a shift we observe in the field: generation is moving from Text-to-Image toward Requirement-to-Image, where the model must interpret complex intentions, implicit constraints, and cross-modal context rather than a single descriptive prompt. The training of Boogu-Image-0.1 is highly cost-effective, enabled by a carefully designed data construction strategy. On top of this, inference-time scaling with an agentic system powered by Deepseek-V4-Flash that rewrites prompts and selects models further improves performance. Comprehensive evaluations demonstrate that Boogu-Image-0.1 has emerged as one of the top open-source models currently available, achieving performance near to leading closed-source models such as NanoBanana-Pro in certain domains. Despite this strong performance, it maintains economical training costs, requiring 208.62M unique images across its full training pipeline at a total cost of around \$400K.

While Boogu achieves competitive performance under limited resources, it still has several limitations. In terms of world knowledge, Boogu still trails leading closed-source systems on tasks demanding extensive common-sense and domain knowledge (\eg, artistic styles, landmarks, public figures, and commercial products), and this gap is hard to measure reliably, so we expect it to exceed what reported scores suggest. For text rendering, only Chinese and English are currently optimized. In terms of body structure, multi-person interactions, heavy occlusions, or atypical viewpoints can yield anatomical inconsistencies such as distorted hands and limbs. Because we adopt the open-source FLUX.1 VAE, its reconstruction error further bottlenecks fine-grained details like small faces, limbs, and text, which native 2K generation only partially alleviates.

With the open-source release of {\model}-Image-0.1, we aim to advance image generation toward systems that genuinely understand what users want. We believe the following directions are essential for the field to advance, and we call on the community to jointly explore them:
\begin{itemize}
    \item \textbf{Toward more transparent and faithful evaluation.} Static benchmarks have largely saturated and grown misleading, as models increasingly optimize for the benchmark itself, so high academic scores no longer guarantee real-world efficacy. Even a model that dominates standard benchmarks can deliver an inferior experience, since true utility lies in faithfully translating complex, multi-faceted intents into images (through multi-turn editability, identity-preserving consistency, and robust adherence to long-context prompts) rather than single-shot aesthetic over-fitting.

    \item \textbf{Overcoming the open-source data bottleneck.} Public datasets suffer from pervasive semantic noise and a lack of cognitive depth, imposing a strict ceiling on performance, and closing the gap to frontier models requires a shift toward high-fidelity data synthesis and expert-level curation.

    \item \textbf{Toward stronger agentic image generation.} Injecting understanding at inference time has proven effective, but current agentic pipelines remain shallow. We see substantial headroom in tighter reasoning-generation loops, self-verification, and tool use, moving generation from a passive mapping toward a system that actively plans, critiques, and refines its own outputs.

\end{itemize}
\clearpage

\bibliography{reference}

\begin{thebibliography}{148}
\providecommand{\natexlab}[1]{#1}
\providecommand{\url}[1]{\texttt{#1}}
\expandafter\ifx\csname urlstyle\endcsname\relax
  \providecommand{\doi}[1]{doi: #1}\else
  \providecommand{\doi}{doi: \begingroup \urlstyle{rm}\Url}\fi

\bibitem[are(2026)]{arena_ai_2026}
Arena ai: The official ai ranking \& llm leaderboard.
\newblock \url{https://arena.ai/}, 2026.

\bibitem[AI(2026)]{ideogram-4-2026}
Ideogram AI.
\newblock {Ideogram 4}.
\newblock \url{https://ideogram.ai/blog/ideogram-4.0/}, 2026.

\bibitem[Bai et~al.(2025{\natexlab{a}})Bai, Cai, Chen, Chen, Chen, Cheng, Deng, Ding, Gao, Ge, Ge, Guo, Huang, Huang, Huang, Hui, Jiang, Li, Li, Li, Li, Lin, Lin, Liu, Liu, Liu, Liu, Liu, Liu, Lu, Luo, Lv, Men, Meng, Ren, Ren, Song, Sun, Tang, Tu, Wan, Wang, Wang, Wang, Wang, Xie, Xu, Xu, Xu, Yang, Yang, Yang, Yang, Yu, Zhang, Zhang, Zhang, Zheng, Zhong, Zhou, Zhou, Zhou, Zhu, and Zhu]{bai2025qwen3vl}
Shuai Bai, Yuxuan Cai, Ruizhe Chen, Keqin Chen, Xionghui Chen, Zesen Cheng, Lianghao Deng, Wei Ding, Chang Gao, Chunjiang Ge, Wenbin Ge, Zhifang Guo, Qidong Huang, Jie Huang, Fei Huang, Binyuan Hui, Shutong Jiang, Zhaohai Li, Mingsheng Li, Mei Li, Kaixin Li, Zicheng Lin, Junyang Lin, Xuejing Liu, Jiawei Liu, Chenglong Liu, Yang Liu, Dayiheng Liu, Shixuan Liu, Dunjie Lu, Ruilin Luo, Chenxu Lv, Rui Men, Lingchen Meng, Xuancheng Ren, Xingzhang Ren, Sibo Song, Yuchong Sun, Jun Tang, Jianhong Tu, Jianqiang Wan, Peng Wang, Pengfei Wang, Qiuyue Wang, Yuxuan Wang, Tianbao Xie, Yiheng Xu, Haiyang Xu, Jin Xu, Zhibo Yang, Mingkun Yang, Jianxin Yang, An~Yang, Bowen Yu, Fei Zhang, Hang Zhang, Xi~Zhang, Bo~Zheng, Humen Zhong, Jingren Zhou, Fan Zhou, Jing Zhou, Yuanzhi Zhu, and Ke~Zhu.
\newblock Qwen3-vl technical report, 2025{\natexlab{a}}.
\newblock URL \url{https://arxiv.org/abs/2511.21631}.

\bibitem[Bai et~al.(2025{\natexlab{b}})Bai, Chen, Liu, Wang, Ge, Song, Dang, Wang, Wang, Tang, Zhong, Zhu, Yang, Li, Wan, Wang, Ding, Fu, Xu, Ye, Zhang, Xie, Cheng, Zhang, Yang, Xu, and Lin]{bai2025qwen25vl}
Shuai Bai, Keqin Chen, Xuejing Liu, Jialin Wang, Wenbin Ge, Sibo Song, Kai Dang, Peng Wang, Shijie Wang, Jun Tang, Humen Zhong, Yuanzhi Zhu, Mingkun Yang, Zhaohai Li, Jianqiang Wan, Pengfei Wang, Wei Ding, Zheren Fu, Yiheng Xu, Jiabo Ye, Xi~Zhang, Tianbao Xie, Zesen Cheng, Hang Zhang, Zhibo Yang, Haiyang Xu, and Junyang Lin.
\newblock Qwen2.5-vl technical report, 2025{\natexlab{b}}.
\newblock URL \url{https://arxiv.org/abs/2502.13923}.

\bibitem[bench authors(2023)]{srivastava2023beyond}
BIG bench authors.
\newblock Beyond the imitation game: Quantifying and extrapolating the capabilities of language models.
\newblock \emph{Transactions on Machine Learning Research}, 2023.
\newblock ISSN 2835-8856.
\newblock URL \url{https://openreview.net/forum?id=uyTL5Bvosj}.

\bibitem[Betker et~al.(2023)Betker, Goh, Jing, Brooks, Wang, Li, Ouyang, Zhuang, Lee, Guo, et~al.]{betker2023improving}
James Betker, Gabriel Goh, Li~Jing, Tim Brooks, Jianfeng Wang, Linjie Li, Long Ouyang, Juntang Zhuang, Joyce Lee, Yufei Guo, et~al.
\newblock Improving image generation with better captions.
\newblock \emph{Computer Science. https://cdn. openai. com/papers/dall-e-3. pdf}, 2\penalty0 (3):\penalty0 8, 2023.

\bibitem[{Black Forest Labs}(2024)]{flux2024blackforest}
{Black Forest Labs}.
\newblock Announcing {Black Forest Labs}: Black forest labs' inaugural model suite, {FLUX.1}, 2024.
\newblock URL \url{https://blackforestlabs.ai/announcing-black-forest-labs/}.
\newblock Accessed: 2026-04-25.

\bibitem[{Black Forest Labs}(2025{\natexlab{a}})]{bfl2025flux2}
{Black Forest Labs}.
\newblock {FLUX.2}: Analyzing and enhancing the latent space of {FLUX} -- representation comparison, 11 2025{\natexlab{a}}.
\newblock URL \url{https://bfl.ai/research/representation-comparison}.

\bibitem[{Black Forest Labs}(2025{\natexlab{b}})]{blackforestlabs2025flux2}
{Black Forest Labs}.
\newblock Flux.2: Production-grade ai image generation and editing model with 4mp photorealistic output and multi-reference control.
\newblock \url{https://bfl.ai/models/flux-2}, 2025{\natexlab{b}}.

\bibitem[Brock et~al.(2019)Brock, Donahue, and Simonyan]{brock2018large}
Andrew Brock, Jeff Donahue, and Karen Simonyan.
\newblock Large scale {GAN} training for high fidelity natural image synthesis.
\newblock In \emph{International Conference on Learning Representations}, 2019.
\newblock URL \url{https://openreview.net/forum?id=B1xsqj09Fm}.

\bibitem[Brooks et~al.(2023)Brooks, Holynski, and Efros]{brooks2023instructpix2pix}
Tim Brooks, Aleksander Holynski, and Alexei~A Efros.
\newblock Instructpix2pix: Learning to follow image editing instructions.
\newblock In \emph{Proceedings of the IEEE/CVF conference on computer vision and pattern recognition}, pp.\  18392--18402, 2023.

\bibitem[Brooks et~al.(2024)Brooks, Peebles, Holmes, DePue, Guo, Jing, Schnurr, Taylor, Luhman, Luhman, et~al.]{brooks2024video}
Tim Brooks, Bill Peebles, Connor Holmes, Will DePue, Yufei Guo, Leo Jing, David Schnurr, Joe Taylor, Troy Luhman, Eric Luhman, et~al.
\newblock Video generation models as world simulators.
\newblock \emph{OpenAI Blog}, 1\penalty0 (8):\penalty0 1, 2024.

\bibitem[Byeon et~al.(2022)Byeon, Park, Kim, Lee, Baek, and Kim]{kakaobrain2022coyo}
Minwoo Byeon, Beomhee Park, Haecheon Kim, Sungjun Lee, Woonhyuk Baek, and Saehoon Kim.
\newblock Coyo-700m: Image-text pair dataset.
\newblock \url{https://github.com/kakaobrain/coyo-dataset}, 2022.

\bibitem[{ByteDance Seed Team}(2026{\natexlab{a}})]{seedream4d5}
{ByteDance Seed Team}.
\newblock Seedream 4.5.
\newblock \url{https://seed.bytedance.com/en/seedream4_5}, 2026{\natexlab{a}}.

\bibitem[{ByteDance Seed Team}(2026{\natexlab{b}})]{seedream5lite}
{ByteDance Seed Team}.
\newblock Seedream 5.0 lite.
\newblock \url{https://seed.bytedance.com/en/blog/deeper-thinking-more-accurate-generation-introducing-seedream-5-0-lite}, 2026{\natexlab{b}}.

\bibitem[Cai et~al.(2025)Cai, Yang, and Hu]{cai2025mm}
Huanqia Cai, Yijun Yang, and Winston Hu.
\newblock Mm-iq: Benchmarking human-like abstraction and reasoning in multimodal models.
\newblock \emph{arXiv preprint arXiv:2502.00698}, 2025.

\bibitem[Cai et~al.(2026)Cai, Chen, Gao, Gong, Li, Mei, Pan, Peng, Qiu, Yao, Yu, Zhang, et~al.]{hidreamolimage}
Qi~Cai, Jingwen Chen, Chengmin Gao, Zijian Gong, Yehao Li, Tao Mei, Yingwei Pan, Yi~Peng, Zhaofan Qiu, Ting Yao, Kai Yu, Yiheng Zhang, et~al.
\newblock Hidream-o1-image: A natively unified image generative foundation model with pixel-level unified transformer.
\newblock \emph{arXiv preprint arXiv:2605.11061}, 2026.

\bibitem[Cao et~al.(2024)Cao, Wang, He, Yuan, Li, Chen, Liu, Zhao, et~al.]{cao2024rwku}
Pengfei Cao, Chenhao Wang, Zhitao He, Hongbang Yuan, Jiachun Li, Yubo Chen, Kang Liu, Jun Zhao, et~al.
\newblock Rwku: Benchmarking real-world knowledge unlearning for large language models.
\newblock \emph{Advances in Neural Information Processing Systems}, 37:\penalty0 98213--98263, 2024.

\bibitem[Cao et~al.(2025)Cao, Chen, Chen, Cheng, Cui, Deng, Dong, Gong, Gu, Gu, et~al.]{cao2025hunyuanimage}
Siyu Cao, Hangting Chen, Peng Chen, Yiji Cheng, Yutao Cui, Xinchi Deng, Ying Dong, Kipper Gong, Tianpeng Gu, Xiusen Gu, et~al.
\newblock Hunyuanimage 3.0 technical report.
\newblock \emph{arXiv preprint arXiv:2509.23951}, 2025.

\bibitem[Carlini et~al.(2023)Carlini, Hayes, Nasr, Jagielski, Sehwag, Tramer, Balle, Ippolito, and Wallace]{carlini2023extracting}
Nicolas Carlini, Jamie Hayes, Milad Nasr, Matthew Jagielski, Vikash Sehwag, Florian Tramer, Borja Balle, Daphne Ippolito, and Eric Wallace.
\newblock Extracting training data from diffusion models.
\newblock In \emph{32nd USENIX security symposium (USENIX Security 23)}, pp.\  5253--5270, 2023.

\bibitem[Chang et~al.(2025)Chang, Fang, Xing, Wu, Cheng, Wang, Zeng, Yu, and Chen]{chang2025oneigbenchomnidimensionalnuancedevaluation}
Jingjing Chang, Yixiao Fang, Peng Xing, Shuhan Wu, Wei Cheng, Rui Wang, Xianfang Zeng, Gang Yu, and Hai-Bao Chen.
\newblock Oneig-bench: Omni-dimensional nuanced evaluation for image generation, 2025.
\newblock URL \url{https://arxiv.org/abs/2506.07977}.

\bibitem[Chen et~al.(2026)Chen, Wei, Wan, Chen, Zhang, Zhao, Zhang, Yue, Liang, Guo, et~al.]{chen2026lens}
Dong Chen, Fangyun Wei, Ziyu Wan, Dongdong Chen, Jiawei Zhang, Jinjing Zhao, Sirui Zhang, Yang Yue, Zhiyang Liang, Baining Guo, et~al.
\newblock Lens: Rethinking training efficiency for foundational text-to-image models.
\newblock \emph{arXiv preprint arXiv:2605.21573}, 2026.

\bibitem[Chen et~al.(2025{\natexlab{a}})Chen, Xue, Xu, Pan, Yang, Qin, Yan, Zhou, Chen, Huang, Zhou, Li, Savarese, Xiong, and Xu]{chen2025blip3onextfrontiernativeimage}
Jiuhai Chen, Le~Xue, Zhiyang Xu, Xichen Pan, Shusheng Yang, Can Qin, An~Yan, Honglu Zhou, Zeyuan Chen, Lifu Huang, Tianyi Zhou, Junnan Li, Silvio Savarese, Caiming Xiong, and Ran Xu.
\newblock Blip3o-next: Next frontier of native image generation, 2025{\natexlab{a}}.
\newblock URL \url{https://arxiv.org/abs/2510.15857}.

\bibitem[Chen et~al.(2023{\natexlab{a}})Chen, Yu, Ge, Yao, Xie, Wu, Wang, Kwok, Luo, Lu, and Li]{chen2023pixartalpha}
Junsong Chen, Jincheng Yu, Chongjian Ge, Lewei Yao, Enze Xie, Yue Wu, Zhongdao Wang, James Kwok, Ping Luo, Huchuan Lu, and Zhenguo Li.
\newblock Pixart-$\alpha$: Fast training of diffusion transformer for photorealistic text-to-image synthesis, 2023{\natexlab{a}}.

\bibitem[Chen et~al.(2023{\natexlab{b}})Chen, Zaharia, and Zou]{chen2023frugalgptuselargelanguage}
Lingjiao Chen, Matei Zaharia, and James Zou.
\newblock Frugalgpt: How to use large language models while reducing cost and improving performance, 2023{\natexlab{b}}.
\newblock URL \url{https://arxiv.org/abs/2305.05176}.

\bibitem[Chen et~al.(2025{\natexlab{b}})Chen, Huang, Fan, Wang, Zhou, Guan, and Lin]{chen2025reproducible}
Ziliang Chen, Xin Huang, Xiaoxuan Fan, Keze Wang, Yuyu Zhou, Quanlong Guan, and Liang Lin.
\newblock Reproducible vision-language models meet concepts out of pre-training.
\newblock In \emph{Proceedings of the Computer Vision and Pattern Recognition Conference}, pp.\  14701--14711, 2025{\natexlab{b}}.

\bibitem[Chiang et~al.(2024)Chiang, Zheng, Sheng, Angelopoulos, Li, Li, Zhang, Zhu, Jordan, Gonzalez, et~al.]{chiang2024chatbot}
Wei-Lin Chiang, Lianmin Zheng, Ying Sheng, Anastasios~Nikolas Angelopoulos, Tianle Li, Dacheng Li, Hao Zhang, Banghua Zhu, Michael Jordan, Joseph~E Gonzalez, et~al.
\newblock Chatbot arena: An open platform for evaluating llms by human preference.
\newblock \emph{arXiv preprint arXiv:2403.04132}, 2024.

\bibitem[Chrystal \& Mizen(2003)Chrystal and Mizen]{chrystal2003goodhart}
K~Alec Chrystal and Paul~D Mizen.
\newblock Goodhart's law: its origins, meaning and implications for monetary policy.
\newblock \emph{Central banking, monetary theory and practice: Essays in honour of Charles Goodhart}, 1:\penalty0 221--243, 2003.

\bibitem[Dehghani et~al.(2023)Dehghani, Djolonga, Mustafa, Padlewski, Heek, Gilmer, Steiner, Caron, Geirhos, Alabdulmohsin, Jenatton, Beyer, Tschannen, Arnab, Wang, Riquelme, Minderer, Puigcerver, Evci, Kumar, van Steenkiste, Elsayed, Mahendran, Yu, Oliver, Huot, Bastings, Collier, Gritsenko, Birodkar, Vasconcelos, Tay, Mensink, Kolesnikov, Pavetić, Tran, Kipf, Lučić, Zhai, Keysers, Harmsen, and Houlsby]{dehghani2023scalingvisiontransformers22}
Mostafa Dehghani, Josip Djolonga, Basil Mustafa, Piotr Padlewski, Jonathan Heek, Justin Gilmer, Andreas Steiner, Mathilde Caron, Robert Geirhos, Ibrahim Alabdulmohsin, Rodolphe Jenatton, Lucas Beyer, Michael Tschannen, Anurag Arnab, Xiao Wang, Carlos Riquelme, Matthias Minderer, Joan Puigcerver, Utku Evci, Manoj Kumar, Sjoerd van Steenkiste, Gamaleldin~F. Elsayed, Aravindh Mahendran, Fisher Yu, Avital Oliver, Fantine Huot, Jasmijn Bastings, Mark~Patrick Collier, Alexey Gritsenko, Vighnesh Birodkar, Cristina Vasconcelos, Yi~Tay, Thomas Mensink, Alexander Kolesnikov, Filip Pavetić, Dustin Tran, Thomas Kipf, Mario Lučić, Xiaohua Zhai, Daniel Keysers, Jeremiah Harmsen, and Neil Houlsby.
\newblock Scaling vision transformers to 22 billion parameters, 2023.
\newblock URL \url{https://arxiv.org/abs/2302.05442}.

\bibitem[Deng et~al.(2009)Deng, Dong, Socher, Li, Li, and Fei-Fei]{deng2009imagenet}
Jia Deng, Wei Dong, Richard Socher, Li-Jia Li, Kai Li, and Li~Fei-Fei.
\newblock Imagenet: A large-scale hierarchical image database.
\newblock In \emph{2009 IEEE conference on computer vision and pattern recognition}, pp.\  248--255. Ieee, 2009.

\bibitem[Ding et~al.(2024)Ding, Mallick, Wang, Sim, Mukherjee, R{\"u}hle, Lakshmanan, and Awadallah]{ding2024hybrid}
Dujian Ding, Ankur Mallick, Chi Wang, Robert Sim, Subhabrata Mukherjee, Victor R{\"u}hle, Laks Lakshmanan, and Ahmed~H Awadallah.
\newblock Hybrid llm: Cost-efficient and quality-aware query routing.
\newblock In \emph{International Conference on Learning Representations}, volume 2024, pp.\  41348--41366, 2024.

\bibitem[Ding et~al.(2021)Ding, Yang, Hong, Zheng, Zhou, Yin, Lin, Zou, Shao, Yang, and Tang]{ming2021cogview}
Ming Ding, Zhuoyi Yang, Wenyi Hong, Wendi Zheng, Chang Zhou, Da~Yin, Junyang Lin, Xu~Zou, Zhou Shao, Hongxia Yang, and Jie Tang.
\newblock Cogview: Mastering text-to-image generation via transformers.
\newblock In M.~Ranzato, A.~Beygelzimer, Y.~Dauphin, P.S. Liang, and J.~Wortman Vaughan (eds.), \emph{Advances in Neural Information Processing Systems}, volume~34, pp.\  19822--19835. Curran Associates, Inc., 2021.
\newblock URL \url{https://proceedings.neurips.cc/paper_files/paper/2021/file/a4d92e2cd541fca87e4620aba658316d-Paper.pdf}.

\bibitem[Dodge et~al.(2021)Dodge, Sap, Marasovi{\'c}, Agnew, Ilharco, Groeneveld, Mitchell, and Gardner]{dodge2021documenting}
Jesse Dodge, Maarten Sap, Ana Marasovi{\'c}, William Agnew, Gabriel Ilharco, Dirk Groeneveld, Margaret Mitchell, and Matt Gardner.
\newblock Documenting large webtext corpora: A case study on the colossal clean crawled corpus.
\newblock In \emph{Proceedings of the 2021 conference on empirical methods in natural language processing}, pp.\  1286--1305, 2021.

\bibitem[Dosovitskiy et~al.(2021)Dosovitskiy, Beyer, Kolesnikov, Weissenborn, Zhai, Unterthiner, Dehghani, Minderer, Heigold, Gelly, Uszkoreit, and Houlsby]{dosovitskiy2020image}
Alexey Dosovitskiy, Lucas Beyer, Alexander Kolesnikov, Dirk Weissenborn, Xiaohua Zhai, Thomas Unterthiner, Mostafa Dehghani, Matthias Minderer, Georg Heigold, Sylvain Gelly, Jakob Uszkoreit, and Neil Houlsby.
\newblock An image is worth 16x16 words: Transformers for image recognition at scale.
\newblock In \emph{International Conference on Learning Representations (ICLR)}, 2021.
\newblock URL \url{https://openreview.net/forum?id=YicbFdNTTy}.

\bibitem[Du et~al.(2025)Du, Chen, Chen, Gao, Chen, Jiang, Yang, and Tai]{du2025textcrafter}
Nikai Du, Zhennan Chen, Zhizhou Chen, Shan Gao, Xi~Chen, Zhengkai Jiang, Jian Yang, and Ying Tai.
\newblock Textcrafter: Accurately rendering multiple texts in complex visual scenes.
\newblock \emph{arXiv e-prints}, pp.\  arXiv--2503, 2025.

\bibitem[Esser et~al.(2024)Esser, Kulal, Blattmann, Entezari, M\"{u}ller, Saini, Levi, Lorenz, Sauer, Boesel, Podell, Dockhorn, English, and Rombach]{esser2024scaling}
Patrick Esser, Sumith Kulal, Andreas Blattmann, Rahim Entezari, Jonas M\"{u}ller, Harry Saini, Yam Levi, Dominik Lorenz, Axel Sauer, Frederic Boesel, Dustin Podell, Tim Dockhorn, Zion English, and Robin Rombach.
\newblock Scaling rectified flow transformers for high-resolution image synthesis.
\newblock In \emph{Proceedings of the 41st International Conference on Machine Learning}, ICML'24. JMLR.org, 2024.

\bibitem[Gadre et~al.(2023)Gadre, Ilharco, Fang, Hayase, Smyrnis, Nguyen, Marten, Wortsman, Ghosh, Zhang, Orgad, Entezari, Daras, Pratt, Ramanujan, Bitton, Marathe, Mussmann, Vencu, Cherti, Krishna, Koh, Saukh, Ratner, Song, Hajishirzi, Farhadi, Beaumont, Oh, Dimakis, Jitsev, Carmon, Shankar, and Schmidt]{gadredatacomp2023}
Samir~Yitzhak Gadre, Gabriel Ilharco, Alex Fang, Jonathan Hayase, Georgios Smyrnis, Thao Nguyen, Ryan Marten, Mitchell Wortsman, Dhruba Ghosh, Jieyu Zhang, Eyal Orgad, Rahim Entezari, Giannis Daras, Sarah Pratt, Vivek Ramanujan, Yonatan Bitton, Kalyani Marathe, Stephen Mussmann, Richard Vencu, Mehdi Cherti, Ranjay Krishna, Pang~Wei Koh, Olga Saukh, Alexander Ratner, Shuran Song, Hannaneh Hajishirzi, Ali Farhadi, Romain Beaumont, Sewoong Oh, Alex Dimakis, Jenia Jitsev, Yair Carmon, Vaishaal Shankar, and Ludwig Schmidt.
\newblock Datacomp: in search of the next generation of multimodal datasets.
\newblock In \emph{Proceedings of the 37th International Conference on Neural Information Processing Systems}, NIPS '23, Red Hook, NY, USA, 2023. Curran Associates Inc.

\bibitem[Gao et~al.(2025{\natexlab{a}})Gao, Zhuo, Liu, Du, Luo, Qiu, Zhang, Huang, Geng, Zhang, Xie, Shao, Jiang, Yang, Ye, He, He, He, Qiao, and Li]{gao2024luminat2x}
Peng Gao, Le~Zhuo, Dongyang Liu, Ruoyi Du, Xu~Luo, Longtian Qiu, Yuhang Zhang, Rongjie Huang, Shijie Geng, Renrui Zhang, Junlin Xie, Wenqi Shao, Zhengkai Jiang, Tianshuo Yang, Weicai Ye, Tong He, Jingwen He, Junjun He, Yu~Qiao, and Hongsheng Li.
\newblock Lumina-t2x: Scalable flow-based large diffusion transformer for flexible resolution generation.
\newblock In \emph{The Thirteenth International Conference on Learning Representations}, 2025{\natexlab{a}}.
\newblock URL \url{https://openreview.net/forum?id=EbWf36quzd}.

\bibitem[Gao et~al.(2025{\natexlab{b}})Gao, Gong, Guo, Hou, Lai, Li, Li, Lian, Liao, Liu, Liu, Shi, Sun, Tian, Tian, Wang, Wang, Wang, Wang, Wang, Wu, Wu, Xia, Xiao, Zhai, Zhang, Zhang, Zhang, Zhao, Yang, and Huang]{gao2025seedream3}
Yu~Gao, Lixue Gong, Qiushan Guo, Xiaoxia Hou, Zhichao Lai, Fanshi Li, Liang Li, Xiaochen Lian, Chao Liao, Liyang Liu, Wei Liu, Yichun Shi, Shiqi Sun, Yu~Tian, Zhi Tian, Peng Wang, Rui Wang, Xuanda Wang, Xun Wang, Ye~Wang, Guofeng Wu, Jie Wu, Xin Xia, Xuefeng Xiao, Zhonghua Zhai, Xinyu Zhang, Qi~Zhang, Yuwei Zhang, Shijia Zhao, Jianchao Yang, and Weilin Huang.
\newblock Seedream 3.0 technical report, 2025{\natexlab{b}}.
\newblock URL \url{https://arxiv.org/abs/2504.11346}.

\bibitem[Gemini~Team(2025)]{comanici2025gemini25pushingfrontier}
Google Gemini~Team.
\newblock Gemini 2.5: Pushing the frontier with advanced reasoning, multimodality, long context, and next generation agentic capabilities, 2025.
\newblock URL \url{https://arxiv.org/abs/2507.06261}.

\bibitem[Geng et~al.(2025)Geng, Wang, Ma, Li, Rao, Gu, Zhong, Lu, Hu, Zhang, et~al.]{geng2025x}
Zigang Geng, Yibing Wang, Yeyao Ma, Chen Li, Yongming Rao, Shuyang Gu, Zhao Zhong, Qinglin Lu, Han Hu, Xiaosong Zhang, et~al.
\newblock X-omni: Reinforcement learning makes discrete autoregressive image generative models great again.
\newblock \emph{arXiv preprint arXiv:2507.22058}, 2025.

\bibitem[Ghosh et~al.(2023)Ghosh, Hajishirzi, and Schmidt]{ghosh2023geneval}
Dhruba Ghosh, Hannaneh Hajishirzi, and Ludwig Schmidt.
\newblock Geneval: An object-focused framework for evaluating text-to-image alignment.
\newblock \emph{Advances in Neural Information Processing Systems}, 36:\penalty0 52132--52152, 2023.

\bibitem[{Google}(2025)]{google2025nanobananapro}
{Google}.
\newblock Introducing nano banana pro, 2025.
\newblock URL \url{https://blog.google/innovation-and-ai/products/nano-banana-pro/}.

\bibitem[{Google}(2026)]{google2026nanobanana2}
{Google}.
\newblock Nano banana 2: Combining pro capabilities with lightning-fast speed, 2026.
\newblock URL \url{https://blog.google/innovation-and-ai/technology/ai/nano-banana-2/}.

\bibitem[{Google DeepMind}(2025)]{googleimagen4}
{Google DeepMind}.
\newblock Imagen 4.0.
\newblock \url{https://deepmind.google/models/imagen/}, 2025.

\bibitem[{Google DeepMind}(2026)]{deepmind2026gemini31pro}
{Google DeepMind}.
\newblock Gemini 3.1 pro - model card.
\newblock \url{https://deepmind.google/models/model-cards/gemini-3-1-pro/}, February 2026.
\newblock Accessed: 2026-05-05.

\bibitem[Goyal et~al.(2018)Goyal, Dollár, Girshick, Noordhuis, Wesolowski, Kyrola, Tulloch, Jia, and He]{goyal2018wwarmup1}
Priya Goyal, Piotr Dollár, Ross Girshick, Pieter Noordhuis, Lukasz Wesolowski, Aapo Kyrola, Andrew Tulloch, Yangqing Jia, and Kaiming He.
\newblock Accurate, large minibatch sgd: Training imagenet in 1 hour, 2018.
\newblock URL \url{https://arxiv.org/abs/1706.02677}.

\bibitem[Gugger et~al.(2022)Gugger, Debut, Wolf, Schmid, Mueller, Mangrulkar, Sun, and Bossan]{gugger2022accelerate}
Sylvain Gugger, Lysandre Debut, Thomas Wolf, Philipp Schmid, Zachary Mueller, Sourab Mangrulkar, Marc Sun, and Benjamin Bossan.
\newblock Accelerate: Training and inference at scale made simple, efficient and adaptable.
\newblock \url{https://github.com/huggingface/accelerate}, 2022.

\bibitem[Guo et~al.(2025)Guo, Xu, Zhang, Song, Peng, Deng, Dong, Nakayama, Geng, Wang, et~al.]{guo2025r}
Meng-Hao Guo, Jiajun Xu, Yi~Zhang, Jiaxi Song, Haoyang Peng, Yi-Xuan Deng, Xinzhi Dong, Kiyohiro Nakayama, Zhengyang Geng, Chen Wang, et~al.
\newblock R-bench: Graduate-level multi-disciplinary benchmarks for llm \& mllm complex reasoning evaluation.
\newblock \emph{arXiv preprint arXiv:2505.02018}, 2025.

\bibitem[Gupta et~al.(2018)Gupta, Koren, and Singer]{gupta2018shampoo}
Vineet Gupta, Tomer Koren, and Yoram Singer.
\newblock Shampoo: Preconditioned stochastic tensor optimization, 2018.
\newblock URL \url{https://arxiv.org/abs/1802.09568}.

\bibitem[Hao et~al.(2023)Hao, Chi, Dong, and Wei]{hao2023optimizing}
Yaru Hao, Zewen Chi, Li~Dong, and Furu Wei.
\newblock Optimizing prompts for text-to-image generation.
\newblock \emph{Advances in Neural Information Processing Systems}, 36:\penalty0 66923--66939, 2023.

\bibitem[He et~al.(2016)He, Zhang, Ren, and Sun]{he2016deep}
Kaiming He, Xiangyu Zhang, Shaoqing Ren, and Jian Sun.
\newblock Deep residual learning for image recognition.
\newblock In \emph{Proceedings of the IEEE conference on computer vision and pattern recognition}, pp.\  770--778, 2016.

\bibitem[He et~al.(2026)He, Huang, Qu, Li, Zhu, Cheng, and Yang]{he2026gems}
Zefeng He, Siyuan Huang, Xiaoye Qu, Yafu Li, Tong Zhu, Yu~Cheng, and Yang Yang.
\newblock Gems: Agent-native multimodal generation with memory and skills.
\newblock \emph{arXiv preprint arXiv:2603.28088}, 2026.

\bibitem[Hessel et~al.(2021)Hessel, Holtzman, Forbes, Le~Bras, and Choi]{hessel2021clipscore}
Jack Hessel, Ari Holtzman, Maxwell Forbes, Ronan Le~Bras, and Yejin Choi.
\newblock Clipscore: A reference-free evaluation metric for image captioning.
\newblock In \emph{Proceedings of the 2021 conference on empirical methods in natural language processing}, pp.\  7514--7528, 2021.

\bibitem[Ho \& Salimans(2021)Ho and Salimans]{ho2022cfg}
Jonathan Ho and Tim Salimans.
\newblock Classifier-free diffusion guidance.
\newblock In \emph{NeurIPS 2021 Workshop on Deep Generative Models and Downstream Applications}, 2021.
\newblock URL \url{https://openreview.net/forum?id=qw8AKxfYbI}.

\bibitem[Ho et~al.(2020)Ho, Jain, and Abbeel]{ho2020ddpm}
Jonathan Ho, Ajay Jain, and Pieter Abbeel.
\newblock Denoising diffusion probabilistic models.
\newblock In \emph{Proceedings of the 34th International Conference on Neural Information Processing Systems}, NIPS '20, Red Hook, NY, USA, 2020. Curran Associates Inc.
\newblock ISBN 9781713829546.

\bibitem[Hu et~al.(2024)Hu, Wang, Fang, Fu, Cheng, and Yu]{hu2024ella}
Xiwei Hu, Rui Wang, Yixiao Fang, Bin Fu, Pei Cheng, and Gang Yu.
\newblock Ella: Equip diffusion models with llm for enhanced semantic alignment.
\newblock \emph{arXiv preprint arXiv:2403.05135}, 2024.

\bibitem[Hu et~al.(2023)Hu, Liu, Kasai, Wang, Ostendorf, Krishna, and Smith]{hu2023tifa}
Yushi Hu, Benlin Liu, Jungo Kasai, Yizhong Wang, Mari Ostendorf, Ranjay Krishna, and Noah~A Smith.
\newblock Tifa: Accurate and interpretable text-to-image faithfulness evaluation with question answering.
\newblock In \emph{Proceedings of the IEEE/CVF International Conference on Computer Vision}, pp.\  20406--20417, 2023.

\bibitem[Huang et~al.(2023)Huang, Sun, Xie, Li, and Liu]{huang2023t2i}
Kaiyi Huang, Kaiyue Sun, Enze Xie, Zhenguo Li, and Xihui Liu.
\newblock T2i-compbench: A comprehensive benchmark for open-world compositional text-to-image generation.
\newblock \emph{Advances in Neural Information Processing Systems}, 36:\penalty0 78723--78747, 2023.

\bibitem[Huang et~al.(2026)Huang, Wu, Wang, Cao, Chen, Fidler, Ling, and Ren]{huang2026ape}
Zijian Huang, Jay~Zhangjie Wu, Zian Wang, Tianshi Cao, Jiasi Chen, Sanja Fidler, Huan Ling, and Xuanchi Ren.
\newblock Ape: Agentic prompt enhancer for image generation and editing.
\newblock \emph{arXiv preprint arXiv:2606.00204}, 2026.

\bibitem[Jia et~al.(2021)Jia, Yang, Xia, Chen, Parekh, Pham, Le, Sung, Li, and Duerig]{jia2021scaling}
Chao Jia, Yinfei Yang, Ye~Xia, Yi-Ting Chen, Zarana Parekh, Hieu Pham, Quoc Le, Yun-Hsuan Sung, Zhen Li, and Tom Duerig.
\newblock Scaling up visual and vision-language representation learning with noisy text supervision.
\newblock In \emph{International conference on machine learning}, pp.\  4904--4916. PMLR, 2021.

\bibitem[Jiang et~al.(2026)Jiang, Sun, Zeng, Yang, Zhang, Wu, Cheng, Yu, Yang, and Wen]{jiang2026geditbench}
Zhangqi Jiang, Zheng Sun, Xianfang Zeng, Yufeng Yang, Xuanyang Zhang, Yongliang Wu, Wei Cheng, Gang Yu, Xu~Yang, and Bihan Wen.
\newblock Geditbench v2: A human-aligned benchmark for general image editing.
\newblock \emph{arXiv preprint arXiv:2603.28547}, 2026.

\bibitem[Jordan et~al.(2024)Jordan, Jin, Boza, You, Cesista, Newhouse, and Bernstein]{jordan2024muon}
Keller Jordan, Yuchen Jin, Vlado Boza, Jiacheng You, Franz Cesista, Laker Newhouse, and Jeremy Bernstein.
\newblock Muon: An optimizer for hidden layers in neural networks, 2024.
\newblock URL \url{https://kellerjordan.github.io/posts/muon/}.

\bibitem[Kaplan et~al.(2020)Kaplan, McCandlish, Henighan, Brown, Chess, Child, Gray, Radford, Wu, and Amodei]{kaplan2020scaling}
Jared Kaplan, Sam McCandlish, Tom Henighan, Tom~B Brown, Benjamin Chess, Rewon Child, Scott Gray, Alec Radford, Jeffrey Wu, and Dario Amodei.
\newblock Scaling laws for neural language models.
\newblock \emph{arXiv preprint arXiv:2001.08361}, 2020.

\bibitem[Karras et~al.(2022)Karras, Aittala, Laine, and Aila]{2022karrasODE}
Tero Karras, Miika Aittala, Samuli Laine, and Timo Aila.
\newblock Elucidating the design space of diffusion-based generative models.
\newblock In \emph{Proceedings of the 36th International Conference on Neural Information Processing Systems}, NIPS '22, Red Hook, NY, USA, 2022. Curran Associates Inc.
\newblock ISBN 9781713871088.

\bibitem[Karras et~al.(2024)Karras, Aittala, Lehtinen, Hellsten, Aila, and Laine]{karras2024altercfg}
Tero Karras, Miika Aittala, Jaakko Lehtinen, Janne Hellsten, Timo Aila, and Samuli Laine.
\newblock Analyzing and improving the training dynamics of diffusion models, 2024.
\newblock URL \url{https://arxiv.org/abs/2312.02696}.

\bibitem[Kingma \& Welling(2013)Kingma and Welling]{kingma2013auto}
Diederik~P Kingma and Max Welling.
\newblock Auto-encoding variational bayes.
\newblock \emph{arXiv preprint arXiv:1312.6114}, 2013.

\bibitem[Kirstain et~al.(2023)Kirstain, Polyak, Singer, Matiana, Penna, and Levy]{kirstain2023pick}
Yuval Kirstain, Adam Polyak, Uriel Singer, Shahbuland Matiana, Joe Penna, and Omer Levy.
\newblock Pick-a-pic: An open dataset of user preferences for text-to-image generation.
\newblock \emph{Advances in neural information processing systems}, 36:\penalty0 36652--36663, 2023.

\bibitem[Krizhevsky et~al.(2012)Krizhevsky, Sutskever, and Hinton]{krizhevsky2012imagenet}
Alex Krizhevsky, Ilya Sutskever, and Geoffrey~E Hinton.
\newblock Imagenet classification with deep convolutional neural networks.
\newblock \emph{Advances in neural information processing systems}, 25, 2012.

\bibitem[{Kuaishou Technology}(2025)]{klingimage21}
{Kuaishou Technology}.
\newblock Kling-image-2.1.
\newblock \url{https://kling.ai/explore/kling_2.1_api}, 2025.

\bibitem[Kynk{\"a}{\"a}nniemi et~al.(2019)Kynk{\"a}{\"a}nniemi, Karras, Laine, Lehtinen, and Aila]{kynkaanniemi2019improved}
Tuomas Kynk{\"a}{\"a}nniemi, Tero Karras, Samuli Laine, Jaakko Lehtinen, and Timo Aila.
\newblock Improved precision and recall metric for assessing generative models.
\newblock \emph{Advances in neural information processing systems}, 32, 2019.

\bibitem[Lee et~al.(2026)Lee, Millon, Zhuo, Newton, Filatov, Devarinti, Zhong, Djordjevic, Menezes, Beddow, Ebbecke, Petrescu, Fahey, Saß, Gil, and Perez]{krea-2-2026}
Sangwu Lee, Erwann Millon, Le~Zhuo, Matthew Newton, Andrei Filatov, Naga Sai~Abhinay Devarinti, Dazhi Zhong, Avram Djordjevic, Gabriel Menezes, Will Beddow, Titus Ebbecke, Mihai Petrescu, Owen Fahey, Gian Saß, Felix Gil, and Victor Perez.
\newblock {Krea 2}.
\newblock \url{https://www.krea.ai/blog/krea-2-technical-report}, 2026.

\bibitem[Li et~al.(2026)Li, Hu, Qiao, Ba, Hong, Shen, Wang, Zhou, Kang, Shang, He, Wang, Li, Li, Zhang, Gao, Yan, Jiang, Tang, Yin, Wu, Xu, Chen, Chen, Shu, Zhang, Chen, Xu, Zhang, Wang, Liu, Zhou, Shi, Wang, Zhao, Wei, Qu, and Wu]{li2026qwenimagebenchgenerationcreationtexttoimage}
Niantong Li, Guangzheng Hu, Weixu Qiao, Ying Ba, Qichen Hong, Shijun Shen, Jinlin Wang, Fan Zhou, Jianye Kang, Xin Shang, Ziyi He, Wei Wang, Dalin Li, Jiahao Li, Jie Zhang, Kaiyuan Gao, Kun Yan, Lihan Jiang, Ningyuan Tang, Shengming Yin, Tianhe Wu, Xiao Xu, Xiaoyue Chen, Yuxiang Chen, Yan Shu, Yanran Zhang, Yilei Chen, Yixian Xu, Zekai Zhang, Zhendong Wang, Zihao Liu, Zikai Zhou, Hongzhu Shi, Yi~Wang, Bing Zhao, Hu~Wei, Lin Qu, and Chenfei Wu.
\newblock Qwen-image-bench: From generation to creation in text-to-image evaluation, 2026.
\newblock URL \url{https://arxiv.org/abs/2605.28091}.

\bibitem[Li et~al.(2025{\natexlab{a}})Li, Chen, Wang, Wang, Ye, Jin, Chen, He, Gao, Cui, Yu, Tian, Zhou, Xu, Wang, Wei, Li, Zhang, Zhang, Cai, Wen, Yan, Chu, Wang, Dou, Tian, Zhu, Lu, Chen, He, Lu, Wang, Wang, Lin, Qiao, Shi, He, and Dai]{li2025omnicorpus}
Qingyun Li, Zhe Chen, Weiyun Wang, Wenhai Wang, Shenglong Ye, Zhenjiang Jin, Guanzhou Chen, Yinan He, Zhangwei Gao, Erfei Cui, Jiashuo Yu, Hao Tian, Jiasheng Zhou, Chao Xu, Bin Wang, Xingjian Wei, Wei Li, Wenjian Zhang, Bo~Zhang, Pinlong Cai, Licheng Wen, Xiangchao Yan, Pei Chu, Yi~Wang, Min Dou, Changyao Tian, Xizhou Zhu, Lewei Lu, Yushi Chen, Junjun He, Tong Lu, Yali Wang, Limin Wang, Dahua Lin, Yu~Qiao, Botian Shi, Conghui He, and Jifeng Dai.
\newblock Omnicorpus: A unified multimodal corpus of 10 billion-level images interleaved with text.
\newblock In \emph{The Thirteenth International Conference on Learning Representations}, 2025{\natexlab{a}}.
\newblock URL \url{https://openreview.net/forum?id=kwqhn2VuG4}.

\bibitem[Li et~al.(2025{\natexlab{b}})Li, Kallidromitis, Gokul, Koneru, Kato, Kozuka, and Grover]{li2025reflect}
Shufan Li, Konstantinos Kallidromitis, Akash Gokul, Arsh Koneru, Yusuke Kato, Kazuki Kozuka, and Aditya Grover.
\newblock Reflect-dit: Inference-time scaling for text-to-image diffusion transformers via in-context reflection.
\newblock In \emph{Proceedings of the IEEE/CVF International Conference on Computer Vision}, pp.\  15657--15668, 2025{\natexlab{b}}.

\bibitem[Liang et~al.(2022)Liang, Bommasani, Lee, Tsipras, Soylu, Yasunaga, Zhang, Narayanan, Wu, Kumar, et~al.]{liang2022holistic}
Percy Liang, Rishi Bommasani, Tony Lee, Dimitris Tsipras, Dilara Soylu, Michihiro Yasunaga, Yian Zhang, Deepak Narayanan, Yuhuai Wu, Ananya Kumar, et~al.
\newblock Holistic evaluation of language models.
\newblock \emph{arXiv preprint arXiv:2211.09110}, 2022.

\bibitem[Liang et~al.(2024)Liang, He, Yang, and Dai]{liang2024scaling_dit}
Zhengyang Liang, Hao He, Ceyuan Yang, and Bo~Dai.
\newblock Scaling laws for diffusion transformers.
\newblock \emph{arXiv preprint arXiv:2410.08184}, 2024.

\bibitem[Lin et~al.(2026)Lin, Dong, Shi, Lei, Zhang, Liu, and Liu]{lin2026aegis}
Jintao Lin, Bowen Dong, Weikang Shi, Chenyang Lei, Suiyun Zhang, Rui Liu, and Xihui Liu.
\newblock Aegis: Exploring the limit of world knowledge capabilities for unified mulitmodal models.
\newblock \emph{arXiv preprint arXiv:2601.00561}, 2026.

\bibitem[Lipman et~al.(2023)Lipman, Chen, Ben-Hamu, Nickel, and Le]{lipman2023flowmatching}
Yaron Lipman, Ricky T.~Q. Chen, Heli Ben-Hamu, Maximilian Nickel, and Matt Le.
\newblock Flow matching for generative modeling, 2023.
\newblock URL \url{https://arxiv.org/abs/2210.02747}.

\bibitem[Liu et~al.(2023)Liu, Li, Wu, and Lee]{liu2024visual}
Haotian Liu, Chunyuan Li, Qingyang Wu, and Yong~Jae Lee.
\newblock Visual instruction tuning.
\newblock \emph{Advances in neural information processing systems}, 36:\penalty0 34892--34916, 2023.

\bibitem[Liu et~al.(2026)Liu, Feng, Zou, Qian, Zhu, Xia, Dong, Lin, Xiong, Chen, et~al.]{liu2026ernie}
Jiaxiang Liu, Zhida Feng, Pengyu Zou, Zhenyu Qian, Tianrui Zhu, Jun Xia, Yuehu Dong, Yanzheng Lin, Honglin Xiong, Anqi Chen, et~al.
\newblock Ernie-image technical report.
\newblock \emph{arXiv preprint arXiv:2605.25347}, 2026.

\bibitem[Liu et~al.(2025{\natexlab{a}})Liu, Liu, Liang, Li, Liu, Wang, Wan, Zhang, and Ouyang]{liu2025flowgrpotrainingflowmatching}
Jie Liu, Gongye Liu, Jiajun Liang, Yangguang Li, Jiaheng Liu, Xintao Wang, Pengfei Wan, Di~Zhang, and Wanli Ouyang.
\newblock Flow-grpo: Training flow matching models via online rl, 2025{\natexlab{a}}.
\newblock URL \url{https://arxiv.org/abs/2505.05470}.

\bibitem[Liu et~al.(2025{\natexlab{b}})Liu, Han, Xing, Yin, Wang, Cheng, Liao, Wang, Fu, Han, et~al.]{liu2025step1x}
Shiyu Liu, Yucheng Han, Peng Xing, Fukun Yin, Rui Wang, Wei Cheng, Jiaqi Liao, Yingming Wang, Honghao Fu, Chunrui Han, et~al.
\newblock Step1x-edit: A practical framework for general image editing.
\newblock \emph{arXiv preprint arXiv:2504.17761}, 2025{\natexlab{b}}.

\bibitem[Liu et~al.(2022)Liu, Gong, and Liu]{liu2022rectifiedflow}
Xingchao Liu, Chengyue Gong, and Qiang Liu.
\newblock Flow straight and fast: Learning to generate and transfer data with rectified flow, 2022.
\newblock URL \url{https://arxiv.org/abs/2209.03003}.

\bibitem[Lu et~al.(2022)Lu, Zhou, Bao, Chen, Li, and Zhu]{lu2022dpmsolver}
Cheng Lu, Yuhao Zhou, Fan Bao, Jianfei Chen, Chongxuan Li, and Jun Zhu.
\newblock Dpm-solver: A fast ode solver for diffusion probabilistic model sampling in around 10 steps, 2022.
\newblock URL \url{https://arxiv.org/abs/2206.00927}.

\bibitem[Lu et~al.(2025)Lu, Zhou, Bao, Chen, Li, and Zhu]{lu2025dpmsolverpp}
Cheng Lu, Yuhao Zhou, Fan Bao, Jianfei Chen, Chongxuan Li, and Jun Zhu.
\newblock Dpm-solver++: Fast solver for guided sampling of diffusion probabilistic models.
\newblock \emph{Machine Intelligence Research}, 22\penalty0 (4):\penalty0 730–751, June 2025.
\newblock ISSN 2731-5398.
\newblock \doi{10.1007/s11633-025-1562-4}.
\newblock URL \url{http://dx.doi.org/10.1007/s11633-025-1562-4}.

\bibitem[Lu et~al.(2024)Lu, Li, Chen, Xu, Luo, Zhang, and Ye]{lu2024ovis}
Shiyin Lu, Yang Li, Qing-Guo Chen, Zhao Xu, Weihua Luo, Kaifu Zhang, and Han-Jia Ye.
\newblock Ovis: Structural embedding alignment for multimodal large language model.
\newblock \emph{arXiv preprint arXiv:2405.20797}, 2024.

\bibitem[Murphy(2023)]{pml2Book}
Kevin~P. Murphy.
\newblock \emph{Probabilistic Machine Learning: Advanced Topics}.
\newblock MIT Press, 2023.
\newblock URL \url{http://probml.github.io/book2}.

\bibitem[Neumann(1937)]{Neumann1937}
Johann~von Neumann.
\newblock Some matrix-inequalities and metrization of matric-space.
\newblock \emph{Tomsk. Univ. Rev.}, 1:\penalty0 286--300, 1937.
\newblock Reprinted in {\it John von Neumann: Collected Works} (A. H. Taub, ed.), Vol. IV, Pergamon Press, Oxford, 1962, pp. 205-218.

\bibitem[Nichol \& Dhariwal(2021)Nichol and Dhariwal]{nichol2021improved}
Alexander~Quinn Nichol and Prafulla Dhariwal.
\newblock Improved denoising diffusion probabilistic models.
\newblock In \emph{International conference on machine learning}, pp.\  8162--8171. PMLR, 2021.

\bibitem[Niu et~al.(2025)Niu, Ning, Zheng, Jin, Lin, Jin, Liao, Ning, Feng, Zhu, and Yuan]{niu2025wise}
Yuwei Niu, Munan Ning, Mengren Zheng, Weiyang Jin, Bin Lin, Peng Jin, Jiaqi Liao, Kunpeng Ning, Chaoran Feng, Bin Zhu, and Li~Yuan.
\newblock Wise: A world knowledge-informed semantic evaluation for text-to-image generation.
\newblock \emph{arXiv preprint arXiv:2503.07265}, 2025.

\bibitem[Ong et~al.(2024)Ong, Almahairi, Wu, Chiang, Wu, Gonzalez, Kadous, and Stoica]{ong2024routellm}
Isaac Ong, Amjad Almahairi, Vincent Wu, Wei-Lin Chiang, Tianhao Wu, Joseph~E Gonzalez, M~Waleed Kadous, and Ion Stoica.
\newblock Routellm: Learning to route llms with preference data.
\newblock \emph{arXiv preprint arXiv:2406.18665}, 2024.

\bibitem[{OpenAI}(2025)]{openai2025gptimage1}
{OpenAI}.
\newblock Gpt image 1.
\newblock \url{https://developers.openai.com/api/docs/models/gpt-image-1}, 2025.

\bibitem[{OpenAI}(2026)]{openai2026gptimage2}
{OpenAI}.
\newblock Introducing chatgpt images 2.0, 2026.
\newblock URL \url{https://openai.com/index/introducing-chatgpt-images-2-0/}.

\bibitem[Parashar et~al.(2024)Parashar, Lin, Liu, Dong, Li, Ramanan, Caverlee, and Kong]{parashar2024neglected}
Shubham Parashar, Zhiqiu Lin, Tian Liu, Xiangjue Dong, Yanan Li, Deva Ramanan, James Caverlee, and Shu Kong.
\newblock The neglected tails in vision-language models, 2024.
\newblock URL \url{https://arxiv.org/abs/2401.12425}.

\bibitem[Parmar et~al.(2022)Parmar, Zhang, and Zhu]{parmar2022aliased}
Gaurav Parmar, Richard Zhang, and Jun-Yan Zhu.
\newblock On aliased resizing and surprising subtleties in gan evaluation.
\newblock In \emph{Proceedings of the IEEE/CVF conference on computer vision and pattern recognition}, pp.\  11410--11420, 2022.

\bibitem[Peebles \& Xie(2023)Peebles and Xie]{peebles2023dit}
William Peebles and Saining Xie.
\newblock Scalable diffusion models with transformers, 2023.
\newblock URL \url{https://arxiv.org/abs/2212.09748}.

\bibitem[Podell et~al.(2024)Podell, English, Lacey, Blattmann, Dockhorn, M{\"u}ller, Penna, and Rombach]{podell2024sdxl}
Dustin Podell, Zion English, Kyle Lacey, Andreas Blattmann, Tim Dockhorn, Jonas M{\"u}ller, Joe Penna, and Robin Rombach.
\newblock Sdxl: Improving latent diffusion models for high-resolution image synthesis.
\newblock In \emph{International Conference on Learning Representations}, volume 2024, pp.\  1862--1874, 2024.

\bibitem[{Qwen Team, Alibaba Group}(2025{\natexlab{a}})]{qwenimage2512}
{Qwen Team, Alibaba Group}.
\newblock Qwen-image-2512: Finer details, greater realism.
\newblock \url{https://qwen.ai/blog?id=qwen-image-2512}, 2025{\natexlab{a}}.
\newblock Apache-2.0 License.

\bibitem[{Qwen Team, Alibaba Group}(2025{\natexlab{b}})]{qwenimageedit2509}
{Qwen Team, Alibaba Group}.
\newblock Qwen-image-edit-2509: Multi-image support, improved consistency.
\newblock \url{https://qwen.ai/blog?id=7a90090115ee193ce6a7f619522771dd9696dd93}, 2025{\natexlab{b}}.
\newblock Apache-2.0 License.

\bibitem[{Qwen Team, Alibaba Group}(2025{\natexlab{c}})]{qwenimageedit2511}
{Qwen Team, Alibaba Group}.
\newblock Qwen-image-2511.
\newblock \url{https://qwen.ai/blog?id=qwen-image-2512}, 2025{\natexlab{c}}.
\newblock Apache-2.0 License.

\bibitem[Radford et~al.(2021)Radford, Kim, Hallacy, Ramesh, Goh, Agarwal, Sastry, Askell, Mishkin, Clark, et~al.]{radford2021learning}
Alec Radford, Jong~Wook Kim, Chris Hallacy, Aditya Ramesh, Gabriel Goh, Sandhini Agarwal, Girish Sastry, Amanda Askell, Pamela Mishkin, Jack Clark, et~al.
\newblock Learning transferable visual models from natural language supervision.
\newblock In \emph{International conference on machine learning}, pp.\  8748--8763. PmLR, 2021.

\bibitem[Rajbhandari et~al.(2020)Rajbhandari, Rasley, Ruwase, and He]{rajbhandari2020zero}
Samyam Rajbhandari, Jeff Rasley, Olatunji Ruwase, and Yuxiong He.
\newblock Zero: Memory optimizations toward training trillion parameter models, 2020.
\newblock URL \url{https://arxiv.org/abs/1910.02054}.

\bibitem[Ramesh et~al.(2022)Ramesh, Dhariwal, Nichol, Chu, and Chen]{ramesh2022hierarchical}
Aditya Ramesh, Prafulla Dhariwal, Alex Nichol, Casey Chu, and Mark Chen.
\newblock Hierarchical text-conditional image generation with clip latents.
\newblock \emph{arXiv preprint arXiv:2204.06125}, 1\penalty0 (2):\penalty0 3, 2022.

\bibitem[Recht et~al.(2019)Recht, Roelofs, Schmidt, and Shankar]{recht2019imagenet}
Benjamin Recht, Rebecca Roelofs, Ludwig Schmidt, and Vaishaal Shankar.
\newblock Do imagenet classifiers generalize to imagenet?
\newblock In \emph{International conference on machine learning}, pp.\  5389--5400. PMLR, 2019.

\bibitem[Reiter(2018)]{reiter2018structured}
Ehud Reiter.
\newblock A structured review of the validity of bleu.
\newblock \emph{Computational Linguistics}, 44\penalty0 (3):\penalty0 393--401, 2018.

\bibitem[Rombach et~al.(2022)Rombach, Blattmann, Lorenz, Esser, and Ommer]{rombach2022high}
Robin Rombach, Andreas Blattmann, Dominik Lorenz, Patrick Esser, and Björn Ommer.
\newblock High-resolution image synthesis with latent diffusion models, 2022.
\newblock URL \url{https://arxiv.org/abs/2112.10752}.

\bibitem[Ronneberger et~al.(2015)Ronneberger, Fischer, and Brox]{ronneberger2015u}
Olaf Ronneberger, Philipp Fischer, and Thomas Brox.
\newblock U-net: Convolutional networks for biomedical image segmentation.
\newblock In \emph{International Conference on Medical Image Computing and Computer-Assisted Intervention}, pp.\  234--241. Springer, 2015.

\bibitem[Sadat et~al.(2025)Sadat, Hilliges, and Weber]{sadat2025apg}
Seyedmorteza Sadat, Otmar Hilliges, and Romann~M. Weber.
\newblock Eliminating oversaturation and artifacts of high guidance scales in diffusion models, 2025.
\newblock URL \url{https://arxiv.org/abs/2410.02416}.

\bibitem[Saharia et~al.(2022)Saharia, Chan, Saxena, Li, Whang, Denton, Ghasemipour, Gontijo~Lopes, Karagol~Ayan, Salimans, et~al.]{saharia2022photorealistic}
Chitwan Saharia, William Chan, Saurabh Saxena, Lala Li, Jay Whang, Emily~L Denton, Kamyar Ghasemipour, Raphael Gontijo~Lopes, Burcu Karagol~Ayan, Tim Salimans, et~al.
\newblock Photorealistic text-to-image diffusion models with deep language understanding.
\newblock \emph{Advances in neural information processing systems}, 35:\penalty0 36479--36494, 2022.

\bibitem[Schuhmann et~al.(2021)Schuhmann, Vencu, Beaumont, Kaczmarczyk, Mullis, Katta, Coombes, Jitsev, and Komatsuzaki]{schuhmann2021laion}
Christoph Schuhmann, Richard Vencu, Romain Beaumont, Robert Kaczmarczyk, Clayton Mullis, Aarush Katta, Theo Coombes, Jenia Jitsev, and Aran Komatsuzaki.
\newblock Laion-400m: Open dataset of clip-filtered 400 million image-text pairs.
\newblock \emph{arXiv preprint arXiv:2111.02114}, 2021.

\bibitem[Schuhmann et~al.(2022)Schuhmann, Beaumont, Vencu, Gordon, Wightman, Cherti, Coombes, Katta, Mullis, Wortsman, et~al.]{schuhmann2022laion}
Christoph Schuhmann, Romain Beaumont, Richard Vencu, Cade Gordon, Ross Wightman, Mehdi Cherti, Theo Coombes, Aarush Katta, Clayton Mullis, Mitchell Wortsman, et~al.
\newblock Laion-5b: An open large-scale dataset for training next generation image-text models.
\newblock \emph{Advances in neural information processing systems}, 35:\penalty0 25278--25294, 2022.

\bibitem[Seedream et~al.(2025)Seedream, :, Chen, Gao, Gong, Guo, Guo, Guo, Hou, Huang, Huang, Jian, Kuang, Lai, Li, Li, Lian, Liao, Liu, Liu, Lu, Luo, Ou, Shi, Shi, Sun, Tian, Tian, Wang, Wang, Wang, Wang, Wu, Wu, Wu, Wu, Xia, Xiao, Xu, Yan, Yang, Yang, Zhai, Zhang, Zhang, Zhang, Zhang, Zhang, Zhao, Zhao, and Zhu]{seedream2025seedream40nextgenerationmultimodal}
Team Seedream, :, Yunpeng Chen, Yu~Gao, Lixue Gong, Meng Guo, Qiushan Guo, Zhiyao Guo, Xiaoxia Hou, Weilin Huang, Yixuan Huang, Xiaowen Jian, Huafeng Kuang, Zhichao Lai, Fanshi Li, Liang Li, Xiaochen Lian, Chao Liao, Liyang Liu, Wei Liu, Yanzuo Lu, Zhengxiong Luo, Tongtong Ou, Guang Shi, Yichun Shi, Shiqi Sun, Yu~Tian, Zhi Tian, Peng Wang, Rui Wang, Xun Wang, Ye~Wang, Guofeng Wu, Jie Wu, Wenxu Wu, Yonghui Wu, Xin Xia, Xuefeng Xiao, Shuang Xu, Xin Yan, Ceyuan Yang, Jianchao Yang, Zhonghua Zhai, Chenlin Zhang, Heng Zhang, Qi~Zhang, Xinyu Zhang, Yuwei Zhang, Shijia Zhao, Wenliang Zhao, and Wenjia Zhu.
\newblock Seedream 4.0: Toward next-generation multimodal image generation, 2025.
\newblock URL \url{https://arxiv.org/abs/2509.20427}.

\bibitem[Singla et~al.(2024)Singla, Yue, Paul, Shirkavand, Jayawardhana, Ganjdanesh, Huang, Bhatele, Somepalli, and Goldstein]{singla2024pixelsproselargedataset}
Vasu Singla, Kaiyu Yue, Sukriti Paul, Reza Shirkavand, Mayuka Jayawardhana, Alireza Ganjdanesh, Heng Huang, Abhinav Bhatele, Gowthami Somepalli, and Tom Goldstein.
\newblock From pixels to prose: A large dataset of dense image captions, 2024.
\newblock URL \url{https://arxiv.org/abs/2406.10328}.

\bibitem[Sohl-Dickstein et~al.(2015)Sohl-Dickstein, Weiss, Maheswaranathan, and Ganguli]{sohldickstein2015diffusion}
Jascha Sohl-Dickstein, Eric~A. Weiss, Niru Maheswaranathan, and Surya Ganguli.
\newblock Deep unsupervised learning using nonequilibrium thermodynamics, 2015.
\newblock URL \url{https://arxiv.org/abs/1503.03585}.

\bibitem[Somepalli et~al.(2023)Somepalli, Singla, Goldblum, Geiping, and Goldstein]{somepalli2023diffusion}
Gowthami Somepalli, Vasu Singla, Micah Goldblum, Jonas Geiping, and Tom Goldstein.
\newblock Diffusion art or digital forgery? investigating data replication in diffusion models.
\newblock In \emph{Proceedings of the IEEE/CVF conference on computer vision and pattern recognition}, pp.\  6048--6058, 2023.

\bibitem[Song et~al.(2022)Song, Meng, and Ermon]{song2022ddim}
Jiaming Song, Chenlin Meng, and Stefano Ermon.
\newblock Denoising diffusion implicit models, 2022.
\newblock URL \url{https://arxiv.org/abs/2010.02502}.

\bibitem[Song et~al.(2026)Song, Li, Ma, Tang, Wang, Zhang, Yang, Xiao, Liu, Zhang, et~al.]{song2026joyai}
Lin Song, Wenbo Li, Guoqing Ma, Wei Tang, Bo~Wang, Yuan Zhang, Yijun Yang, Yicheng Xiao, Jianhui Liu, Yanbing Zhang, et~al.
\newblock Joyai-image: Awaking spatial intelligence in unified multimodal understanding and generation.
\newblock \emph{arXiv preprint arXiv:2605.04128}, 2026.

\bibitem[Song et~al.(2021)Song, Sohl-Dickstein, Kingma, Kumar, Ermon, and Poole]{song2021sde}
Yang Song, Jascha Sohl-Dickstein, Diederik~P. Kingma, Abhishek Kumar, Stefano Ermon, and Ben Poole.
\newblock Score-based generative modeling through stochastic differential equations, 2021.
\newblock URL \url{https://arxiv.org/abs/2011.13456}.

\bibitem[{Stability AI}(2024)]{stability2024introducing35}
{Stability AI}.
\newblock Introducing stable diffusion 3.5.
\newblock \url{https://stability.ai/news/introducing-stable-diffusion-3-5}, Oct 2024.

\bibitem[Su et~al.(2024)Su, Ahmed, Lu, Pan, Bo, and Liu]{su2024roformer}
Jianlin Su, Murtadha Ahmed, Yu~Lu, Shengfeng Pan, Wen Bo, and Yunfeng Liu.
\newblock Roformer: Enhanced transformer with rotary position embedding.
\newblock \emph{Neurocomputing}, 2024.

\bibitem[Team et~al.(2025)Team, Ma, Tan, Huang, Wu, He, Gao, Xiao, Wei, Ma, et~al.]{team2025longcat}
Meituan~LongCat Team, Hanghang Ma, Haoxian Tan, Jiale Huang, Junqiang Wu, Jun-Yan He, Lishuai Gao, Songlin Xiao, Xiaoming Wei, Xiaoqi Ma, et~al.
\newblock Longcat-image technical report.
\newblock \emph{arXiv preprint arXiv:2512.07584}, 2025.

\bibitem[Team et~al.(2026)Team, Qiao, Hui, Li, Wang, Song, Zhang, Li, Xiang, Wang, et~al.]{team2026firered}
Super~Intelligence Team, Changhao Qiao, Chao Hui, Chen Li, Cunzheng Wang, Dejia Song, Jiale Zhang, Jing Li, Qiang Xiang, Runqi Wang, et~al.
\newblock Firered-image-edit-1.0 technical report.
\newblock \emph{arXiv preprint arXiv:2602.13344}, 2026.

\bibitem[Udandarao et~al.(2024)Udandarao, Prabhu, Ghosh, Sharma, Torr, Bibi, Albanie, and Bethge]{udandarao2024pretraining}
Vishaal Udandarao, Ameya Prabhu, Adhiraj Ghosh, Yash Sharma, Philip H.~S. Torr, Adel Bibi, Samuel Albanie, and Matthias Bethge.
\newblock No "zero-shot" without exponential data: Pretraining concept frequency determines multimodal model performance, 2024.
\newblock URL \url{https://arxiv.org/abs/2404.04125}.

\bibitem[Vaswani et~al.(2017)Vaswani, Shazeer, Parmar, Uszkoreit, Jones, Gomez, Kaiser, and Polosukhin]{vaswani2017attention}
Ashish Vaswani, Noam Shazeer, Niki Parmar, Jakob Uszkoreit, Llion Jones, Aidan~N Gomez, \L~ukasz Kaiser, and Illia Polosukhin.
\newblock Attention is all you need.
\newblock In I.~Guyon, U.~Von Luxburg, S.~Bengio, H.~Wallach, R.~Fergus, S.~Vishwanathan, and R.~Garnett (eds.), \emph{Advances in Neural Information Processing Systems}, volume~30. Curran Associates, Inc., 2017.
\newblock URL \url{https://proceedings.neurips.cc/paper_files/paper/2017/file/3f5ee243547dee91fbd053c1c4a845aa-Paper.pdf}.

\bibitem[Wang et~al.(2025{\natexlab{a}})Wang, Li, Luo, Ao, Zhu, Li, and Wang]{wang2025jcomvton}
Aowen Wang, Wei Li, Hao Luo, Mengxing Ao, Chenyu Zhu, Xinyang Li, and Fan Wang.
\newblock Jco-mvton: Jointly controllable multi-modal diffusion transformer for mask-free virtual try-on.
\newblock \emph{arXiv preprint arXiv:2508.17614}, 2025{\natexlab{a}}.

\bibitem[Wang et~al.(2025{\natexlab{b}})Wang, Zhao, Zhang, Cao, Zhan, Duan, Lu, Fu, Chen, Zhao, Li, and Chen]{wang2025ovisu1technicalreport}
Guo-Hua Wang, Shanshan Zhao, Xinjie Zhang, Liangfu Cao, Pengxin Zhan, Lunhao Duan, Shiyin Lu, Minghao Fu, Xiaohao Chen, Jianshan Zhao, Yang Li, and Qing-Guo Chen.
\newblock Ovis-u1 technical report, 2025{\natexlab{b}}.
\newblock URL \url{https://arxiv.org/abs/2506.23044}.

\bibitem[Wightman(2019)]{rw2019timmwarmup}
Ross Wightman.
\newblock Pytorch image models.
\newblock \url{https://github.com/rwightman/pytorch-image-models}, 2019.

\bibitem[Wightman et~al.(2021)Wightman, Touvron, and Jégou]{wightman2021warmup2}
Ross Wightman, Hugo Touvron, and Hervé Jégou.
\newblock Resnet strikes back: An improved training procedure in timm, 2021.
\newblock URL \url{https://arxiv.org/abs/2110.00476}.

\bibitem[Wu et~al.(2025{\natexlab{a}})Wu, Li, Zhou, Lin, Gao, Yan, ming Yin, Bai, Xu, Chen, Chen, Tang, Zhang, Wang, Yang, Yu, Cheng, Liu, Li, Zhang, Meng, Wei, Ni, Chen, Cao, Peng, Qu, Wu, Wang, Yu, Wen, Feng, Xu, Wang, Zhang, Zhu, Wu, Cai, and Liu]{qwen2025qwenimage}
Chenfei Wu, Jiahao Li, Jingren Zhou, Junyang Lin, Kaiyuan Gao, Kun Yan, Sheng ming Yin, Shuai Bai, Xiao Xu, Yilei Chen, Yuxiang Chen, Zecheng Tang, Zekai Zhang, Zhengyi Wang, An~Yang, Bowen Yu, Chen Cheng, Dayiheng Liu, Deqing Li, Hang Zhang, Hao Meng, Hu~Wei, Jingyuan Ni, Kai Chen, Kuan Cao, Liang Peng, Lin Qu, Minggang Wu, Peng Wang, Shuting Yu, Tingkun Wen, Wensen Feng, Xiaoxiao Xu, Yi~Wang, Yichang Zhang, Yongqiang Zhu, Yujia Wu, Yuxuan Cai, and Zenan Liu.
\newblock Qwen-image technical report, 2025{\natexlab{a}}.
\newblock URL \url{https://arxiv.org/abs/2508.02324}.

\bibitem[Wu et~al.(2025{\natexlab{b}})Wu, Zheng, Yan, Xiao, Luo, Wang, Li, Jiang, Liu, Zhou, Liu, Xia, Li, Deng, Wang, Luo, Zhang, Lian, Wang, Wang, Huang, and Liu]{wu2025omnigen2}
Chenyuan Wu, Pengfei Zheng, Ruiran Yan, Shitao Xiao, Xin Luo, Yueze Wang, Wanli Li, Xiyan Jiang, Yexin Liu, Junjie Zhou, Ze~Liu, Ziyi Xia, Chaofan Li, Haoge Deng, Jiahao Wang, Kun Luo, Bo~Zhang, Defu Lian, Xinlong Wang, Zhongyuan Wang, Tiejun Huang, and Zheng Liu.
\newblock Omnigen2: Exploration to advanced multimodal generation, 2025{\natexlab{b}}.
\newblock URL \url{https://arxiv.org/abs/2506.18871}.

\bibitem[Xiao et~al.(2025)Xiao, Wang, Zhou, Yuan, Xing, Yan, Li, Wang, Huang, and Liu]{xiao2024omnigen}
Shitao Xiao, Yueze Wang, Junjie Zhou, Huaying Yuan, Xingrun Xing, Ruiran Yan, Chaofan Li, Shuting Wang, Tiejun Huang, and Zheng Liu.
\newblock Omnigen: Unified image generation.
\newblock In \emph{Proceedings of the IEEE/CVF Conference on Computer Vision and Pattern Recognition}, pp.\  13294--13304, 2025.

\bibitem[Xie et~al.(2024)Xie, Chen, Chen, Cai, Tang, Lin, Zhang, Li, Zhu, Lu, and Han]{xie2024sana}
Enze Xie, Junsong Chen, Junyu Chen, Han Cai, Haotian Tang, Yujun Lin, Zhekai Zhang, Muyang Li, Ligeng Zhu, Yao Lu, and Song Han.
\newblock Sana: Efficient high-resolution image synthesis with linear diffusion transformer, 2024.
\newblock URL \url{https://arxiv.org/abs/2410.10629}.

\bibitem[Xu et~al.(2023)Xu, Liu, Wu, Tong, Li, Ding, Tang, and Dong]{xu2023imagereward}
Jiazheng Xu, Xiao Liu, Yuchen Wu, Yuxuan Tong, Qinkai Li, Ming Ding, Jie Tang, and Yuxiao Dong.
\newblock Imagereward: Learning and evaluating human preferences for text-to-image generation.
\newblock \emph{Advances in Neural Information Processing Systems}, 36:\penalty0 15903--15935, 2023.

\bibitem[Yang et~al.(2025)Yang, Li, Yang, Zhang, Hui, Zheng, Yu, Gao, Huang, Lv, Zheng, Liu, Zhou, Huang, Hu, Ge, Wei, Lin, Tang, Yang, Tu, Zhang, Yang, Yang, Zhou, Zhou, Lin, Dang, Bao, Yang, Yu, Deng, Li, Xue, Li, Zhang, Wang, Zhu, Men, Gao, Liu, Luo, Li, Tang, Yin, Ren, Wang, Zhang, Ren, Fan, Su, Zhang, Zhang, Wan, Liu, Wang, Cui, Zhang, Zhou, and Qiu]{yang2025qwen3}
An~Yang, Anfeng Li, Baosong Yang, Beichen Zhang, Binyuan Hui, Bo~Zheng, Bowen Yu, Chang Gao, Chengen Huang, Chenxu Lv, Chujie Zheng, Dayiheng Liu, Fan Zhou, Fei Huang, Feng Hu, Hao Ge, Haoran Wei, Huan Lin, Jialong Tang, Jian Yang, Jianhong Tu, Jianwei Zhang, Jianxin Yang, Jiaxi Yang, Jing Zhou, Jingren Zhou, Junyang Lin, Kai Dang, Keqin Bao, Kexin Yang, Le~Yu, Lianghao Deng, Mei Li, Mingfeng Xue, Mingze Li, Pei Zhang, Peng Wang, Qin Zhu, Rui Men, Ruize Gao, Shixuan Liu, Shuang Luo, Tianhao Li, Tianyi Tang, Wenbiao Yin, Xingzhang Ren, Xinyu Wang, Xinyu Zhang, Xuancheng Ren, Yang Fan, Yang Su, Yichang Zhang, Yinger Zhang, Yu~Wan, Yuqiong Liu, Zekun Wang, Zeyu Cui, Zhenru Zhang, Zhipeng Zhou, and Zihan Qiu.
\newblock Qwen3 technical report, 2025.
\newblock URL \url{https://arxiv.org/abs/2505.09388}.

\bibitem[Ye et~al.(2026)Ye, He, Li, Yuan, Yan, Hou, Yuan, et~al.]{ye2026imgedit}
Yang Ye, Xianyi He, Zongjian Li, Shenghai Yuan, Zhiyuan Yan, Bohan Hou, Li~Yuan, et~al.
\newblock Imgedit: A unified image editing dataset and benchmark.
\newblock \emph{Advances in Neural Information Processing Systems}, 38, 2026.

\bibitem[Yu et~al.(2022)Yu, Xu, Koh, Luong, Baid, Wang, Vasudevan, Ku, Yang, Ayan, et~al.]{yu2022scaling}
Jiahui Yu, Yuanzhong Xu, Jing~Yu Koh, Thang Luong, Gunjan Baid, Zirui Wang, Vijay Vasudevan, Alexander Ku, Yinfei Yang, Burcu~Karagol Ayan, et~al.
\newblock Scaling autoregressive models for content-rich text-to-image generation.
\newblock \emph{arXiv preprint arXiv:2206.10789}, 2\penalty0 (3):\penalty0 5, 2022.

\bibitem[{Z-Image Team} et~al.(2025){Z-Image Team}, Cai, Cao, Du, Gao, Hoi, Hou, Huang, Jiang, Jin, Li, Li, Li, Liu, Liu, Shi, Wu, Yu, Zhang, Zhang, and Zhou]{zimage2025}
{Z-Image Team}, Huanqia Cai, Sihan Cao, Ruoyi Du, Peng Gao, Steven Hoi, Zhaohui Hou, Shijie Huang, Dengyang Jiang, Xin Jin, Liangchen Li, Zhen Li, Zhong-Yu Li, David Liu, Dongyang Liu, Junhan Shi, Qilong Wu, Feng Yu, Chi Zhang, Shifeng Zhang, and Shilin Zhou.
\newblock Z-image: An efficient image generation foundation model with single-stream diffusion transformer, 2025.
\newblock URL \url{https://arxiv.org/abs/2511.22699}.

\bibitem[{Z.AI}(2026)]{glmimage2026}
{Z.AI}.
\newblock Glm-image: Auto-regressive for dense-knowledge and high-fidelity image generation.
\newblock \url{https://z.ai/blog/glm-image}, 2026.
\newblock Apache-2.0 License.

\bibitem[Zhang et~al.(2025)Zhang, Jiang, Xu, Chen, Jin, Lu, Zhang, Yong, Luo, and Luo]{zhang2025worldgenbench}
Daoan Zhang, Che Jiang, Ruoshi Xu, Biaoxiang Chen, Zijian Jin, Yutian Lu, Jianguo Zhang, Liang Yong, Jiebo Luo, and Shengda Luo.
\newblock Worldgenbench: A world-knowledge-integrated benchmark for reasoning-driven text-to-image generation.
\newblock \emph{arXiv preprint arXiv:2505.01490}, 2025.

\bibitem[Zhang et~al.(2023)Zhang, Rao, and Agrawala]{zhang2023adding}
Lvmin Zhang, Anyi Rao, and Maneesh Agrawala.
\newblock Adding conditional control to text-to-image diffusion models.
\newblock In \emph{Proceedings of the IEEE/CVF international conference on computer vision}, pp.\  3836--3847, 2023.

\bibitem[Zhang et~al.(2026)Zhang, Li, Zhang, Gao, Yan, Jiang, Tang, Yin, Wu, Chen, Xu, Shu, Zhang, Xu, Chen, Wang, Liu, Zhou, Zhang, Zhao, and Wu]{zhang2026qwenimageagentbridgingcontextgap}
Zekai Zhang, Jiahao Li, Jie Zhang, Kaiyuan Gao, Kun Yan, Lihan Jiang, Ningyuan Tang, Shengming Yin, Tianhe Wu, Xiaoyue Chen, Xiao Xu, Yan Shu, Yanran Zhang, Yixian Xu, Yuxiang Chen, Zhendong Wang, Zihao Liu, Zikai Zhou, Huishuai Zhang, Dongyan Zhao, and Chenfei Wu.
\newblock Qwen-image-agent: Bridging the context gap in real-world image generation, 2026.
\newblock URL \url{https://arxiv.org/abs/2606.26907}.

\bibitem[Zhao et~al.(2026)Zhao, Wu, Li, Meng, Li, Zhang, Zhou, Lin, Gao, Cao, et~al.]{zhao2026qwen2}
Bing Zhao, Chenfei Wu, Deqing Li, Hao Meng, Jiahao Li, Jie Zhang, Jingren Zhou, Junyang Lin, Kaiyuan Gao, Kuan Cao, et~al.
\newblock Qwen-image-2.0 technical report.
\newblock \emph{arXiv preprint arXiv:2605.10730}, 2026.

\bibitem[Zhao et~al.(2023)Zhao, Gu, Varma, Luo, Huang, Xu, Wright, Shojanazeri, Ott, Shleifer, Desmaison, Balioglu, Damania, Nguyen, Chauhan, Hao, Mathews, and Li]{zhao2023pytorchfsdp}
Yanli Zhao, Andrew Gu, Rohan Varma, Liang Luo, Chien-Chin Huang, Min Xu, Less Wright, Hamid Shojanazeri, Myle Ott, Sam Shleifer, Alban Desmaison, Can Balioglu, Pritam Damania, Bernard Nguyen, Geeta Chauhan, Yuchen Hao, Ajit Mathews, and Shen Li.
\newblock Pytorch fsdp: Experiences on scaling fully sharded data parallel, 2023.
\newblock URL \url{https://arxiv.org/abs/2304.11277}.

\bibitem[Zheng et~al.(2024)Zheng, Chiang, Sheng, Zhuang, Wu, Zhuang, Lin, Li, Li, Xing, et~al.]{zheng2024chatbot}
Lianmin Zheng, Wei-Lin Chiang, Ying Sheng, Siyuan Zhuang, Zhanghao Wu, Yonghao Zhuang, Zhi~Zheng Lin, Zi~Ning Li, Dacheng Li, Eric Xing, et~al.
\newblock Chatbot arena: An open platform for evaluating llms based on human preference.
\newblock In \emph{arXiv preprint arXiv:2403.04132}, 2024.

\bibitem[Zhu et~al.(2025)Zhu, Wang, Chen, Liu, Ye, Gu, Tian, Duan, Su, Shao, Gao, Cui, Wang, Cao, Liu, Wei, Zhang, Wang, Xu, Li, Wang, Deng, Li, He, Jiang, Luo, Wang, He, Shi, Zhang, Shao, He, Xiong, Qu, Sun, Jiao, Lv, Wu, Zhang, Deng, Ge, Chen, Wang, Dou, Lu, Zhu, Lu, Lin, Qiao, Dai, and Wang]{zhu2025internvl3exploringadvancedtraining}
Jinguo Zhu, Weiyun Wang, Zhe Chen, Zhaoyang Liu, Shenglong Ye, Lixin Gu, Hao Tian, Yuchen Duan, Weijie Su, Jie Shao, Zhangwei Gao, Erfei Cui, Xuehui Wang, Yue Cao, Yangzhou Liu, Xingguang Wei, Hongjie Zhang, Haomin Wang, Weiye Xu, Hao Li, Jiahao Wang, Nianchen Deng, Songze Li, Yinan He, Tan Jiang, Jiapeng Luo, Yi~Wang, Conghui He, Botian Shi, Xingcheng Zhang, Wenqi Shao, Junjun He, Yingtong Xiong, Wenwen Qu, Peng Sun, Penglong Jiao, Han Lv, Lijun Wu, Kaipeng Zhang, Huipeng Deng, Jiaye Ge, Kai Chen, Limin Wang, Min Dou, Lewei Lu, Xizhou Zhu, Tong Lu, Dahua Lin, Yu~Qiao, Jifeng Dai, and Wenhai Wang.
\newblock Internvl3: Exploring advanced training and test-time recipes for open-source multimodal models, 2025.
\newblock URL \url{https://arxiv.org/abs/2504.10479}.

\bibitem[Zhuo et~al.(2024)Zhuo, Du, Xiao, Li, Liu, Huang, Liu, Zhu, Wang, Ma, Luo, Wang, Zhang, Zhao, Liu, Yue, Ouyang, Qiao, Li, and Gao]{zhuo2024LuminaNext}
Le~Zhuo, Ruoyi Du, Han Xiao, Yangguang Li, Dongyang Liu, Rongjie Huang, Wenze Liu, Xiangyang Zhu, Fu-Yun Wang, Zhanyu Ma, Xu~Luo, Zehan Wang, Kaipeng Zhang, Lirui Zhao, Si~Liu, Xiangyu Yue, Wanli Ouyang, Yu~Qiao, Hongsheng Li, and Peng Gao.
\newblock Lumina-next: making lumina-t2x stronger and faster with next-dit.
\newblock In \emph{Proceedings of the 38th International Conference on Neural Information Processing Systems}, NIPS '24, Red Hook, NY, USA, 2024. Curran Associates Inc.
\newblock ISBN 9798331314385.

\bibitem[汉字处(1988)]{frequently_used_zh}
国家语言文字工作委员会. 汉字处.
\newblock \emph{现代汉语常用字表}.
\newblock 语文出版社, 1988.
\newblock URL \url{https://books.google.com.hk/books?id=Av_tAQAACAAJ}.

\end{thebibliography}
\bibliographystyle{reference}

\clearpage
\appendix

\boogutocchapterstar{Appendix}

\section{Contributors}
\label{app:contributors}

\newcommand{\booguFont}{\sffamily}
\newcommand{\booguColor}{boogured} %

\newcommand{\booguTeamSize}{\fontsize{11.5pt}{14pt}\selectfont}
\newcommand{\booguLabSize}{\fontsize{9.5pt}{11.5pt}\selectfont}

\newcommand{\inlineLogo}[2]{\raisebox{-0.2\height}{\includegraphics[height=#1]{#2}}}

\begin{center}
    \color{\booguColor}\booguFont
    
    {\booguTeamSize Boogu Team} \\[0.2cm]
    
    {\booguLabSize Celia Large Model Application Laboratory (Leibniz)} \\[0.2cm]
    
    {\booguLabSize
        \hspace*{4.5em}%
        \makebox[0pt][r]{\inlineLogo{1.5em}{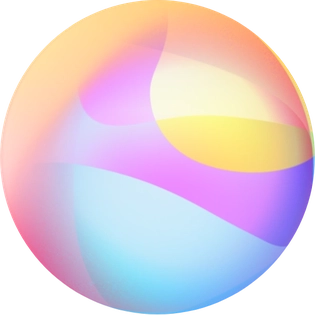}\hspace{0.35em}}%
        小艺大模型应用实验室（莱布尼茨所）}
\end{center}
\vspace{0.4cm} %

\begin{itemize}

  \item \textbf{Project Sponsors:}
        {Rui Liu}$^1$, Chao Huang$^2$, Han Shi$^1$, Haoli Bai$^1$, Xihui Liu$^2$, Hongsheng Li$^6$, Qifeng Chen$^3$

  \item \textbf{Project Leader}: Chenyang Lei$^1$

  \item \textbf{Core Contributors:} 
                {Guoxuan Chen$^{1,2,*}$, Chufeng Xiao$^1$, Haoran Yang$^1$, Siyue Xie$^1$}, Binxiao Huang$^1$, Ming Zhang$^1$, Cheuk Him Chau$^1$, Xinyu Fu$^1$, Yingzhao Lian$^1$,  Tom S.Y. Li$^1$, Jintao Lin$^{1,2,*}$, Bowen Dong$^{1,4,*}$, Zian Qian$^{1,3,*}$, Chenyang Lei$^1$
        
  \item \textbf{Contributors}: Yuhao Liu$^{1,5,*}$, Yuxuan Hu$^{1,6,*}$, Weikang Shi$^{1,6,*}$, Bin Zou$^{1,2,*}$, Bowen Zheng$^{1,7,*}$, Haoxuan Che$^{1,\dagger}$, Chang Chen$^{1,3,*}$, Yuyang He$^1$, Heyang Sun$^{1,3,*}$, Tianyu Huang$^{1,4,*}$, Chong Hou Choi$^1$, Cheng Gong$^1$

  \item \textbf{Correspondence}: Chenyang Lei$^1$, Rui Liu$^1$, Chao Huang$^2$
\end{itemize}

\textbf{Primary Affiliation}:  $^1$Artificial Intelligence Laboratory (Leibniz), Huawei Technologies, Hong Kong.~~~\inlineLogo{2.5em}{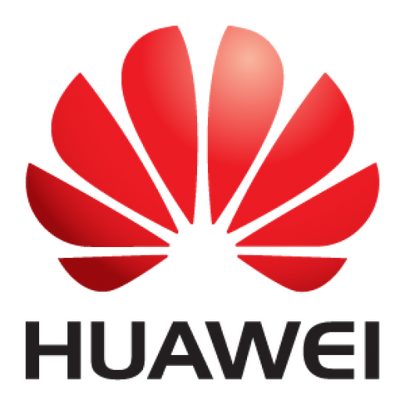}

\textbf{Collaborating Institutions}: $^2$The University of Hong Kong, $^3$The Hong Kong University of Science and Technology, $^4$The Hong Kong Polytechnic University, $^5$City University of Hong Kong, $^6$The Chinese University of Hong Kong, $^7$Nanjing University.

\vspace{0.1in}
\textit{Note}:  * \textit{indicates work completed during an internship}. \(\dagger\) \textit{indicates the author is no longer affiliated with the primary affiliation.}

\clearpage

\section{Background and Technical Choices}

\label{sec:arch_design}
\subsection{Background}

The evolution of visual synthesis has rapidly transitioned from isolated generative tasks to integrated multimodal frameworks. This section reviews the foundational trajectories contextualizing our work: text-to-image diffusion, controllable editing, vision-language models (VLMs), and the recent paradigm shift towards unified generation.

\paragraph{Text-to-Image Diffusion Models.} 
Diffusion models \citep{ho2020ddpm} have fundamentally revolutionized visual synthesis. Foundational works like LDM \citep{rombach2022high} and DALL-E 2 \citep{ramesh2022hierarchical} established the paradigm of high-fidelity text-conditioned generation. Subsequent advancements focused on text-image alignment, with DALL-E 3 \citep{betker2023improving} leveraging recaptioned datasets for superior prompt adherence. Furthermore, the architectural shift from U-Nets \citep{ronneberger2015u} to transformers \citep{vaswani2017attention}, pioneered by DiT \citep{peebles2023dit}, unlocked unprecedented scalability. This transformer-driven scalability is prominently demonstrated by the spatiotemporal video generation of Sora \citep{brooks2024video} and the rectified flow formulations of Stable Diffusion 3 \citep{esser2024scaling} for complex image synthesis. Most recently, this trajectory has culminated in the proprietary and closed-source Nano Banana series \citep{google2025nanobananapro} and ChatGPT Images series \citep{openai2026gptimage2}, which currently define the state-of-the-art in high-fidelity text-to-image generation.

\paragraph{Controllable Generation and Editing.} 
Relying solely on text prompts often falls short when precise spatial control is required. To address this, ControlNet \citep{zhang2023adding} introduced auxiliary conditioning, enabling diffusion models to respect spatial hints (e.g., depth or poses) without compromising the original generative prior. In the domain of image editing, InstructPix2Pix \citep{brooks2023instructpix2pix} pioneered instruction-driven modifications, allowing users to alter existing images via natural language instructions without relying on complex manual masks.

\paragraph{Vision-Language Models (VLMs).} 
Driven by the success of LLMs, VLMs have rapidly advanced visual comprehension. Pioneering works like LLaVA \citep{liu2024visual} aligned visual encoders—predominantly based on Vision Transformers (ViTs) \citep{dosovitskiy2020image}—with LLMs via visual instruction tuning. Recent state-of-the-art models, such as Ovis \citep{lu2024ovis} and the Qwen-VL series \citep{bai2025qwen25vl, bai2025qwen3vl}, have further refined structural alignment and visual reasoning, enabling the processing of high-resolution inputs for intricate multimodal tasks.

\paragraph{Towards Unified Multimodal Generation.} 
Building upon VLMs' reasoning and diffusion models' generative prowess, the community is shifting towards consolidated architectures. OmniGen \citep{xiao2024omnigen} and OmniGen2 \citep{wu2025omnigen2} pioneered this by natively handling diverse generation and editing tasks within a single framework. Concurrently, while proprietary models like Google's Nano Banana series \citep{google2025nanobananapro} have introduced reasoning-guided synthesis via interleaved prompts, open-weight alternatives such as the Qwen-Image series \citep{qwen2025qwenimage} and Z-Image \citep{zimage2025} integrate deep visual understanding with scalable, efficient generation, paving the way for versatile and intuitive multimodal AI systems.

\subsection{Macro-Architecture Design}

Empirical observations of state-of-the-art proprietary models, such as ChatGPT Images series \citep{openai2026gptimage2} and the Nano Banana series \citep{google2025nanobananapro}, reveal a consistent latency prior to generation. We hypothesize that, regardless of their underlying architectures, these systems employ a robust comprehension module to meticulously analyze and deconstruct user instructions, which naturally incurs a temporal overhead. Motivated by this, our preliminary architectural design utilized a massive 72B Vision-Language Model (VLM) to drive a 10B Diffusion Transformer (DiT), where the VLM was kept frozen and only the DiT parameters were updated. However, this configuration imposes prohibitive computational overheads during both training and inference, rendering it inaccessible to the broader research community. To address this, we propose a decoupled, compute-efficient paradigm: we employ a moderately large VLM (e.g., 32B) as an \textit{Instruction Reasoner} for complex logic comprehension, paired with a lightweight VLM (e.g., 8B) acting as an \textit{Instruction Encoder} to extract hidden states as conditioning signals for the DiT. This decoupled strategy significantly mitigates both training and inference costs compared to relying on a monolithic, large VLM.
\begin{figure}[h!]
  \centering
  \includegraphics[width=1\linewidth]{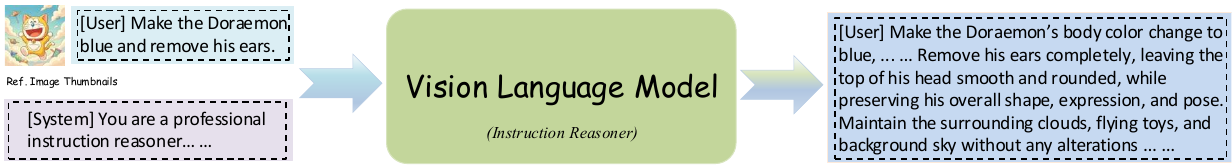}
  \caption{\textbf{Illustration of the Instruction Reasoner workflow.} The input text instruction, formatted via a predefined chat template with a system prompt, is processed by the Reasoner. In image editing scenarios, a reference image is appended to the input. The Reasoner yields a comprehensively analyzed, deconstructed, and rewritten instruction to condition the subsequent generation pipeline. }
  \label{fig:instruct_reasoner}
\end{figure}

\subsubsection{Instruction Reasoner} 
The operational pipeline of the Instruction Reasoner is depicted in Figure~\ref{fig:instruct_reasoner}. We introduce a unified architecture that seamlessly integrates both text-to-image (T2I) generation and image editing. Specifically, the system determines the task modality based on the presence of visual inputs: if the user-provided reference image is omitted, the process defaults to a T2I task and applies the corresponding T2I system prompt. Conversely, the inclusion of a reference image explicitly designates an editing operation, thereby activating an editing-specific system prompt. Ultimately, the Reasoner outputs a meticulously parsed, structurally decomposed, and refined textual prompt, which serves as the optimized conditioning signal for the subsequent generation stages.

\subsubsection{Instruction Encoder and Image Generation}

\begin{figure}[!t]
  \centering
  \includegraphics[width=1\linewidth]{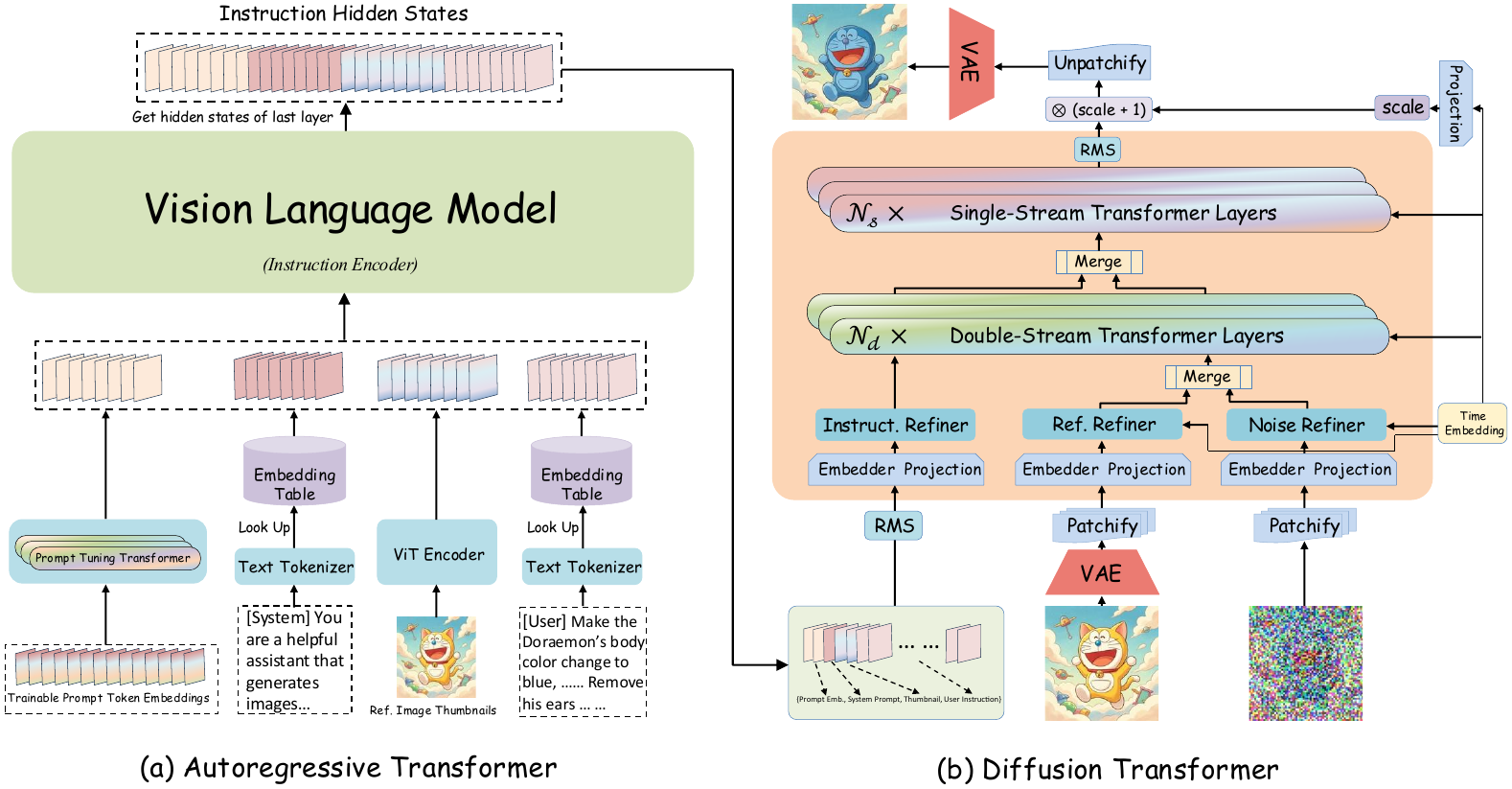}
  \caption{\textbf{Pipeline of the instruction encoder and DiT.} The framework defaults to image editing when a reference image is provided, and text-to-image generation otherwise. A thumbnail of the reference image is fed to the Instruction Encoder to aid comprehension, while the original image (scaled within the resolution limit) is processed by the DiT. The system prompt adaptively adjusts to the specific task. Furthermore, an optional trainable Prompt Tuning Transformer is introduced to enhance instruction parsing. Note that the input user instructions have been pre-analyzed and rewritten by the Instruction Reasoner.}
  \label{fig:instruct_enc_dit}
\end{figure}

\begin{figure}[ht]
  \centering
  \includegraphics[width=0.25\linewidth]{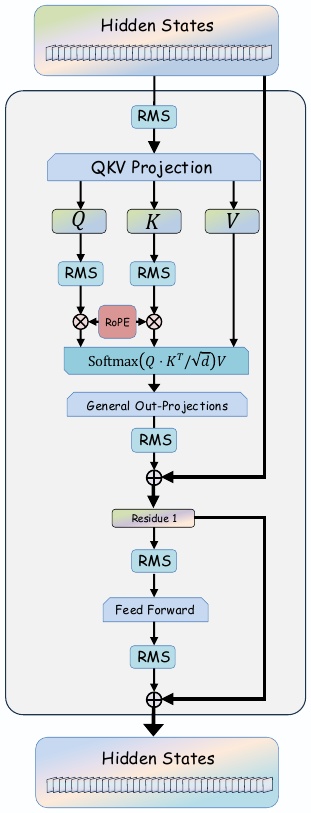}
  \caption{\textbf{Lightweight single-stream transformer layer.} This is a lightweight transformer module designed for various refiners and optional prompt tuning components. This module adopts the classic design \cite{peebles2023dit, esser2024scaling,lipman2023flowmatching, dehghani2023scalingvisiontransformers22, stability2024introducing35,chen2023pixartalpha, gao2024luminat2x, zhuo2024LuminaNext, wang2025ovisu1technicalreport, xiao2024omnigen, wu2025omnigen2, qwen2025qwenimage} with minimal modifications.}
  \label{fig:simple_trans_layer}
\end{figure}

Upon acquiring the text instructions that have been analyzed, deconstructed, and rewritten by the Instruction Reasoner, we employ a subsequent pipeline to generate images. This pipeline primarily consists of two core modules: an Autoregressive Transformer \citep{vaswani2017attention} and a Diffusion Transformer (DiT) \citep{peebles2023dit}. The Autoregressive Transformer essentially functions as a Vision-Language Model (VLM), serving as our Instruction Encoder. The sub-modules comprising the pipeline of the Instruction Encoder are briefly introduced as follows (see Figure.\ref{fig:instruct_enc_dit}):
\paragraph{Text Tokenizer and Embedding Table} This module is responsible for tokenizing the rewritten text instructions and projecting them into a sequence of embedding vectors. This sequence serves as a partial input to the Autoregressive Transformer.
\paragraph{Vision Transformer (ViT)} The number of user-provided reference images can range from zero to multiple. When present, the Vision Transformer (ViT) \citep{dosovitskiy2020image} encodes them into the same latent space as the text prompt embeddings. These visual and textual embeddings are concatenated into a unified sequence before being fed into the Autoregressive Transformer. As a standard component of most VLMs, the ViT processes these reference images, which are here downsampled to a $384 \times 384$ resolution to prevent excessively long vector sequences. Notably, if no reference images are provided, the task is defined as pure text-to-image generation, and the ViT is consequently bypassed.
\paragraph{System Prompt} The task paradigm is determined based on the presence of reference images. Specifically, the absence of any reference images defines a text-to-image generation task, whereas the presence of one or more reference images indicates an image editing task. The system prompt is adaptively formulated corresponding to the specific task type.
\paragraph{Prompt Tuning Transformer (Optional)} This is a highly lightweight, trainable Transformer module. In our implementation, it is fixed at $3$ layers, with the architecture of each layer illustrated in Figure~\ref{fig:simple_trans_layer}. It contains only $32$ trainable prompt token embeddings and employs causal masking. The output from its final layer is prepended to the input sequence of the Instruction Encoder. The Prompt Tuning Transformer serves as an optional trainable module designed to enhance the instruction parsing and comprehension capabilities of the overall text-to-image model. Since the Instruction Encoder remains frozen during training, mathematically, the introduction of this module globally modulates the embedding distribution of the hidden state sequence output by the encoder, thereby facilitating a better alignment with the semantic space required by the DiT.

It is worth noting that the Prompt Tuning Transformer is an optional module, which will be discussed further in the subsequent experimental section. In addition, Figure~\ref{fig:instruct_enc_dit}(b) illustrates the Diffusion Transformer (DiT) component. The sub-modules involved in this component are briefly introduced below:

    \paragraph{Variational Autoencoder (VAE)} When user-provided reference images are present, to balance performance and computational efficiency, we first employ a Variational Autoencoder (VAE) \citep{kingma2013auto} encoder to map them into a latent space before feeding them into the DiT. Consequently, our diffusion training is conducted entirely within this latent space \citep{rombach2022high}. Once mapped, the reference images undergo patchification, followed by processing through an Embedder and a Refiner. The resulting hidden states, along with the hidden states corresponding to the noise, are fed into the DiT layers for sequential processing. The output of the final layer must be unpatchified and passed through the VAE decoder to reconstruct the output image. When no reference images are provided, the VAE encoder is bypassed; however, the VAE decoder remains essential for decoding the final output image, as the noise depicted in the bottom right of Figure~\ref{fig:instruct_enc_dit}(b) essentially represents the noise state within the VAE-encoded latent space.

    \paragraph{Embedder and Refiner} The Embedder is a standard linear layer designed to project hidden states from one latent space to another. The Refiner is a simple, lightweight Transformer \citep{vaswani2017attention}. By default, it consists of only $2$ layers, with each layer's architecture depicted in Figure~\ref{fig:simple_trans_layer} (identical to the structure of the Prompt Tuning Transformer). Compared to the Embedder, the Refiner also accounts for the inter-relationships among elements in the newly patchified sequence. Notably, the user instructions, reference images, and noise each possess their own independent Embedders and Refiners.

    \paragraph{Single-Stream and Dual-Stream Layers} Our DiT employs a stacked architecture comprising \(\mathcal{N}_d\) dual-stream layers followed by \(\mathcal{N}_s\) single-stream layers. After being projected into a unified latent space by their respective Refiners and associated modules, the reference images and noise are concatenated and fed into the image stream of the dual-stream layers. Simultaneously, the hidden states of the user instructions from the final layer of the Instruction Encoder are projected into the instruction latent space by their corresponding Refiner, \etc, and then input into the instruction stream of the dual-stream layers. The outputs from the two streams of the final dual-stream layer are concatenated and fed into the unified stream of the single-stream layers. Following a series of operations, the output from the single-stream layers is ultimately decoded by the VAE decoder to produce the final image.

Figure~\ref{fig:instruct_enc_dit} illustrates the macro-architecture design and overall pipeline for text-to-image generation and image editing. During the actual inference phase, the DiT component typically requires multiple iterative steps to synthesize high-fidelity images \citep{ho2020ddpm, peebles2023dit}. This iterative process is accompanied by the transition of the time step \(t\) from 0 to 1 (or from 1 to 0, depending on the specific formulation), along with corresponding updates to the Time Embedding. For the specific iterative sampling algorithm during generation, one can employ the standard first-order Euler method commonly utilized in Flow Matching \citep{lipman2023flowmatching}, or fast dedicated ODE solvers such as DPM-Solver++ \citep{lu2025dpmsolverpp}.

\subsection{Micro-Architecture Details}

Building upon the macro-level overview presented in the previous section, we now delve into the micro-architectural details of our model. Our specific design draws inspiration from recent foundational works \cite{peebles2023dit, esser2024scaling,lipman2023flowmatching, dehghani2023scalingvisiontransformers22, stability2024introducing35,chen2023pixartalpha, gao2024luminat2x, zhuo2024LuminaNext, wang2025ovisu1technicalreport, xiao2024omnigen, wu2025omnigen2, qwen2025qwenimage}, which we have appropriately modified and extended to suit our objectives. Overall, we propose a unified framework that seamlessly supports both text-to-image generation and image editing. By integrating and adapting the aforementioned designs, we incorporate specialized modules to enhance fine-grained image comprehension and comprehensive multimodal alignment. The detailed components are elaborated below.

\subsubsection{Dual-Stream Layer}

\begin{figure}[!t]
  \centering
  \includegraphics[width=0.9\linewidth]{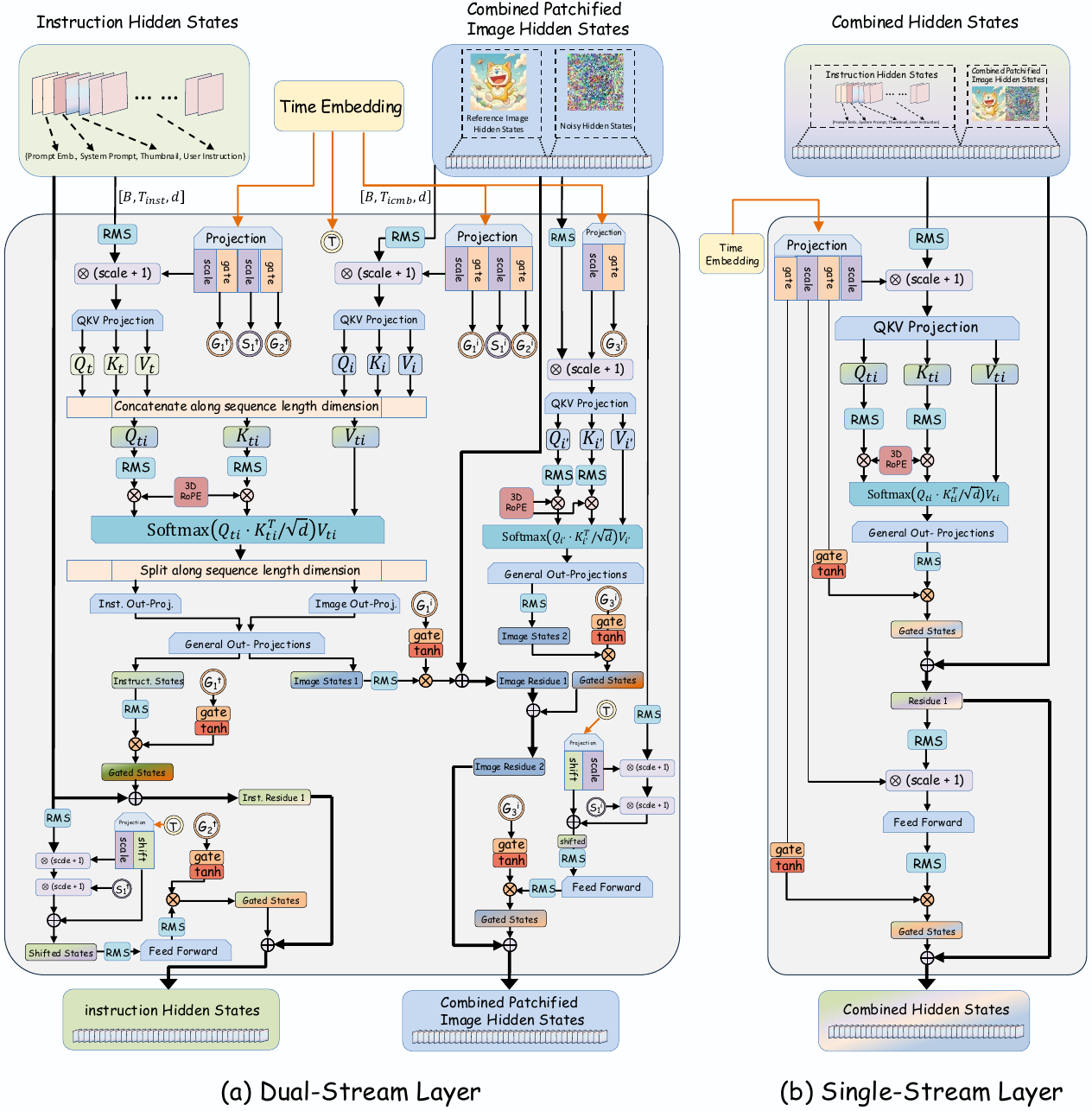}
  \caption{\textbf{Detailed architecture of the dual-stream and single-stream layers}. Bold lines indicate the Main Path constructed via residual connections, while thin lines represent Branch Paths involving complex computations. Retaining only the bold lines reduces both layers to identity mappings. The branch paths merge into the main path at appropriate junctures to influence the parameter distribution of a layer's output.}
  \label{fig:sgl_doub_str_layers}
\end{figure}

The architecture of the dual-stream layer is illustrated in Figure~\ref{fig:sgl_doub_str_layers}(a). In this design, the user instruction stream and the image stream (comprising the noisy latent and the optional reference image) are processed separately within the layer. Unlike conventional dual-stream architectures, which typically either maintain independent linear projections solely for the Query, Key, and Value tensors of the two streams \citep{wang2025jcomvton}, or decouple only a small subset of additional computational modules \citep{esser2024scaling}, our approach draws inspiration from recent advancements \citep{wang2025ovisu1technicalreport, gao2024luminat2x, zhuo2024LuminaNext}. Specifically, it incorporates a more comprehensive set of independent intra-stream and inter-stream modules to facilitate deeper computational processing and feature interaction.

\paragraph{Boosting Multimodal Understanding} 
Profound comprehension of user instructions is crucial for image generation and editing. Our VLM-based Instruction Reasoner deconstructs user prompts, feeding rewritten text instructions to the Instruction Encoder to extract generation conditions. During this process, although this encoder utilizes a ViT to process optional reference images, its primary focus remains text instruction modeling rather than comprehensive multimodal understanding.

To explicitly enhance multimodal understanding, following recent architectures \citep{peebles2023dit,wang2025ovisu1technicalreport, qwen2025qwenimage, gao2024luminat2x, zhuo2024LuminaNext}, we adopt an independent attention module for the image modality (comprising the noisy latent and optional reference image), as shown in Figure~\ref{fig:sgl_doub_str_layers}(a, right). Within the dual-stream layer, the right stream processes image token interactions to deepen visual comprehension. Conversely, the left stream fuses the instruction and image modalities (Figure~\ref{fig:sgl_doub_str_layers}(a, left)). By incorporating instruction residual connections, it inherently emphasizes instruction modeling from a holistic multimodal perspective, despite performing full-modality attention. Subsequently, the image information derived from the left stream's full-modality analysis is fused with the residual and internal self-attention outputs of the right stream. This integration yields a refined representation grounded in holistic comprehension while prioritizing the image modality. Furthermore, the dual-stream layer features a combination of shared and independent computational modules, including three independent Query, Key, and Value (QKV) projections. These QKV sets are allocated to: the text modality, the image modality used for holistic enhancement, and the image modality's internal self-attention, respectively.

Collectively, these two streams—rooted in holistic understanding but respectively emphasizing the instruction and image modalities—enhance the model's multimodal comprehension at a fine-grained level. This synergy paves the way for more macroscopic and comprehensive modeling in the subsequent single-stream layers after the streams converge.

\paragraph{Main Path, Residue, and Modulation}
Our dual-stream layer adopts the classic sandwich norm \citep{ming2021cogview,zhuo2024LuminaNext} and the modulation mechanism \citep{peebles2023dit} to facilitate information fusion and signal control. However, our specific implementation introduces slight modifications (see Figure~\ref{fig:sgl_doub_str_layers}(a)). The multi-norm design significantly enhances model stability during pre-training with large learning rates, thereby accelerating convergence and improving overall robustness. In Figure~\ref{fig:sgl_doub_str_layers}(a), the bold lines denote the \textit{Main Path}, which is fundamentally constructed with residual connections to stabilize the parameter distribution across layers. If only this main path existed, a layer reduces to an identity mapping. Conversely, the thin lines represent \textit{Branch Paths} that incorporate various operations, including attention and modulation. These branches are responsible for analyzing and processing the input representations. The computational results from these branch paths are subsequently merged into the main path at appropriate junctures, ultimately shaping the final output distribution of the layer.

\subsubsection{Single-Stream Layer}

The architecture of our single-stream layer follows numerous classic designs \citep{wang2025ovisu1technicalreport, wu2025omnigen2, gao2024luminat2x, zhuo2024LuminaNext, qwen2025qwenimage}. Specifically, we adopt the classical modulation mechanism \citep{peebles2023dit} and sandwich norm \citep{ming2021cogview, zhuo2024LuminaNext} (see Figure~\ref{fig:sgl_doub_str_layers}(a)). This layer merges the hidden states of both the instruction and image streams from the final dual-stream layer—which have already been deeply comprehended and modeled—into a unified representation, and then it performs comprehensive multi-modal fusion and analysis, processing the sequence layer by layer to yield the final output. The computational logic within this layer is unified and modality-agnostic; both instruction and image inputs share the same parameters, including the QKV projections. As in the dual-stream layer, the bold lines denote the main path, while the thin lines represent the branch paths. Retaining only the main path reduces the layer to an identity mapping. The branch paths are responsible for complex computations and merge into the main path at appropriate junctures.

\subsection{Summary of Architectural Design}

In this section, we summarize our architectural design and detail the specific hyperparameter choices. Specifically, our DiT architecture comprises \(\mathcal{N}_s=32\) single-stream layers and \(\mathcal{N}_d=8\) dual-stream layers. We incorporate 3D Rotary Positional Embeddings (RoPE) (following ~\cite{wu2025omnigen2,su2024roformer}) and, by default, utilize the hidden states from the final layer of the Instruction Encoder as the condition for the DiT. The base DiT model contains 10,292.56M ($\sim$\textbf{10B}) trainable parameters. When the optional Prompt Tuning Transformer is included, the total trainable parameter count slightly increases to 11,022.55M. Further architectural details and hyperparameter settings are comprehensively summarized in Table~\ref{tab:arch_details}.

\begin{table}[htbp]
\centering
\renewcommand{\arraystretch}{1.15}
\begin{tabular}{@{} l c @{\hspace{3em}} l c @{}}
\toprule
\textbf{Parameter} & \textbf{Value} & \textbf{Parameter} & \textbf{Value} \\
\midrule
\multicolumn{4}{c}{\textit{Base Architecture}} \\
\midrule
Patch Size & 2 & Single-Stream Layers & 32 \\
Input Channels & 16 & Dual-Stream Layers & 8 \\
Hidden Size & 3360 & Refiner Layers & 2 \\
Attention Heads & 28 & KV Heads & 7 \\
FFN Multiple Of & 256 & RoPE Axes Dims & $[40, 40, 40]$ \\
\midrule
\multicolumn{4}{c}{\textit{Instruction Feature Configuration}} \\
\midrule
Feature Layers & 1 & Feature Dimension & 4096 \\
Reduction Type & Mean & & \\
\midrule
\multicolumn{4}{c}{\textit{Pretrained Modules}} \\
\midrule
VAE Model & \begin{tabular}[c]{@{}c@{}}FLUX.1-dev \\ \citep{flux2024blackforest}\end{tabular} & Instruction Encoder & \begin{tabular}[c]{@{}c@{}}Qwen3-VL-8B-Instruct \\ \citep{bai2025qwen3vl}\end{tabular} \\
\midrule
\multicolumn{4}{c}{\textit{Prompt Tuning (Optional)}} \\
\midrule
Trainable Tokens & 32 & Causal Mask & True \\
Hidden Size & 4096 & Layers & 3 \\
Attention Heads & 32 & KV Heads & 8 \\
\midrule
\multicolumn{4}{c}{\textit{Trainable Parameters}} \\
\midrule
DiT Model & 10,292.56M & Total (w/ Prompt Tuning) & 11,022.55M \\
\bottomrule
\end{tabular}
\caption{Summary of architectural design, specific hyperparameter configurations, and parameter counts.}
\label{tab:arch_details}
\end{table}
\clearpage

\subsection{Training Process}

Our overall training pipeline follows a progressive curriculum: we first train on low-resolution images before scaling to higher resolutions, and initially train the text-to-image (T2I) model before fine-tuning it into the text-and-image-to-image (TI2I) editing model. The complete training process is summarized in Table~\ref{tab:training_hyperparams}. 

The parameters \textit{Max Output Pixels}, \textit{Max Input Pixels}, and \textit{Max Side Length} define the spatial constraints during training. During data preprocessing, images exceeding these thresholds are dynamically downscaled to fit within the limits, while smaller or equal-sized images remain unchanged, inherently enabling mixed-resolution training. 

For the TI2I training phase, we construct a mixed dataset of approximately 22M samples. This comprises 11.29M pure T2I samples sampled from the 47.19M \model~Syllabus dataset (see \ref{sec:boogu_syllabus}), and 10.70M editing samples augmented from our initially constructed 2.50M TI2I dataset. Consequently, our image editing model is optimized through a combination of mixed-resolution and mixed-task training.

\begin{table}[htbp]
\centering
\resizebox{\linewidth}{!}{
\begin{tabular}{l | c c c | c}
\toprule
\multirow{2}{*}{\textbf{Parameter}} & \multicolumn{3}{c|}{\textbf{T2I}} & \textbf{TI2I} \\
\cmidrule(lr){2-4} \cmidrule(lr){5-5}
 & \textbf{512$\times$512} & \textbf{1024$\times$1024} & \textbf{2048$\times$2048} & \textbf{2048$\times$2048} \\
\midrule
\multicolumn{5}{c}{\textit{Resolution \& Data Constraints}} \\
\midrule
Training Data & \begin{tabular}{@{}c@{}}Open-Source \\ 187M\end{tabular} & \begin{tabular}{@{}c@{}}Open-Source 60M + \\ \model~Syllabus 47M\end{tabular} & \begin{tabular}{@{}c@{}}\model~Syllabus \\ 47M\end{tabular} & \begin{tabular}{@{}c@{}}11M T2I + \\ 11M TI2I\end{tabular} \\
Max Output Pixels & 262,144 & 1,048,576 & \multicolumn{2}{|c}{4,194,304} \\
Max Input Pixels & 262,144 & 1,048,576 & \multicolumn{2}{|c}{4,194,304} \\
Max Side Length & 800 & 2048 & \multicolumn{2}{|c}{4096} \\
Max Instruction Tokens & 256 & 512 & \multicolumn{2}{|c}{944} \\
Instruct. Dropout Prob. & \multicolumn{4}{c}{0.1} \\
Ref. Img. Dropout Prob. & \multicolumn{3}{c|}{N/A} & 0.01 \\
\midrule
\multicolumn{5}{c}{\textit{Batch Size \& Steps}} \\
\midrule
Global Batch Size & 1280 & 2048 & 1024 & 448 \\
Micro Batch Size & 10 & 8 & 2 & 1 \\
Num Train Epochs & 1 & \multicolumn{3}{|c}{2} \\
Gradient Accumulation & \multicolumn{4}{c}{1} \\
\midrule
\multicolumn{5}{c}{\textit{Optimizer \& Learning Rate}} \\
\midrule
Learning Rate & \multicolumn{2}{c|}{1e-4} & \multicolumn{2}{c}{3e-5} \\
Warmup Steps & 1200 & \multicolumn{3}{|c}{800} \\
LR Scheduler & \multicolumn{4}{c}{Constant with Warmup \cite{goyal2018wwarmup1,rw2019timmwarmup,wightman2021warmup2}} \\
Warmup Init LR & \multicolumn{4}{c}{1e-6} \\
Adam $\beta_1, \beta_2$ & \multicolumn{4}{c}{0.9, 0.95} \\
Adam Weight Decay & \multicolumn{4}{c}{0.01} \\
Max Grad Norm & \multicolumn{4}{c}{1.0} \\
\midrule
\multicolumn{5}{c}{\textit{Transport \& Loss}} \\
\midrule
SNR Strategy & \multicolumn{2}{c|}{Dynamic Logit Normal \cite{esser2024scaling,flux2024blackforest, lipman2023flowmatching, liu2022rectifiedflow}} & \multicolumn{2}{c}{Rectified Dynamic Logit Normal (\ref{sec:rect_dyn_timeshift})} \\
Dynamic Time Shifting & \multicolumn{4}{c}{True} \\
\midrule
\multicolumn{5}{c}{\textit{Hardware \& Precision}} \\
\midrule
Sharding Strategy & HYBRID\_SHARD\_ZERO2 \cite{rajbhandari2020zero,zhao2023pytorchfsdp} & \multicolumn{3}{|c}{HYBRID\_SHARD \cite{zhao2023pytorchfsdp,gugger2022accelerate}} \\
Mixed Precision & \multicolumn{4}{c}{bf16} \\
Gradient Checkpointing & \multicolumn{4}{c}{True} \\
\bottomrule
\end{tabular}
}
\caption{Training hyperparameters across different tasks and resolutions}
\label{tab:training_hyperparams}
\end{table}

\clearpage

\clearpage

\subsection{Lightweight Module Details}
\label{app:small_arch}

\clearpage

\section{Boosted Orthogonal Guidance}
\label{sec:bog}

Classifier-free guidance \citep{ho2022cfg} and related inference-time steering techniques have become indispensable for improving prompt adherence in text-to-image DiTs. However, aggressively amplifying guidance often trades alignment for degraded appearance, manifesting as oversaturation, flattened tones, over-smoothing, and visually implausible artifacts. In this section, we introduce \textbf{Boosted Orthogonal Guidance} (BOG), a new training-free inference method for DiT-based text-to-image generation that targets this failure mode while explicitly enhancing fine-grained details and cinematic atmosphere. 

Inspired by the key insight behind APG~\citep{sadat2025apg} -- namely that guidance updates can be meaningfully decomposed into components parallel and orthogonal to the model’s predicted flow -- BOG emphasizes the information-preserving orthogonal contribution to avoid color blow-up and texture collapse. Crucially, BOG departs from the conventional view that a DiT prediction is merely a flattened high-dimensional vector: we treat each per-step predicted tensor as a 2D matrix aligned with the image’s inherent spatial structure, echoing the matrix-aware perspective adopted in optimizer design (e.g., Moun-style~\citep{jordan2024muon} updates). Building on this view, BOG applies an efficient orthogonalization approximation to the predicted matrix at every denoising step, acting as a structured normalization that not only controls magnitude but also enriches the diversity of row/column directions. Unlike vector-norm rescaling -- which preserves a single global direction while adjusting only scale -- our matrix-structured normalization encourages higher effective rank, thereby retaining more informative variation across spatial dimensions. Empirically, this rank-preserving, orthogonality-boosted guidance leads to richer micro-structures, improved tonal depth, and more coherent “film-like” lighting and ambience, while mitigating oversaturation without introducing additional training or meaningful sampling overhead.

\subsection{Matrix-Structured View of DiT Predictions}

Standard inference in DiTs typically treats the model prediction at each flow-matching (or denoising) step as a single flattened 1D update direction in a high-dimensional state space: under conditional or unconditional prompting, the current latent state (often a noisy or partially denoised image representation) is moved along this vector field toward a corresponding clean-image hidden state~\citep{sohldickstein2015diffusion,ho2020ddpm,song2021sde,song2022ddim,lu2022dpmsolver,lipman2023flowmatching,liu2022rectifiedflow,lu2025dpmsolverpp}. While this viewpoint is mathematically convenient, it implicitly discards a key inductive bias of the image modality -- namely, that the predicted hidden state is inherently spatial. Concretely, the DiT output is a tensor of shape 
$[B,C,H,W]$; after ignoring batch and channel dimensions, the remaining $[H,W]$ structure is not an arbitrary reshaping artifact but a meaningful 2D grid aligned with the image plane. This mismatch becomes more pronounced in how normalization is commonly applied: prevailing “norm” or rescaling strategies first collapse the 2D prediction into a 1D vector and then rescale its Euclidean length, which changes only the magnitude in the vector field while leaving the update direction (and its spatial composition) essentially untouched. Others additionally subtract a mean term -- either the batch mean or the per-sample mean over features -- thereby performing mean-centering. As a result, such vector-level normalization still overlooks spatial organization that could be leveraged to preserve informative variation and avoid overly smooth updates. Motivated by recent optimizer designs that explicitly respect the matrix/tensor structure rather than reducing updates to flat vectors~\citep{gupta2018shampoo,jordan2024muon}, we instead adopt a 2D normalization strategy that operates directly on the matrix-structured prediction at each step.

\subsubsection{Matrix Normalization}

\newcommand{\Wlog}{\textit{w.l.o.g.}}

\newcommand{\U}{\bm{U}}
\newcommand{\V}{\bm{V}}
\newcommand{\D}{\bm{D}}
\newcommand{\Omat}{\bm{O}}
\newcommand{\Q}{\bm{Q}}
\newcommand{\Sig}{\bm{\Sigma}}
\newcommand{\e}{\bm{e}}
\newcommand{\Imat}{\bm{I}}
\newcommand{\Rd}{\mathbb{R}}
\newcommand{\Nd}{\mathbb{N}}
\newcommand{\Np}{\mathbb{N}_{+}}

\newcommand{\x}{\bm{x}}
\newcommand{\ctx}{\bm{c}}

\newcommand{\dd}{\mathrm{d}}
\renewcommand{\algorithmiccomment}[1]{\hfill $\triangleright$ \textit{#1}}

Let $\D\in\Rd^{m\times n}$ (Assume $m\ge n$, {\Wlog}), we perform the singular value decomposition (SVD) of the matrix 
$\D$ as follows. 
\begin{equation}
\D=\U\Sig\V^\top
=\U \cdot
\left(
\begin{array}{cccc}
\sigma_{1} & 0 & \cdots & 0 \\
0 & \sigma_{2} & \cdots & 0 \\
\vdots & \vdots & \ddots & \vdots \\
0 & 0 & \cdots & \sigma_{n} \\
\hline
0 & 0 & \cdots & 0 \\
\multicolumn{4}{c}{\vdots}
\end{array}
\right) \cdot
\V^\top,
\qquad \Sig\in\Rd^{m\times n},\ ~\Wlog~\ m\ge n,
\end{equation}

We propose the following scalar function, whose inputs are the singular value index $i$ and the singular value itself.
\begin{equation}
\sigma_i^{\prime}=norm(\sigma_{i}, i)=
\left\{
\begin{array}{ll}
\mathrm{Sigmoid}(\sigma_{i}), \\[6pt]
\lambda \cdot \dfrac{\exp(\sigma_i/\tau)}{\sum_{j=1}^{n}\exp(\sigma_j/\tau)}, \\[10pt]
\cdots, \\[6pt]
1.
\end{array}
\right.
\end{equation}

where the $norm$ function can be chosen as needed; for example, we provide the options as above. Future work could further and more thoroughly investigate the differences, advantages, and disadvantages of these choices. Here $\lambda$ is a hyperparameter, which can, for example, be set to around $ (\frac{1}{n} \sum_{i=1}^n\delta_i)^{-1}$ in which $\delta_i={e^{\sigma_i/\tau}}/{\sum_{j=1}^{n}e^{\sigma_j/\tau}}$. Thus, we propose the following matrix-oriented normalization method.
\begin{equation}
\text{MNorm}(\D)
=\U \cdot
\left(
\begin{array}{cccc}
norm(\sigma_{1},1) & 0 & \cdots & 0 \\
0 & norm(\sigma_{2},2) & \cdots & 0 \\
\vdots & \vdots & \ddots & \vdots \\
0 & 0 & \cdots & norm(\sigma_{n},n) \\
\hline
0 & 0 & \cdots & 0 \\
\multicolumn{4}{c}{\vdots}
\end{array}
\right) \cdot
\V^\top
\end{equation}
It is worth noting that when we set \(norm(\sigma_i,i)=1\), we obtain the following equivalent form.
\begin{equation}
\text{MNorm}(\D)= \arg\min_{\Omat \in \Rd^{m  \times n}}\{ \|\Omat - \D\|_F ~~\Big|~ \Omat^\top\Omat=\I_n ~~\text{or}~~ \Omat\Omat^\top=\I_m\} 
\end{equation}
Therefore, in this case, \(\text{MNorm}(\D)\) is essentially the optimal orthogonal (semi-orthogonal) approximation of the matrix \(\D\) (see Appendix~\ref{app:osa} for a detailed proof). In practical image generation, we do not necessarily need to compute the SVD explicitly; instead, we can estimate \(\text{MNorm}(\D)\) using the Newton--Schulz method~\citep{jordan2024muon}. The specific algorithmic procedure is also provided in Appendix~\ref{app:ns}.

\subsubsection{Matrix Orthogonal Decomposition}
\paragraph{Preliminary - Ordinary Differential Equation (ODE)}
We briefly review key concepts in diffusion models and introduce the notation used in this paper. $\x \!\sim\! p_{data}(\x)$ denotes a sample, and here we let ~$\x_t \!=\! \alpha(t)\x + \sigma(t)\bm{\epsilon}$~~with ~$\bm{\epsilon} \!\sim\! \mathcal{N}(\bm{0},\Imat)$, which describes the forward noising process in diffusion models. Here $\sigma(t)$ with $t \in [0,1]$ gives the noise shedule, \ie, the amount of noise at the time step $t$. Following~\cite{zhuo2024LuminaNext,wu2025omnigen2}, we adopt the setting \(\sigma(1)\!=\!0\) and \(\sigma(0)\!=\!\sigma_{\max}\), so that \(t\!=\!0\) corresponds to (approximately) pure noise and \(t\!=\!1\) corresponds to a clean sample. Building upon the Probability Flow ODE (PF-ODE) framework~\citep{song2021sde,2022karrasODE}, the forward diffusion process can be equivalently described by a deterministic ODE that preserves the marginal probability densities. From this continuous-time perspective, the generative process simply amounts to solving this empirical ODE to map a pure noise distribution to the clean data manifold, which is formulated as:
{%
\begingroup
\renewcommand{\thefootnote}{$\text{\#}^1$}
\begin{equation}
\dd\x_t = -\sigma(t) \nabla_{\x_t}\!\log p_t(\x_t)\dd\sigma(t) = -\dot{\sigma}(t)\sigma(t)\nabla_{\x_t}\!\log p_t(\x_t)\dd t~~\footnotemark[1]
\end{equation}
\footnotetext[1]{~For our setting \(\sigma(1)\!=\!0\) and \(\sigma(0)\!=\!\sigma_{\max}\), ~\(\sigma(t) \!\ge\! 0\) is a monotonically decreasing function with \(\dot{\sigma}(t) \!\le\! 0\), ~and \(\dd t \!\ge\! 0\) during generation.}
\endgroup
\addtocounter{footnote}{-1}%
}%
\!\!where \(p_t(\x_t)\) represents the marginal distribution of the perturbed data at time \(t\), bridging the Gaussian noise \(p_0\!=\! \mathcal{N}(\bm{0}, \sigma_{\max}^2\Imat)\) and the target data distribution \(p_1\!=\!p_{\text{data}}\). Generative sampling is performed by numerically integrating this ODE forward from \(t\!=\!0\) to \(t\!=\!1\). To operationalize this process, the analytically intractable score \(\nabla_{\x_t}\!\log p_t(\x_t)\) is approximated by a Diffusion Transformer (DiT) \(\D_{\theta}(\x_t, t)\), which is optimized to recover the uncorrupted sample \(\x\) from the noisy observation \(\x_t\). For conditional generation, the DiT architecture is readily augmented to \(\D_{\theta}(\x_t, t, \ctx)\), enabling auxiliary inputs \(\ctx\), such as text instructions or class labels, to guide the generation.

\paragraph{Preliminary - Classifier-Free Guidance (CFG)}
CFG~\citep{ho2022cfg} is a widely adopted inference technique that enhances sample quality and condition alignment by extrapolating between conditional and unconditional model predictions. Given a null condition\footnote{Here, \(\ctx_{\text{neg}}\) denotes a context condition derived from negative text prompts (e.g., ``low quality, deformed, blurry, ...''). In practice, many models substitute the standard empty condition \(\ctx_{\varnothing}\) with \(\ctx_{\text{neg}}\). \Wlog, we use \(\ctx_{\varnothing}\) to broadly represent either the true unconditional state \(\ctx_{\varnothing}\) or \(\ctx_{\text{neg}}\).} $\ctx_{\varnothing}$, CFG modifies the output of the DiT denoiser at each sampling time step $t$ as follows:
\begin{equation}
\begin{aligned}
\D_{CFG}(\x_t,t,\ctx) &= \D_{\theta}(\x_t,t,\ctx_{\varnothing}) + \omega \left( \D_{\theta}(\x_t,t,\ctx) - \D_{\theta}(\x_t, t, \ctx_{\varnothing}) \right)\\
&= \D_{\theta}(\x_t,t,\ctx) + (\omega - 1)\Delta \D_{\theta,\x_t,t}
\end{aligned}
\label{eq:cfg}
\end{equation}
where $\Delta \D_{\theta,\x_t,t} = \D_{\theta}(\x_t,t,\ctx) - \D_{\theta}(\x_t, t, \ctx_{\varnothing})$. The DiT output is typically a tensor of shape $[B,C,H,W]$; in some cases for simplicity, we can omit the batch size $B$ and channel dimension $C$, treating it as an $[H, W]$ matrix. The unconditional prediction $\D_{\theta}(\x_t,t,\ctx_{\varnothing})$ is conventionally obtained by randomly dropping the condition $\ctx$ during training, though a separate unconditional DiT denoiser can also be trained for this purpose~\citep{karras2024altercfg}. Similar to the truncation trick in GANs~\citep{brock2018large}, CFG improves image quality at the expense of generation diversity~\citep{pml2Book}. While a larger guidance scale $\omega$ generally yields better instruction alignment, recent studies indicate that at high $\omega$ values, the parallel component of $\Delta \D_{\theta,\x_t,t}$ with respect to $\D_{\theta}(\x_t,t,\ctx)$ (when flattened into a 1D vector) leads to image over-saturation~\citep{sadat2025apg}.

\paragraph{Parallel-Orthogonal Decomposition}
For notational convenience, we denote $\Delta \D \!\coloneq\! \Delta \D_{\theta,\x_t,t} \!\in\! \Rd^{H \times W}$ and $\D^{(\ctx)} \!\coloneq\! \D_{\theta}(\x_t,t,\ctx) \!\in\! \Rd^{H \times W}$. Furthermore, let $\D^{(\ctx)}_{[i,:]}$ denote the $i$-th row vector of the DiT output matrix $\D^{(\ctx)}$, and similarly, let $\D^{(\ctx)}_{[:,j]}$ denote its $j$-th column vector ; this notation applies analogously to all other matrices. For instance, taking a column vector $\D^{(\ctx)}_{[:,j]}$ from the conditional DiT output matrix as the reference direction, we can perform a parallel-orthogonal decomposition using vector functions $\text{Parall}(\cdot,\cdot)$ and $\text{Ortho}(\cdot,\cdot)$. Akin to Gram-Schmidt orthogonalization, this process is formulated as:
\begin{equation}
\begin{aligned}
\Delta \D^{\parallel}_{[:,j]} &= \text{Parall}(\D_{[:,j]},~\D^{(\ctx)}_{[:,j]})  =  \frac{ \Delta {\D_{[:,j]}}^{\top} \D^{(\ctx)}_{[:,j]} }{ \|  \D^{(\ctx)}_{[:,j]} \|^2_2 } \cdot\D^{(\ctx)}_{[:,j]} \\
\Delta \D^{\perp}_{[:,j]} &= \text{Ortho}(\D_{[:,j]},~\D^{(\ctx)}_{[:,j]}) = \Delta \D_{[:,j]} - \Delta \D^{\parallel}_{[:,j]}\\
\end{aligned}    
\end{equation}
By applying the aforementioned decomposition to each column $j$, we obtain the full matrices $\Delta \D^{\parallel}_{\text{col}} \!\in\! \Rd^{H \times W}$ and $\Delta \D^{\perp}_{\text{col}} \!\in\! \Rd^{H \times W}$. Similarly, performing this decomposition on each row $i$ yields the matrices $\Delta \D^{\parallel}_{\text{row}},~\Delta \D^{\perp}_{\text{row}} \!\in\! \Rd^{H \times W}$. This process is formulated as:
\begin{equation}
\begin{aligned}
\Delta \D^{\parallel}_{\text{col}} = \text{Stack}\left(\{ \text{Parall}(\D_{[:,j]},~\D^{(\ctx)}_{[:,j]}) ~\Big|~ j=0,\cdots,W-1 \} \right) \\
\Delta \D^{\parallel}_{\text{row}} = \text{Stack}\left(\{ \text{Parall}(\D_{[i,:]},~\D^{(\ctx)}_{[i,:]}) ~\Big|~ i=0,\cdots,H-1 \} \right) 
\end{aligned}    
\end{equation}
Similarly, $\Delta \D^{\perp}_{\text{col}}$ and $\Delta \D^{\perp}_{\text{row}}$ can be obtained using $\text{Ortho}(\cdot,\cdot)$. In summary, we have:
\begin{equation}
\begin{aligned}
\Delta\D_{\text{pod}} = \text{POD}(\Delta \D, \D^{(\ctx)})  = \lambda_{c} \cdot (\Delta \D^{\perp}_{\text{col}} + \mu \Delta \D^{\parallel}_{\text{col}} ) + \lambda_r \cdot (\Delta \D^{\perp}_{\text{row}} + \mu \Delta \D^{\parallel}_{\text{row}} )
\end{aligned}    
\end{equation}
where $\lambda_c = \frac{W}{H+W+\zeta}$ and $\lambda_r=\frac{H}{H+W+\zeta}$, with $\zeta$ being a small constant. The hyperparameter $\mu$ controls the magnitude of the parallel components. In practice, we set $\mu \le 1$ (e.g., $0, 0.1, 0.25$) to mitigate image over-saturation when $\omega$ is relatively large \citep{sadat2025apg}.

\subsection{Rolling-Sum Momentum}
Inspired by momentum strategies commonly used in optimization, we introduce a similar mechanism during the inference of text-to-image generation. This is motivated by the intuition that the "inertia" of the trajectory within the flow matching vector field influences the final generation. Our momentum update is simply formulated as $\overline{\Delta \D}_{(i)}= \eta \Delta \D_{(i)} + \rho\overline{\Delta \D}_{(i-1)}$, with typical default values of $\eta=0.9$ and $\rho=0.1$. Here, $i \!\in\! \Np$ denotes the inference iteration step, which is related to the diffusion time step $t \!\in\! \Rd$ fed into the DiT via a scheduler function, \ie, $t = \text{scheduler}(i)$. We abstract this momentum update as a function:
\begin{equation}
\begin{aligned}
\overline{\Delta \D}_{(i)} = \text{RollSumUpdate}(\Delta \D, i) = \eta \Delta \D_{(i)} + \rho\overline{\Delta \D}_{(i-1)} ~~\Big|~~\overline{\Delta \D}_{(0)}=\bm{0}
\end{aligned}    
\end{equation}
This contrasts with recent findings by \cite{sadat2025apg}, who propose a "reverse momentum" (\ie, $\rho \!<\! 0$). They argue that this pushes the model away from previous CFG update directions and encourages it to focus more on the current one. However, our empirical observations indicate that setting $\rho \!<\! 0$ causes directional jitter for $\overline{\Delta \D}_{(i)}$ within the vector field across iterations.

\subsection{Putting It All Together}

We now present the complete pipeline of our proposed method. Algorithm~\ref{alg:bog_step} summarizes the computation of the Boosted Orthogonal Guidance (BOG) at a single inference iteration step \(i\). 

By explicitly treating the predicted tensors as spatial matrices, BOG systematically refines the guidance direction. It first smooths the trajectory using momentum, then normalizes the matrix to preserve spatial rank, and finally isolates the orthogonal components---effectively diminishing the projections parallel to the conditional prediction---to prevent oversaturation. Ultimately, this rank-preserving and orthogonality-boosted approach significantly elevates the visual quality, producing richer micro-details, enhanced tonal depth, and a realistic cinematic lighting, all achieved zero-shot without meaningful overhead.
\begin{algorithm}[ht]
\caption{Single Step of Boosted Orthogonal Guidance (BOG)}
\label{alg:bog_step}
\textbf{Input:} Inference step \(i \in \mathbb{N}_{+}\), current latent state \(\x_t\), condition \(\ctx\), unconditional state \(\ctx_{\varnothing}\), guidance scale \(\omega\). \\
\textbf{Output:} The BOG modified prediction \(\D_{{BOG}_{(i)}}\).
\begin{algorithmic}[1]
\STATE \(t \leftarrow \text{scheduler}(i)\) \COMMENT{Obtain the continuous diffusion time step}
\STATE \(\Delta \D^{(\ctx)}_{(i)} \leftarrow \D_{\theta}(\x_t, t, \ctx)\), \quad \(\Delta \D^{(\ctx_{\varnothing})}_{(i)} \leftarrow \D_{\theta}(\x_t, t, \ctx_{\varnothing})\) \COMMENT{Compute DiT predictions}
\STATE \(\Delta \D_{(i)} \leftarrow \Delta \D^{(\ctx)}_{(i)} - \Delta \D^{(\ctx_{\varnothing})}_{(i)}\) \COMMENT{Calculate the raw guidance direction}
\STATE \(\widetilde{\Delta \D}_{(i)} \leftarrow \text{MNorm}\Big(\text{RollSumUpdate}(\Delta \D_{(i)}, i)\Big)\) \COMMENT{Apply momentum \& matrix normalization}
\STATE \(\widehat{\Delta \D}_{(i)} \leftarrow \text{POD}\big(\widetilde{\Delta \D}_{(i)}, \Delta \D^{(\ctx)}_{(i)}\big)\) \COMMENT{Parallel-Orthogonal Decomposition}
\STATE \(\D_{{BOG}_{(i)}} \leftarrow \Delta \D^{(\ctx)}_{(i)} + (\omega - 1)\widehat{\Delta \D}_{(i)}\) \COMMENT{Compute final BOG tensor}
\RETURN \(\D_{{BOG}_{(i)}}\)
\end{algorithmic}
\end{algorithm}

\paragraph{Limitations} While BOG effectively enhances the cinematic atmosphere of generated images, it introduces certain limitations. Specifically, it can increase the probability of structural distortions and reduce the success rate of text rendering. Consequently, we treat BOG as an optional module, akin to a selectable filter.

To mitigate these issues, we introduce a simple hyperparameter: the BOG Interval (\(\Delta_{\text{BOG}}\)). During inference, BOG is applied every \(\Delta_{\text{BOG}}\) diffusion steps, while standard CFG is used for the remaining steps. For instance, in a 10-step denoising process with \(\Delta_{\text{BOG}}=2\), BOG is executed on even steps (2, 4, 6, 8, 10) and standard CFG on odd steps. Thus, \(\Delta_{\text{BOG}}=1\) denotes full BOG inference. We recommend \(\Delta_{\text{BOG}}=2\) as the default setting, which significantly alleviates the aforementioned artifacts, albeit with a slight trade-off in the enhanced BOG-styled photographic texture.
\clearpage

\subsection{Optimal Semi-Orthogonal Approximation}
\label{app:osa}
\begin{proof}
Let $\D\in\Rd^{m\times n}$ with $m\ge n$, and let $\Omat\in\Rd^{m\times n}$ be semi-orthogonal, i.e.\ $\Omat^{\top}\Omat=\I_{n}$.
Let the (thin) SVD of $\D$ be
\[
\D = \U \Sig \V^{\top},
\]
where $\U\in\Rd^{m\times n}$ satisfies $\U^{\top}\U=\I_{n}$, $\V\in\Rd^{n\times n}$ is orthogonal, and
\[
\Sig=\mathrm{diag}(\sigma_{1},\dots,\sigma_{n}),\qquad
\sigma_{1}\ge \cdots \ge \sigma_{n}\ge 0.
\]

\paragraph{Remark on $\Q$.}
Although $\U$ and $\Omat$ are rectangular (semi-orthogonal), the matrix
\[
\Q := \U^{\top}\Omat\V
\]
is \emph{square} ($n\times n$). Moreover,
\[
\Q^{\top}\Q=\I_{n},
\]
so $\Q$ is an orthogonal matrix in $\Rd^{n\times n}$ (not merely semi-orthogonal). Indeed, for a square matrix, $\Q^{\top}\Q=\I_{n}$ already implies $\Q$ is orthogonal (equivalently, $\Q^{-1}=\Q^{\top}$ and also $\Q\Q^{\top}=\I_{n}$).

\paragraph{Step 1 (Reduction to trace maximization).}
Expand the squared Frobenius norm:
\begin{align*}
\|\D-\Omat\|_{F}^{2}
&= \mathrm{tr}\!\big((\D-\Omat)^{\top}(\D-\Omat)\big) \\
&= \mathrm{tr}(\D^{\top}\D) + \mathrm{tr}(\Omat^{\top}\Omat) - 2\,\mathrm{tr}(\Omat^{\top}\D) \\
&= \|\D\|_{F}^{2} + \|\Omat\|_{F}^{2} - 2\,\mathrm{tr}(\Omat^{\top}\D).
\end{align*}
Since $\Omat^{\top}\Omat=\I_{n}$,
\[
\|\Omat\|_{F}^{2}=\mathrm{tr}(\Omat^{\top}\Omat)=\mathrm{tr}(\I_{n})=n.
\]
Thus minimizing $\|\D-\Omat\|_{F}^{2}$ over $\Omat^{\top}\Omat=\I_{n}$ is equivalent to maximizing $\mathrm{tr}(\Omat^{\top}\D)$.

\paragraph{Step 2 (SVD coordinates).}
Using $\D=\U\Sig\V^{\top}$ and cyclicity of the trace,
\begin{align*}
\mathrm{tr}(\Omat^{\top}\D)
&= \mathrm{tr}\!\big(\Omat^{\top}\U\Sig\V^{\top}\big) \\
&= \mathrm{tr}\!\big(\V^{\top}\Omat^{\top}\U\Sig\big) \\
&= \mathrm{tr}\!\big(\Q^{\top}\Sig\big),
\end{align*}
where $\Q=\U^{\top}\Omat\V\in\Rd^{n\times n}$.

\paragraph{Step 3 (Elementwise bound via diagonal entries).}
Since $\Sig$ is diagonal,
\[
\mathrm{tr}(\Q^{\top}\Sig)=\sum_{i=1}^{n}\sigma_{i}\,q_{ii}.
\]
For each $i$, let $\e_{i}$ be the $i$th standard basis vector. Then $q_{ii}=\e_{i}^{\top}\Q\e_{i}$ and
\begin{align*}
|q_{ii}|
=|\e_{i}^{\top}\Q\e_{i}|
&\le \|\e_{i}\|_{2}\,\|\Q\e_{i}\|_{2} \\
&\le \|\e_{i}\|_{2}\,\|\Q\|_{2}\,\|\e_{i}\|_{2}.
\end{align*}
Because $\Q$ is orthogonal, $\|\Q\|_{2}=1$; also $\|\e_{i}\|_{2}=1$. Hence $|q_{ii}|\le 1$.
Since $\sigma_{i}\ge 0$, we obtain
\begin{align*}
\mathrm{tr}(\Q^{\top}\Sig)
=\sum_{i=1}^{n}\sigma_{i}q_{ii}
&\le \sum_{i=1}^{n}\sigma_{i}|q_{ii}|
\le \sum_{i=1}^{n}\sigma_{i}.
\end{align*}
Therefore, for all feasible $\Omat$,
\[
\mathrm{tr}(\Omat^{\top}\D)\le \sum_{i=1}^{n}\sigma_{i}.
\]

\paragraph{Step 4 (Achievability).}
Let
\[
\Omat_{\star}:=\U\V^{\top}.
\]
Then $\Omat_{\star}^{\top}\Omat_{\star}=\V\U^{\top}\U\V^{\top}=\I_{n}$, and
\[
\Q_{\star}=\U^{\top}\Omat_{\star}\V=\U^{\top}(\U\V^{\top})\V=\I_{n}.
\]
Hence
\[
\mathrm{tr}(\Omat_{\star}^{\top}\D)
=\mathrm{tr}(\Q_{\star}^{\top}\Sig)
=\mathrm{tr}(\Sig)
=\sum_{i=1}^{n}\sigma_{i},
\]
so the upper bound is tight. Thus $\Omat_{\star}$ maximizes $\mathrm{tr}(\Omat^{\top}\D)$ and therefore minimizes $\|\D-\Omat\|_{F}$.

\paragraph{Minimum value.}
Using Step~1 and $\max_{\Omat^{\top}\Omat=\I_{n}}\mathrm{tr}(\Omat^{\top}\D)=\sum_{i=1}^{n}\sigma_{i}$,
\[
\min_{\Omat^{\top}\Omat=\I_{n}}\|\D-\Omat\|_{F}^{2}
=\|\D\|_{F}^{2}+n-2\sum_{i=1}^{n}\sigma_{i}.
\]
\end{proof}

\bigskip

\noindent\textbf{Supplement (same bound via Von Neumann's trace inequality).}
With $\Q=\U^{\top}\Omat\V$ we have $\mathrm{tr}(\Omat^{\top}\D)=\mathrm{tr}(\Q^{\top}\Sig)$.
Von Neumann's trace inequality~\citep{Neumann1937} states that for any $A,B\in\Rd^{n\times n}$,
\[
|\mathrm{tr}(A^{\top}B)| \le \sum_{i=1}^{n}\sigma_{i}(A)\,\sigma_{i}(B).
\]
Taking $A=\Q$ and $B=\Sig$ gives
\begin{align*}
\mathrm{tr}(\Q^{\top}\Sig)
&\le |\mathrm{tr}(\Q^{\top}\Sig)|
\le \sum_{i=1}^{n}\sigma_{i}(\Q)\,\sigma_{i}(\Sig).
\end{align*}
Since $\Q$ is orthogonal, $\sigma_{i}(\Q)=1$ for all $i$, and $\sigma_{i}(\Sig)=\sigma_{i}$, hence
\[
\mathrm{tr}(\Omat^{\top}\D)=\mathrm{tr}(\Q^{\top}\Sig)\le \sum_{i=1}^{n}\sigma_{i},
\]
with equality at $\Q=\I_{n}$ (equivalently, $\Omat=\U\V^{\top}$), recovering the same optimizer and the same minimum value.

\clearpage
\subsection{Newton-Schulz Approximation}
\label{app:ns}

To compute \(\text{MNorm}(\D)\), it is not strictly necessary to perform Singular Value Decomposition (SVD). Specifically, for the case where \(\text{norm}(\sigma_i, i) = 1\), we can directly estimate it using the Newton-Schulz method, following the approach of \citep{jordan2024muon} (see Code \ref{code:newton_schulz}).

\definecolor{codegreen}{rgb}{0,0.6,0}
\definecolor{codegray}{rgb}{0.5,0.5,0.5}
\definecolor{codepurple}{rgb}{0.58,0,0.82}
\definecolor{backcolour}{rgb}{0.95,0.95,0.95} %

\renewcommand{\lstlistingname}{Code}

\lstdefinestyle{pystyle}{
    backgroundcolor=\color{backcolour},   
    commentstyle=\color{codegreen},
    keywordstyle=\color{blue},
    numberstyle=\tiny\color{codegray},
    stringstyle=\color{codepurple},
    basicstyle=\ttfamily\footnotesize,
    breakatwhitespace=false,         
    breaklines=true,                 
    captionpos=b,                    
    keepspaces=true,                 
    numbers=left,                    
    numbersep=5pt,                  
    showspaces=false,                
    showstringspaces=false,
    showtabs=false,                  
    tabsize=4,
    language=Python
}

\begin{lstlisting}[style=pystyle, caption={Batched Newton-Schulz Iteration in PyTorch}, label={code:newton_schulz}]
def _newtonschulz5_batched(G: torch.Tensor, steps: int = 5, eps: float = 1e-7):
    """
    Batched Newton-Schulz iteration.

    Accepts:
    - (H, W)          -> returns (H, W)
    - (N, H, W)       -> returns (N, H, W)
    - (B, C, H, W)    -> returns (B, C, H, W)
    """
    a, b, c = (3.4445, -4.7750, 2.0315)

    orig_ndim = G.ndim
    if orig_ndim == 2:
        G3 = G.unsqueeze(0)                 # (1, H, W)
        out_shape = None
    elif orig_ndim == 3:
        G3 = G                               # (N, H, W)
        out_shape = None
    elif orig_ndim == 4:
        B, C, H, W = G.shape
        G3 = G.reshape(B * C, H, W)          # (N, H, W)
        out_shape = (B, C, H, W)
    else:
        raise ValueError(f"Expected 2D/3D/4D tensor, got ndim={G.ndim}")

    # Match the original behavior: decide whether to transpose based on H/W
    H, W = G3.shape[-2], G3.shape[-1]

    # Compute in bfloat16 (keeps the original logic)
    X = G3.to(torch.bfloat16)

    # Normalize each matrix by its Frobenius norm: X /= (||X||_F + eps)
    # Frobenius norm = sqrt(sum_ij X^2)
    nrm = torch.linalg.norm(X, ord="fro", dim=(-2, -1))  # (N,)
    X = X / (nrm.unsqueeze(-1).unsqueeze(-1) + eps)

    transposed = False
    if H > W:
        # Transpose the last two dims so we iterate on the "shorter" dimension first
        X = X.transpose(-2, -1)  # (N, W, H)
        transposed = True

    # Newton-Schulz iterations (batched GEMMs)
    for _ in range(steps):
        A = X @ X.transpose(-2, -1)   # (N, m, m)
        Bm = b * A + c * (A @ A)      # (N, m, m)
        X = a * X + (Bm @ X)          # (N, m, n)

    # Transpose back if we transposed at the beginning
    if transposed:
        X = X.transpose(-2, -1)

    # Restore original shape
    if orig_ndim == 2:
        return X.squeeze(0)
    if out_shape is not None:
        return X.reshape(out_shape)
    return X
\end{lstlisting}

\clearpage

\end{CJK}
\end{document}